\documentclass[letter, 12pt]{article} % Remove draft to render pics 
\usepackage{newtxtext,newtxmath, amsmath, amsfonts, bm,mathtools}
\usepackage{paralist, url, natbib, parskip}
\usepackage{adjustbox, todonotes} % , 
\usepackage{graphicx, subfig}
\usepackage{rotating, booktabs, multirow}
\usepackage[makeroom]{cancel}
\usepackage[section]{placeins}
\usepackage[linesnumbered,lined,boxed,commentsnumbered]{algorithm2e}
\usepackage{xcolor}
\usepackage{placeins}
\usepackage{arydshln}
\usepackage{pgf,tikz,pgfplots}
\pgfplotsset{compat=1.15}
\usepackage{mathrsfs}
\usepackage{arydshln}
\usetikzlibrary{arrows}
\usetikzlibrary{decorations.markings}
\usetikzlibrary{positioning, shadows}
\graphicspath{{figures/}}

\usepackage[margin=3cm]{geometry}
\usepackage[margin=2cm]{caption}
\usepackage{titling}
\usepackage{setspace}
\DeclareCaptionStyle{italic}[justification=centering]{labelfont={bf},textfont={it},labelsep=colon}
\SetKwInOut{Parameter}{parameter}
\DeclareMathOperator{\sign}{sign}
\mathtoolsset{showonlyrefs}


\tikzset{pics/gear/.style n args={3}{
    code={
        \def\modu{#1}
        \def\Zb{#2}
        \def\AngleA{#3}

        \pgfmathsetmacro{\Rpr}{\Zb*\modu/2}
        \pgfmathsetmacro{\Rb}{\Rpr*cos(\AngleA)}
        \pgfmathsetmacro{\Rt}{\Rpr+\modu}
        \pgfmathsetmacro{\Rp}{\Rpr-1.25*\modu}
        \pgfmathsetmacro{\AngleT}{pi/180*acos(\Rb/\Rt)}
        \pgfmathsetmacro{\AnglePr}{pi/180*acos(\Rb/\Rpr)}
        \pgfmathsetmacro{\demiAngle}{180/\Zb}
        \pgfmathsetmacro{\Angledecal}{(\demiAngle-2*\AnglePr)/2}

        \path[pic actions] foreach \zz in{1,...,\Zb}{
            \ifnum\zz=1
                (\zz/\Zb*360-\Angledecal:\Rp)
            \else
                -- (\zz/\Zb*360-\Angledecal:\Rp)
            \fi
            to[bend right=\demiAngle]
            (\zz/\Zb*360+\Angledecal:\Rp)
            --
            plot[domain=-0:\AngleT,smooth,variable=\t]
                ({{180/pi*(-\t+tan(180/pi*\t)) +\zz/\Zb*360+\Angledecal}:\Rb/cos(180/pi*\t)})
            to[bend right=\demiAngle]
                ({{180/pi*(\AngleT+tan(180/pi*-\AngleT)) +(\zz+1)/\Zb*360-\Angledecal}:
                \Rb/cos(180/pi*-\AngleT)})
            --
            plot[domain=-\AngleT:-0,smooth,variable=\t]
            ({{180/pi*(-\t+tan(180/pi*\t)) +(\zz+1)/\Zb*360-\Angledecal}:\Rb/cos(180/pi*\t)})
        } -- cycle;
    }
}}
\tikzstyle{startstop} = [rectangle, rounded corners, minimum width=3cm, minimum height=1cm,text centered, draw=black, fill=black!30]
\tikzstyle{arrow} = [thick,->,>=stealth]

\allowdisplaybreaks
\begin{document}
% =======================================================================
% \tikzset{->-/.style={decoration={
%   markings,
%   mark=at position #1 with {\arrow{>}}},postaction={decorate}}}

\title{Comprehensive Algorithm Portfolio Evaluation using Item Response Theory}

\author{Sevvandi Kandanaarachchi$^1$, Kate Smith-Miles$^2$}
\date{%
   \scriptsize{ $^1$ CSIRO's Data61, Clayton, VIC 3168, Australia\\ [2ex]%
   $^2$ School of Mathematics and Statistics, University of Melbourne, Parkville, VIC 3010, Australia.} \\
    \today \\
}
       
%\editor{Kevin Murphy and Bernhard Sch{\"o}lkopf}
\maketitle

\begin{abstract}
Item Response Theory (IRT) has been proposed within the field of Educational Psychometrics to assess student ability as well as test question difficulty and discrimination power. More recently, IRT has been applied to evaluate machine learning algorithm performance on a single classification dataset, where the student is now an algorithm, and the test question is an observation to be classified by the algorithm. In this paper we present a modified IRT-based framework for evaluating a portfolio of algorithms across a repository of datasets, while simultaneously eliciting a richer suite of characteristics - such as algorithm consistency and anomalousness - that describe important aspects of algorithm performance. These characteristics arise from a novel inversion and reinterpretation of the traditional IRT model without requiring additional dataset feature computations. We test this framework on algorithm portfolios for a wide range of applications, demonstrating the broad applicability of this method as an insightful algorithm evaluation tool. Furthermore, the explainable nature of IRT parameters yield an increased understanding of algorithm portfolios. 
\end{abstract}


%\newgeometry{top=1.5cm,bottom=2cm,right=1.5cm,left=1.5cm}
% =======================================================================
\section{Introduction}\label{sec:Intro}
% =======================================================================
Evaluating a diverse set algorithms across a comprehensive set of test problems contributes to an increased understanding of the interplay between test problem characteristics, algorithm mechanisms and algorithm performance. Such an evaluation helps determine an algorithm's strengths and weaknesses, and provides a broad overview of the collective capabilities of an algorithm portfolio. The drawback of  many studies that evaluate only a small number of algorithms on a limited set of test problems is that they fail to reveal where any algorithm belongs within a state-of-the-art algorithm portfolio's capabilities, or where the unique strengths and weaknesses of algorithms lie considering a diverse range of test problem difficulties and challenges. After several decades of calls for a more "empirical science" of algorithm testing \citep{hooker1994needed,hooker1995testing}, research communities in many fields are now pulling together the components needed for rigorous evaluation of algorithms - open source algorithms and shared test problem repositories - that provide the foundation for new methodologies for empirical evaluations \citep{mcgeoch1996toward,hall2010generation,smith2014towards,casalicchio2019openml,Bischl2016}.

In this paper we present a framework that evaluates a portfolio of algorithms based on a novel adaptation of Item Response Theory (IRT). The general premise of IRT is that there is a hidden ``quality'' or a trait, such as verbal or mathematical ability, that cannot be directly measured \citep{hambleton2013item} but can be inferred from responses to well-designed test questions that are suitably difficult and discriminating. A test instrument such as a questionnaire or an exam containing test items is used to capture participant responses.  Using the participant responses to the test items, an IRT model is fitted to estimate the discrimination and difficulty of test items and the ability of participants.   In an educational setting, the ability relates to the knowledge of the subject matter tested on the exam; the discrimination of test items inform us which items are better at discriminating between strong and weak students; and the difficulty parameters indicate the  difficulty of each test item given the response profile from the participants.

IRT's ability to evaluate performance data and obtain useful insights has made it a natural fit for adaptation to the machine learning domain. 
\cite{martinez2019item} used IRT to evaluate the performance of machine learning algorithms (students in the educational analogy) on a single classification dataset (exam), with the individual observations in a classification dataset (exam questions) used to assess algorithm performance. They train and test many classifiers on a single dataset, and obtain insights about the individual observations and about the portfolio of classifiers on that dataset. As a result they obtain a set of classifier characteristic curves for the dataset. Another IRT based evaluation of algorithm portfolios was carried out by \cite{chen2019beta}. They proposed a model called $\beta^3$-IRT, which extends the Beta IRT model for continuous responses discussed by \cite{YvonnickNoel2007}. \cite{chen2019beta} consider new parametrizations so that the resulting item characteristic curves are not limited to logistic curves and use their model to assess machine learning classifiers. They too evaluate an algorithm portfolio on an individual dataset and draw their conclusions about which algorithm is best for a given observation within a dataset.  Both \cite{martinez2019item} and \cite{chen2019beta} investigate IRT on an individual dataset, which we call a test instance. 

These exciting directions have motivated us to expand the use of IRT for understanding the strengths and weaknesses of a portfolio of algorithms when applied to \emph{any} dataset, not just a single dataset. In this case, the test instance is an entire dataset comprising observations, and the `exam' is comprised of many datasets to evaluate the ability of an algorithm.  Extending in this direction is important because the limited amount of diversity contained within a single dataset can shed only a limited amount of light on a portfolio of classifiers, and the classifier characteristic curves heavily depend on the dataset. To obtain a better understanding of the strengths and weaknesses of a portfolio of classifiers, indeed any type of algorithm, we need to evaluate the portfolio on a broader range of datasets from diverse repositories. The excellent foundational work showing how IRT models - with both discrete \citep{martinez2019item} and continuous \citep{chen2019beta} performance metrics - can be used to study performance of machine learning algorithms  is ripe for extension to see how the insights that can be generated from an IRT perspective compare with recent advances in algorithm portfolio evaluation and construction. 

In recent decades the call for a more empirical approach to algorithm testing~\citep{hooker1994needed} has seen efforts to move beyond  a standard statistical analysis  to evaluate algorithm portfolios, where strong algorithms have best ``on-average" performance across a chosen set of test instances.  Machine learning approaches such as meta-learning (``learning to learn'') have been used to learn how algorithm portfolios perform based on characteristics of the test instances \citep{vilalta2009meta}, with efforts encompassing a large body of research on topics such as algorithm selection, rankings, recommendation, and ensembles to name a few \citep{Lemke2015}. 
 \cite{Kevin2016} use Shapley values --  a concept from coalition game theory measuring a component's marginal contribution to the portfolio -- to gain insights into the value of an algorithm in a portfolio.
 
In a related but orthogonal direction,  emphasis in the literature on dataset repository design to facilitate unbiased algorithm evaluation is also a growing research area  \citep{Marcia2014,Bischl2016}, motivated by the fact that algorithms are frequently claimed to be superior without testing them on a demonstrably broad range of test instances. Demonstrating that a selected set of test instances or datasets is unbiased and sufficiently diverse is one of the major contributions of the Instance Space Analysis methodology \citep{smith2012measuring, smith2014towards,smith2015generating, munoz2018instance}, developed by Smith-Miles and co-authors by extending Rice's algorithm selection framework \citep{rice1976algorithm}. A $2D$ instance space is constructed by projecting all test instances into the instance space in a manner that maximize visual interpretation of the relationships between instance features and algorithm performance. The mathematical boundary defining the instance space can be determined, and the diversity of the test instances within the instance space can be scrutinized. Furthermore, the instance space analysis methodology can be used to answer the question posed by the Algorithm Selection Problem~\citep{rice1976algorithm}, ``Which algorithm is best suited for my problem?''. This aspect is missed by the standard statistical analysis, which focuses on average performances, and leaves hidden the unique strengths and weaknesses of algorithms  and relationships to test instance characteristics. 

Our main contribution in this paper is proposing a novel framework for evaluating algorithm portfolios across diverse suites of test instances based on IRT concepts. We call this framework AIRT -- Algorithmic IRT. The word \textit{airt} is an old Scottish word which means ``to guide''. By re-mapping the educational analogies of students, exams and test questions in a manner that is essentially flipped from the original approach on a single dataset of \cite{martinez2019item}, we propose an inverted model that yields a richer set of evaluation metrics for algorithm portfolios. Adapting continuous IRT models, we introduce measures for quantifying  algorithm \textit{consistency}, an algorithm's \textit{difficulty limit} in terms of the instances it can handle, and the degree of \textit{anomalousness} of an algorithm's behavior compared to others in the portfolio. %We also explore the problem space and find regions of best performance for algorithms and compute the proportion of the problem space occupied by each algorithm. 
We also explore the problem space and find regions of good and bad performance, which are effectively algorithm strengths and weaknesses. These other measures are not computed by standard statistical methodology used for ranking algorithms, nor are they available from the standard IRT mapping~\citep{martinez2019item,chen2019beta}. For example, the algorithm with the best overall performance on a suite of test problems may not be stable or consistent in the sense that a small change in a test instance may result in large changes in performance. Or there may be an anomalous algorithm that performs well on test instances for which other algorithms perform poorly, and such insights may be lost in standard `on-average' statistical analysis. Indeed, it is AIRT's focus on revealing insights into algorithm strengths and weaknesses, based on new methods for visual exploratory data analysis from empirical performance data results, that adds significant value beyond standard statistical analysis or  algorithm selection studies.

It is worthwhile noting that methodologies in social sciences  focus on explanations as opposed to accurate predictions \citep{Shmueli2010}.  As such, quantitative models in social sciences only have a handful of parameters which have meaningful interpretations. Explanations are often linked with causality. \cite{Lewis1986} states ``Here is my main thesis: \textit{to explain an event is to provide some information about its causal history.''} \cite{MILLER20191} presents an argument for linkages with social sciences stating that ``the field of explainable artificial intelligence can build on  existing research, and reviews relevant papers from philosophy, cognitive psychology/science, and social psychology, which study these topics.'' Indeed, AIRT is such a linkage.  In educational psychometrics IRT is used  to explain the student performance in terms of student ability and test item discrimination and difficulty. For example, difficult test items generally yield lower scores than easy test items. Similarly, students with high ability obtain higher scores compared to students with low ability. Thus, IRT model parameters are used to explain the student and test item characteristics and have causal interpretations. These explainable interpretations get translated to the algorithm evaluation setting as follows: problems with high difficulty generally result in low performance values. Algorithms with high difficulty limits can handle harder problems. Algorithms that are consistent give similar results irrespective of the problem difficulty. Anomalous algorithms behave in an unusual fashion by giving better results to harder problems compared to easier problems. We realise these statements are simple and obvious. But that is an attribute of an explanation;  {\color{black} \cite{OxfordEnglishDictionary} defines it as \textit{a thing which explains, makes clear, or accounts for something}.} Therefore, AIRT metrics come from an explainable model in educational psychometrics and contribute to increasing the explainability of algorithm performance.

Beyond insights and explanations however, AIRT can also be used for algorithm selection to construct a strong portfolio. In this paper we compare the predictive power of the AIRT portfolio to others generated by Shapley values \citep{Kevin2016} and best on average performance. The AIRT portfolio showcases algorithm strengths in different parts of the problem space. In addition to introducing these measures that capture different aspects of algorithm performance and constructing algorithm portfolios, we also assess the goodness of the IRT model by comparing the IRT predicted performance with the actual performance. As a further contribution, we make this work available in the R package \texttt{airt} \citep{airt}. Another point of interest is that, unlike in instance space analysis, we do not need to compute test instance features for AIRT, avoiding the additional computational expense, as well as the somewhat arbitrariness of certain feature choices. AIRT computes a 1-dimensional problem space based on dataset difficulty, which is calculated from the performance results of the algorithm portfolio.  Characteristics such as  algorithm consistency and anomalousness can be calculated as overall characteristics based only on an algorithm's performance metric, while the region of the problem space for which an algorithm shows superiority can be revealed without the need for features. The fact that similar insights can be obtained from the case studies presented in this paper without the need for feature calculation required by instance space analysis is one of the main advantages of AIRT focused on the broader goal of generating insights into algorithm performance, in addition to constructing strong algorithm portfolios, i.e. addressing both questions of which algorithm should be used for a particular instance, and why?

The remainder of the paper is organized as follows: In Section~\ref{sec:Method} we provide an introduction to polytomous and continuous IRT models and discuss the contextual differences between traditional applications that use IRT for evaluating educational outcomes and adaptations to evaluate algorithms. We then discuss our alternative adaptation, essentially an inverted model, which creates a rich new set of algorithm evaluation metrics defined by reframing the interpretation of the IRT parameters in Section~\ref{subsec:AlgoEvalMetrics}. Using these new metrics, we can visualize the strengths and weaknesses of algorithms in the problem space and construct algorithm portfolios using AIRT. Furthermore, to assess the goodness of the models built within our AIRT framework, we define additional measures based on model predicted performance and actual performance on test instances. AIRT expands on the IRT framework to including such enhancements to enable its application to the broader challenge of understanding algorithm strengths and weaknesses. In Section~\ref{sec:Results} we illustrate the complete functionality of AIRT -- including the algorithm metrics, problem space analysis, strengths and weaknesses of algorithms, algorithm portfolio evaluation and model goodness results -- using the detailed case study of  OpenML-Weka classification algorithms and test instances available at ASlib repository \citep{Bischl2016}. We refer the reader to  Appendix~\ref{sec:App1} where further results are summarized on nine more case studies using a variety of ASlib scenarios including from satisfiability (SAT)  and constraint satisfaction problem domains. These case studies demonstrate  the functionality of AIRT as an exploratory data analysis tool for algorithm portfolio evaluation and how the user can construct a competitive algorithm portfolio using AIRT with the objective of minimizing performance gap. Finally, we discuss future work and present our conclusions in Section~\ref{sec:conclusions}.

% =======================================================================
\section{IRT: Traditional setting and new mapping}\label{sec:Method}
% =======================================================================

% \subsection{A brief introduction to IRT }\label{subsec:IntroIRT}
Item Response Theory (IRT) \citep{lord1980applications,embretson2013item, van2013handbook}  refers to a family of latent trait models that is used to explain the relationship between unobservable characteristics such as intelligence or political preference and their observed outcomes such as responses to questionnaires. Attributes such as verbal or mathematical ability, racial prejudice and stress proneness, which  cannot be measured directly can be modeled as latent variables. The observed outcomes such as test items and questionnaire responses can be explained using latent trait models. IRT builds a connection between the items of a bigger unit such as a test with the participants' latent traits, thus placing each participant in a latent trait continuum. IRT is commonly used in psychometrics \citep{cooper2010psychometric} and educational testing \citep{YEN1986}.     

\subsection{Dichotomous and polytomous IRT models}\label{subsec:diandpoly}
 
We introduce some IRT concepts for dichotomous and polytomous models using the notation of \cite{JSSv048i06}  and \cite{JSSv017i05}. Let $i = 1, \ldots N$ represent participants or testees, $j= 1, \ldots n$ represent the test items with $N > n$, and let $\theta$ denote the latent variable such as intelligence or ability. An example includes a test with $n$ questions, which is administered to a class of $N$ students with the aim of measuring their ability $\theta$ to perform certain tasks. The response of the $i^{\text{th}}$ participant for the $j^{\text{th}}$ item is denoted by $x_{ij}$. The discrimination parameter for test item $j$ is denoted by $\alpha_j$ and the difficulty parameter by $d_j$. These two parameters are used to build the 2-Parameter Logistic (2PL) model, while  an additional guessing parameter $\gamma_j$ is incorporated in the 3-Parameter Logistic (3PL) model. 
 
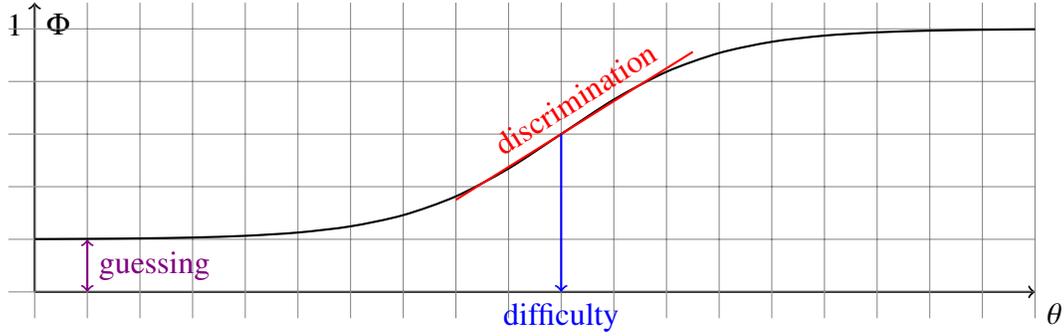
\begin{figure}[!ht]
     \centering
     \begin{tikzpicture}[scale=3.5]
        \draw[thick,->] (0,0) -- (3.8,0) node[anchor=north west] {$\theta$};
        \draw[thick,->] (0,0) -- (0,1.1) node[anchor=north east] {1} node[anchor=north west]{$\Phi$};
        \draw[step=0.2cm, gray,very thin] (-0.1,-0.1) grid (3.8,1.1);
        \draw [black, thick] plot [smooth] coordinates { ( 0.0, 0.2008829) (0.2,  0.2017423) (0.4, 0.2034346) ( 0.6, 0.2067567) (0.8, 0.2132384) (1.0, 0.2257366)  (1.2, 0.2492945) (1.4, 0.2918581) (1.6, 0.3631843) (1.8, 0.4688662) (2.0, 0.6000000) (2.2, 0.7311338) (2.4, 0.8368157)  (2.6, 0.9081419) (2.8, 0.9507055) (3.0, 0.9742634) (3.2, 0.9867616) (3.4, 0.9932433)  (3.6, 0.9965654) (3.8, 0.9982577)  };
        \draw[thick, red] (1.6, 0.35 ) -- (1.8, 0.475 ) -- (2,0.6) -- (2.2,0.725) node[midway, sloped,above]{discrimination};
         \draw[thick, red](2.2,0.725) -- (2.4,0.85) -- (2.5,0.9125); 
         \draw [thick, blue, ->](2, 0.6) -- (2,0) node[below]{difficulty};
         \draw [thick, violet, <->](0.2, 0) -- (0.2,0.2) node[midway, right]{guessing};
    \end{tikzpicture}
     \caption{Probability of a correct response for a given item using a 3PL model. Difficulty corresponds to $d_j$, discrimination to $\alpha_j$ and guessing to $\gamma_j$ in equation~\eqref{eq:3plbasic}. }
     \label{fig:IRT3PL}
 \end{figure}

 For dichotomous data researchers are interested in modeling the probability of correct response for each item given the ability level $\theta_i$. %SHOULD THIS BE SUBSCRIPTED FOR PARTICIPANT i?.  
 The 3PL model defines the probability of a correct response for participant $i$ for item $j$ as
 \begin{equation}\label{eq:3plbasic}
     \Phi\left( x_{ij} =1 |\theta_i, \alpha_j, d_j, \gamma_j \right) =  \gamma_j + \frac{1 - \gamma_j}{1 + \exp\left(-D \alpha_j \left( \theta_i - d_j \right) \right)}
 \end{equation}
 where $D$ is the scaling adjustment traditionally set at $1.702$. The role of $D$ is to make the logistic curve similar to the cumulative distribution function of the normal distribution \citep{reckase2009multidimensional}.  Figure~\ref{fig:IRT3PL} shows the resulting probability for a given item $j$ with fixed $\alpha, d$ and $\gamma$ disregarding the scaling constant $D$. The greater the ability $\theta_i$ of the participant, the higher the probability of the correct response. 
 
For polytomous data, we briefly present the multi-response ordinal models described in \cite{samejima1969estimation}. For example, self-esteem surveys have questions such as \textit{I feel that I am a person of worth, at least on an equal plane with others} with responses \{\textit{strongly disagree, disagree, neutral, agree, strongly agree}\}. In this case the original responses, which are the participants answers, are used to fit the IRT model \citep{Gray-Little1997}. By definition ordinal responses are ordered, i.e.,  \textit{strongly disagree} $<$ \textit{disagree} $<$ \textit{neutral} $<$ \textit{agree} $<$ \textit{strongly agree}. The responses need to be ordinal because the resulting latent trait continuum is ordered from low ability to high ability.  In educational testing an accuracy measure such as marks, derived from the original responses are used to fit the IRT model. For example, for each question in a test, the participants write their  answers and marks are derived by the person who grades them. For simplicity, suppose the marks for each question can take the values \{0, 1, 2, 3, 4, 5\}. The marks, which is a derived accuracy measure are the responses in this case and is used to fit the IRT model. Similarly, for multiple choice questions with marks taking the values \{0, 1\} a dichotomous IRT model is fitted. Whether a derived accuracy measure or the original responses are used, these are called \textit{responses} in IRT literature.   %An example of  polytomous ordinal data is the set of responses \{strongly disagree, disagree, neutral, agree, strongly agree\}.  Another example is obtaining marks $\{0,1,2\}$  for question $j$ with the assumption that half marks are not given. The input data to the IRT model may be derived from the original responses, such as marks, or may contain the responses themselves as in a course evaluation questionnaire with 5 options ranging from 1 to 5 where the level of satisfaction  increases from 1 to 5 for all questions. Whether derived or not, these are called \textit{responses} in IRT literature. 
We note that the word \textit{response} is confusing to non-IRT researchers when it refers to grades or other type of measures derived from the original responses. However, as this is the standard term used in IRT literature, we will use the same for easier cross-referencing.  
If there are $C_j$ unique response categories 

for item $j$ with $0 < 1 < \cdots < C_{j}-1  $, difficulty parameters $\bm{d}_j = \left(d_1, \ldots, d_{C_j-1} \right)$ and discrimination parameter $\alpha_j$,  the cumulative probabilities are defined as 
\begin{align}\label{eq:polytomous1}
    \Phi\left(x_{ij} \geq 0|\theta_i, \alpha_j, \bm{d}_j \right) & =  1 \, , \notag \\
     \Phi\left(x_{ij} \geq 1|\theta_i, \alpha_j, \bm{d}_j \right) & =  \frac{1}{1 + \exp\left(-D \alpha_j \left( \theta_i - d_1 \right) \right) } \, , \notag  \\
    \Phi\left(x_{ij} \geq 2|\theta_i, \alpha_j, \bm{d}_j \right) & =  \frac{1}{1 + \exp\left(-D \alpha_j \left( \theta_i - d_2 \right) \right) } \, ,  \notag \\
       \vdots \,  \notag \\
    \Phi\left(x_{ij} \geq C_{j}-1|\theta_i, \alpha_j, \bm{d}_j \right) & =  \frac{1}{1 + \exp\left(-D \alpha_j \left( \theta_i - d_{C_{j}-1} \right) \right) } \, , \\
     \Phi\left(x_{ij} \geq C_{j}|\theta_i, \alpha_j, \bm{d}_j \right) & =  0 \, , \notag 
\end{align}
where $x_{ij}$ is the response of participant $i$ for question $j$. This gives the probability of the response 
$x_{ij} = k $ as
\begin{equation}\label{eq:polytomous2}
     \Phi\left(x_{ij} = k|\theta_i, \alpha_j, \bm{d}_j \right)  =  \Phi\left(x_{ij} \geq k|\theta_i, \alpha_j, \bm{d}_j \right) - \Phi\left(x_{ij} \geq (k+1)|\theta_i, \alpha_j, \bm{d}_j \right) \, .
\end{equation}

\begin{figure}[!ht]
    \centering
    \includegraphics[trim={0.2cm 1cm 0 0cm},clip, scale=2.5]{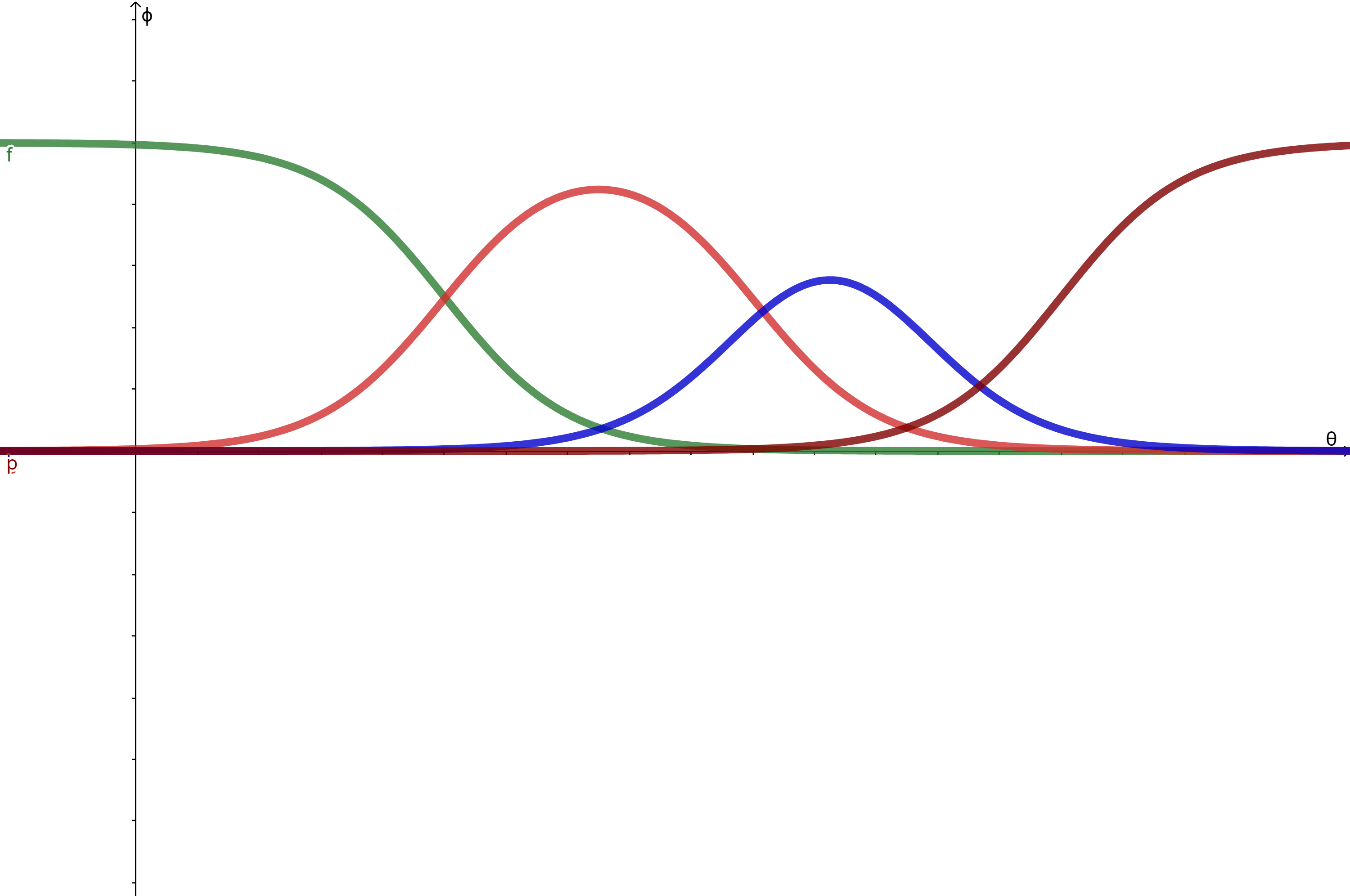}
    \caption{The probability of the response $x_{ij} = k$ for different $k \in \{0, 1, 2, 3 \}$ with $\theta$ on the horizontal axis and $\Phi$ on the vertical axis. The most likely outcome is different depending on the ability levels $\theta$. }
    \label{fig:polytomousIRT}
\end{figure}
Figure~\ref{fig:polytomousIRT} shows the probability density functions for different responses $x_{ij}=k$ for $k \in \{0, \ldots, (C_j-1)\}$.  In the educational testing scenario discussed above, each curve denotes the probability that marks are equal to $k$ for $k \in \{0, 1, 2, 3\}$. From Figure~\ref{fig:polytomousIRT} we see that the green curve, which gives the probability density function for marks = 0, has high probability when the participant ability $\theta$ is low. Similarly, the dark red curve corresponding to marks = 3 has a higher probability for high participant ability. We see that a participant with a lower ability/latent trait is more likely to obtain a response corresponding to  a low value of $k$ compared to a participant with a higher ability. 

\subsection{Continuous IRT models}\label{subsec:ctsIRTmodels}
In addition to the polytomous IRT models, \cite{Samejima1973, Samejima1974} introduced Continuous Response Models (CRM) to extend polytomous models to continuous responses. \cite{Wang1998} introduced an expectation-maximization (EM) algorithm for Samejima's continuous item response model. This EM algorithm was further optimized by \cite{SHOJIMA2005} by proposing a non-iterative solution for each EM cycle. 

% Section 2.2 Joint maximum likelihood estimation -Shijoma
In this section we use the notation used by \cite{Wang1998} and \cite{SHOJIMA2005}. They consider $N$ examinees with trait variables $\theta_i$ where $i \in \{1, \ldots, N\}$ and $n$ test instances with parameters $\bm{\lambda}_j = (\alpha_j, \beta_j, \gamma_j)^T$ for $j \in \{1, \ldots, n\}$. The item parameters $\alpha_j$ represents discrimination, $\beta_j$  difficulty  and $\gamma_j$ a scaling coefficient that defines a scaling transformation from the original rating scale to the $\theta$ scale.

% Wang and Zeng equation 1 and 2
Using the normal density type CRM, \cite{Wang1998} considered the probability of an examinee with an ability $\theta$ obtaining a score of $y_j$ or higher on a given item $j$ as
\begin{equation}\label{eq:CRM1}
P\left( Y \geq y_j |\theta \right) = \frac{1}{\sqrt{2\pi}} \int_{-\infty}^{v} e^{-\frac{t^2}{2}} dt \, , 
\end{equation}
where
\begin{equation}\label{eq:CRM2}
    v = \alpha_j\left(\theta - \beta_j - \gamma_j \ln \frac{y_j}{k_j - y_j} \right)\, , 
\end{equation}
and the continuous score range of $y_j$ is $(0, k_j)$. The continuous score range of $(0, k_j)$  is opened up to $(-\infty, \infty)$ with the reparametrization  
\begin{equation}\label{eq:CRM3}
    z_j = \ln \frac{y_j}{k_j - y_j} \, .
\end{equation}
Using this reparametrization they obtain the probability density function $f\left(z_j|\theta \right)$ by differentiating the cumulative density function obtained using equation~\eqref{eq:CRM1} as 
% Wang and Zeng eq 5
\begin{equation}\label{eq:CRM4}
    f\left(z_j|\theta \right) = \frac{d}{dz_j} \left( 1- P(Z \geq z_j | \theta) \right)  = \frac{\alpha_j\gamma_j}{\sqrt{2\pi}} \exp \left( -\frac{\alpha_j^2}{2} \left( \theta - \beta_j - \gamma_j z_j \right)^2 \right) \, .
\end{equation}

Comparing the parameters $\alpha_j$, $\beta_j$ and $\gamma_j$ with those of Section~\ref{subsec:diandpoly} we note that the parameter $\alpha_j$ denotes discrimination as in Section~\ref{subsec:diandpoly} and the parameter $\beta_j$ denotes the difficulty level of item $j$, which was denoted by $d_j$ in Section~\ref{subsec:diandpoly}. However, the parameter $\gamma_j$ is quite different to the guessing parameter used in  Section~\ref{subsec:diandpoly}, in that it denotes a scaling factor which we will inspect soon. 

\begin{figure}[!ht]
    \centering
    \includegraphics[scale=0.9]{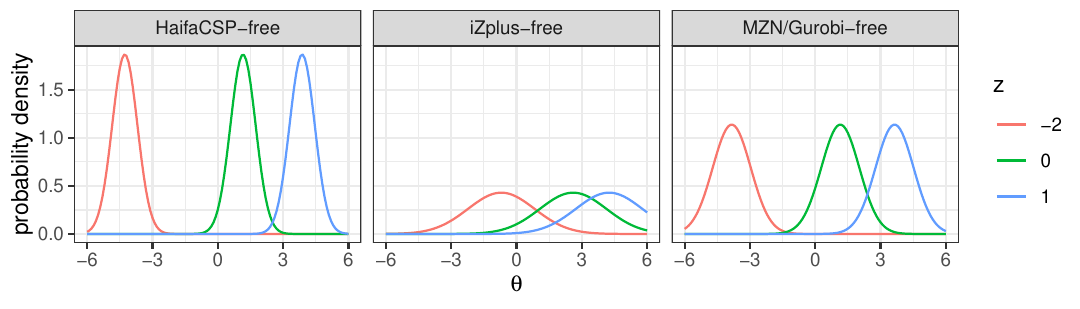}
    \caption{  Probability density curves for $z=-2$, $z=0$ and $z=1$ for three items with different CRM parameters. The items are from CSP-Minizinc-2016 algorithm portfolio. }
    \label{fig:crmparameters}
\end{figure}

\begin{figure}[!ht]
    \centering
    \includegraphics[scale=0.8]{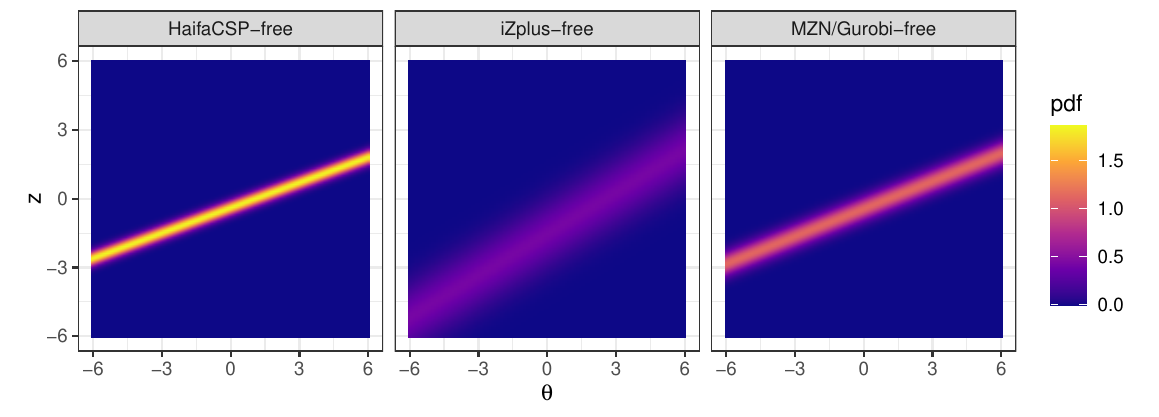}
    \caption{  The heatmap of probability density functions for the items in Figure \ref{fig:crmparameters}}
    \label{fig:crmparameters2}
\end{figure}

For every $z \in \mathbb{R}$, there is an associated probability density function given by $f(z|\theta)$. Figure~\ref{fig:crmparameters} shows the item response functions obtained for $z \in \{-2, 0, 1 \}$ for different items, which have different CRM parameters. Figure~\ref{fig:crmparameters2} shows the heatmap of $f(z|\theta)$ for the same items for continuous $z$ and $\theta$ values.  The first pane in both Figures show the curves/heatmap for the first item, HaifaCSP-free,  with  $\alpha=1.73$, $\beta = 1.16$ and $\gamma =2.72$. The second item, iZplus-free, has CRM parameters $\alpha= 0.65$, $\beta = 2.6$ and $\gamma = 1.65$. The third item, MZN/Gurobi-free, has CRM parameters $\alpha=1.14$, $\beta=1.15$ and $\gamma=2.49$. We will give more context on these items later.   The second item has a higher difficulty level compared to the first and the third we see that $\beta=2.6$ shifts the curves to the right in Figure~\ref{fig:crmparameters} and the high density regions have moved to the right in Figure~\ref{fig:crmparameters2}. The first item has higher $\alpha$ values making the curves steeper in Figure~\ref{fig:crmparameters} compared with items 2 and 3. Similarly, the high density regions are narrower and sharper in  Figure~\ref{fig:crmparameters2} due to higher discrimination. %The last pane shows curves/heatmap for an item with somewhat unexplored characteristics. Item 4 has negative discrimination, i.e. $\alpha= -1$. As the probability density $f\left(z_j|\theta \right)$ in equation~\eqref{eq:CRM4} needs to be positive, this is possible with  negative $\gamma$ values. For an item with negative discrimination examinees with high ability obtain low scores while examinees with low ability obtain high scores. We see that the order of the curves are reversed in Figure~\ref{fig:crmparameters} and similarly the high density region has a negative slope in Figure~\ref{fig:crmparameters2}.

\cite{Wang1998} estimated the item parameters used in equation~\eqref{eq:CRM4} using an EM algorithm. \cite{SHOJIMA2005} enhanced the algorithm by proposing a non-iterative step for the expectation cycle, which made the item parameter computation much faster. However, in their estimation \cite{SHOJIMA2005} only considers $\alpha, \gamma > 0$. As such, their algorithm does not accommodate negative discrimination items.  This reflects the current practice regarding negative discrimination items in educational and psychometric testing. Negative discrimination items are generally considered as non-value adding and as such revised or removed in traditional educational testing \citep{hambleton2013item}. However, in algorithm performance negative discrimination plays an important role and we do not remove such items from the pool. 

We accommodate negative discrimination items by modifying the existing algorithm discussed by \cite{SHOJIMA2005}. Before discussing these modifications we give a brief overview of their method. First they  rescale $y_{ij}$, such that $x_{ij} = y_{ij}/k_j$ lies in $(0,1)$ and consider $z_{ij} = \ln x_{ij}/(1-x_{ij})$.  They denote the item response vector of examinee $i$ by $\bm{z}_i$. % Section 2.2 the joint maximum likelihood estimation
Then they perform a marginal maximum likelihood estimation with the expectation maximization algorithm (MML-EM). Using a normal prior for $\theta_i$, i.e. $\mathscr{N}\left(\theta_i |\mu, \sigma \right)$ % Section 2.3 marginal ML
they obtain an estimate for the posterior distribution of $\theta_i$, given $\bm{z}_i$ and the current estimates of item parameters as
\begin{equation}\label{eq:CRM5}
    p\left( \theta_i|\bm{\Lambda}^{(t)}, \bm{z}_i \right) = \mathscr{N}\left(\theta_i \vert \mu_i^{(t)}, \sigma^{(t)2} \right) \, , 
\end{equation}
where $\bm{\Lambda}^{(t)} = \left(\bm{\lambda_1}^{(t)}, \ldots, \bm{\lambda_n}^{(t)} \right)$, $\bm{\lambda_j}^{(t)} = \left(\alpha_j^{(t)}, \beta_j^{(t)}, \gamma_j^{(t)} \right)^T$ and ${(t)}$ denotes the iteration. The parameters $\mu_i^{(t)}$ and $\sigma^{(t)}$  are given by % EQ 19 and 20 in Shijoma
\begin{align}\label{eq:CRM6}
    \sigma^{(t)2} & = \left( \sum_j \alpha_j^{(t)2} + \sigma^{-2}  \right)^{-1} \, , \\
    \mu_i^{(t)} & =  \sigma^{(t)2} \left( \sum_j \alpha_j^{(t)2} \left( \beta_j^{(t)}  +  \gamma_j^{(t)} z_{ij} \right) + \mu   \right)\, ,
\end{align}
where $\mu$ and $\sigma$ denote the initial prior parameters of $\theta_i$. Then they obtain the expectation of the log-likelihood
\begin{equation}\label{eq:CRM7}
E_{\bm{\theta}|\Lambda^{(t)}, \bm{Z}}\left[ \ln p\left( \bm{\Lambda} \vert \bm{\theta}, \bm{Z} \right) \right] = N \sum_{j=1}^n \left(\ln \alpha_j  + \ln\gamma_j \right) - \frac{1}{2} \sum_{i=1}^N \sum_{j=1}^n \alpha_j^2 \left( \left(  \beta_j + \gamma_j z_{ij} - \mu_i^{(t)} \right)^2 + \sigma^{(t)2} \right) + \ln p\left(\bm\Lambda \right) + \text{const}
\end{equation} 
where $\bm{Z}$ is the item response matrix of all examinees for $n$ items and $p$ denotes the probability. They optimize this expectation with flat priors for item parameters and obtain
\begin{align}
    \gamma_j^{(t+1)} & = \frac{V \left(\mu_i^{(t)}\right) + \sigma^{(t)2}}{C_j\left(z_{ij}, \mu_i^{(t)} \right)}  \, , \label{eq:CRM8.1}\\
    \beta_j^{(t+1)} & = M\left( \mu_i^{(t)}\right) - \gamma_j^{(t+1)} M_j \left( z_{ij}\right) \, , \label{eq:CRM8.2} \\
    \alpha_j^{(t+1)} & = \left( \gamma_j^{(t+1)2} V_j(z_{ij}) - V \left(\mu_i^{(t)}\right) - \sigma^{(t)2} \right)^{-1/2} \, ,  \label{eq:CRM8.3}
\end{align}
where $M$, $V$ and $C$ denote the mean, variance and covariance terms defined by
\begin{align}\label{eq:CRM9}
     M_j \left(z_{ij}\right) & = \frac{\sum_i z_{ij}}{N} \, , \notag \\
     M\left( \mu_i^{(t)} \right) & = \frac{\sum_i \mu_i^{(t)}}{N} \, , \notag \\
     V\left( z_{ij}\right) & = \frac{\sum_i z_{ij}^2}{N} - M_j\left( z_{ij}\right)^2 \, , \notag \\
      V\left( \mu_i^{(t)} \right) & = \frac{\sum_i \mu_i^{(t)2}}{N} - M\left( \mu_i^{(t)} \right)^2 \, , \notag \\
\text{and} \qquad      C_j\left( z_{ij}, \, \mu_i^{(t)} \right) & =  \frac{\sum_i z_{ij} \mu_i^{(t)} }{N}  -   M_j \left(z_{ij}\right) M\left( \mu_i^{(t)} \right) \, .  
\end{align}
% Eq 26, 27, 28 and the next one in Shojima
For each iteration using $\mu_i^{(t)}$, $\sigma^{(t)}$, and $M$, $V$ and $C$ quantities listed above, the parameter $\gamma_j^{(t+1)}$ is computed as in equation~\eqref{eq:CRM8.1}. This value of $\gamma_j^{(t+1)}$ is used to compute $\beta_j^{(t+1)}$ and $\alpha_j^{(t+1)}$ in equations~\eqref{eq:CRM8.2} and~\eqref{eq:CRM8.3}. Using the parameter values $\alpha_j^{(t+1)}$, $\beta_j^{(t+1)}$ and $\gamma_j^{(t+1)}$,  the log-likelihood given in equation~\eqref{eq:CRM7} is computed. This whole process is repeated until the difference in log-likelihoods for successive iterations becomes smaller than a predefined level of convergence. We note that this is a brief overview of this method and refer to \cite{SHOJIMA2005} for more details. 

%\subsubsection{Accommodating negative discrimination items in continuous IRT}
With the current formulation we see that if $\gamma_j^{(t+1)}$ in equation~\eqref{eq:CRM8.1} is negative due to a negative covariance term $C_j\left(z_{ij}, \mu_i^{(t)} \right)$ computed as in equation~\eqref{eq:CRM9}, this results in the log-likelihood in equation~\eqref{eq:CRM7} being incalculable as it requires $\ln \gamma_j$. This forces the MML-EM algorithm to stop, preventing convergence. As a result, this formulation only works when all test instances have  $\alpha_j >0$ and $\gamma_j>0$ as permitted by the assumption.

However, we see that the probability density function $f(z_j|\theta)$ in equation~\eqref{eq:CRM4} contains the product $\alpha_j \gamma_j$ and is valid when both $\alpha_j$ and $\gamma_j$ have the same sign. Similarly, equation~\eqref{eq:CRM7} can be rewritten with the product $\ln \left(\alpha_j \gamma_j \right)$ instead of the sum of log terms and is valid when both  $\alpha_j$ and $\gamma_j$ have the same sign. 

Therefore, if we remove the assumption used by \cite{SHOJIMA2005}, that  $\alpha_j >0$ and $\gamma_j>0$ and update it with $\alpha_j\gamma_j>0$,  we incorporate test items with  $\alpha_j, \, \gamma_j<0$ as well as test items with  $\alpha_j, \, \gamma_j>0$. That is, effectively we are adding the assumption  $\sign(\alpha_j) = \sign(\gamma_j)$, instead of $\alpha_j >0$ and $\gamma_j>0$. More importantly, we are opening the IRT model to negative discrimination items. 

With the updated assumption we can rewrite the log-likelihood as
\begin{equation}\label{eq:CRM12}
E_{\bm{\theta}|\Lambda^{(t)}, \bm{Z}}\left[ \ln p\left( \bm{\Lambda} \vert \bm{\theta}, \bm{Z} \right) \right] = N \sum_{j=1}^n \left(\ln |\alpha_j|  + \ln|\gamma_j| \right) - \frac{1}{2} \sum_{i=1}^N \sum_{j=1}^n \alpha_j^2 \left( \left(  \beta_j + \gamma_j z_{ij} - \mu_i^{(t)} \right)^2 + \sigma^{(t)2} \right) + \ln p\left(\bm\Lambda \right) + \text{const} \, ,
\end{equation} 
making the log-likelihood tractable for any $\alpha_j$ and $\gamma_j$. Then following through the computation we obtain
\begin{equation}\label{eq:CRM11}
      \alpha_j^{(t+1)}  =  \sign\left(\gamma_j^{(t+1)}\right)\left( \gamma_j^{(t+1)2} V_j(z_{ij}) - V \left(\mu_i^{(t)}\right) - \sigma^{(t)2} \right)^{-1/2} \, .  
\end{equation}
The parameters $\gamma_j^{(t+1)}$ and $\beta_j^{(t+1)}$ stay the same as given by equations~\eqref{eq:CRM8.2} and~\eqref{eq:CRM8.3} with the updated assumption. These modifications allow us to fit both negative and positive discrimination items in our continuous IRT model.

\begin{figure}[!ht]
    \centering
    \begin{tikzpicture}[
roundnode/.style={rectangle, draw=black!60, very thick, text width=2cm},
]
    \node[roundnode](participant){ Participant $i$};
    \node[roundnode](testquestion)[right=of participant]{Question $j$};
    \node[roundnode](marks)[below=of participant, xshift=1.65cm]{Marks $x_{ij}$};
    \draw[ultra thick, ->] (participant) -- (marks);
    \draw[ultra thick, ->] (testquestion) -- (marks);
    \node[roundnode](ability)[right=of testquestion]{Participant Ability $\theta_i$};
    \node[roundnode](discrimination)[right=of ability]{Question Discrimination $\alpha_j$};
    \node[roundnode](difficulty)[right=of discrimination]{Question Difficulty $\beta_j$};
    \node[roundnode](marks2)[below=of discrimination]{Marks $x_{ij}$};
    \draw[ultra thick, ->] (ability) -- (marks2);
    \draw[ultra thick, ->] (discrimination) -- (marks2);
    \draw[ultra thick, ->] (difficulty) -- (marks2);
    \end{tikzpicture}
    
    \caption{\color{black} Left: A DAG showing  participant $i$ and question $j$ giving rise to marks $x_{ij}$. Right: The DAG composed of participant and question attributes.}
    \label{fig:dagtraditional}
\end{figure}
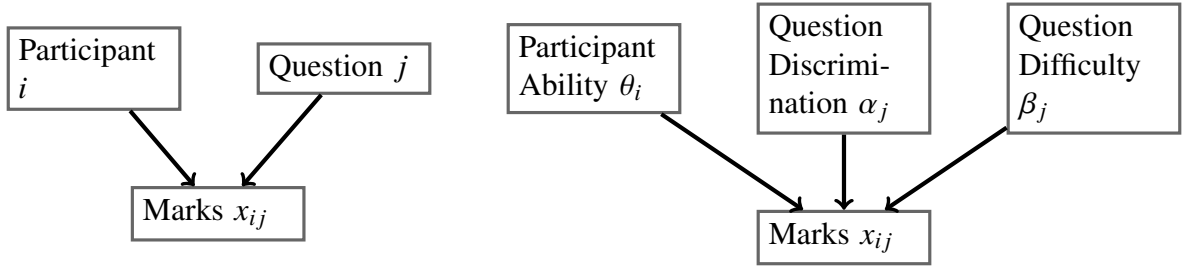

{\color{black}
The causal interpretation of traditional IRT presumes that the attributes of participant $i$ and test question $j$ give rise to marks $x_{ij}$. The attributes are the  discrimination and difficulty parameters of question $j$ and the ability of the participant $i$. This is shown in the Directed Acyclic Graph (DAG) in Figure \ref{fig:dagtraditional}. While traditional IRT texts do not include DAGs, more recent work \citep{Kelly2023} makes these causal interpretations explicit. 
}

% =======================================================================
\subsection{Applications to machine learning and algorithm evaluation}\label{subsec:AlgoEval} % Mapping IRT to the challenge of algorithm evaluation

 In the traditional IRT setting $N$ participants' responses for $n$ test instances are used to fit an IRT model and obtain the discrimination and difficulty of test instances as well as the ability of the participants.  A natural way to use the IRT framework on algorithms and test instances is to consider an algorithm as a participant and  test instances as test questions/items. If we formulate our problem this way, then we can obtain the test instance characteristics difficulty and discrimination using the IRT framework. In addition, IRT will also give us the latent scores or the ability of the algorithms.  \cite{martinez2019item} and \cite{chen2019beta} formulated their problem this way and used the IRT framework to evaluate observations in a dataset and obtain the ability of the classifiers for that dataset. 
 
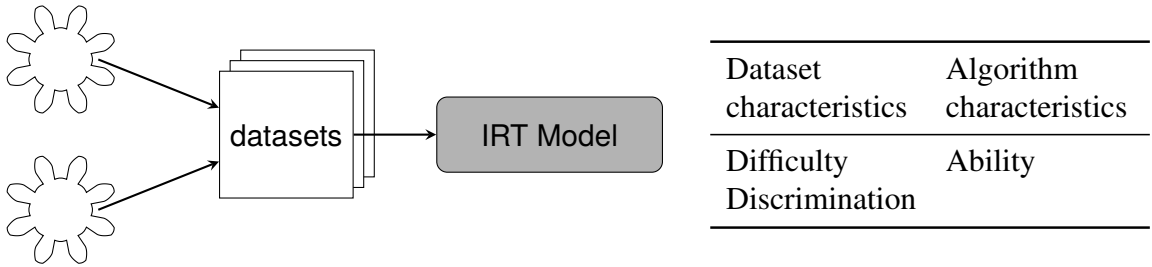
\begin{figure}
\centering
\begin{minipage}{.6\textwidth}
\begin{tikzpicture}[font=\sffamily,
doc/.style={draw, minimum height=4em, minimum width=2em, 
            fill=white, double copy shadow={shadow xshift=4pt, shadow yshift=4pt, fill=white, draw}},]
    \node[doc] at (3,-1) (doc) {datasets};                
    \pic(g1)[draw] at (0,0)   {gear={0.15}{8}{14}};
    \pic(g2)[draw] at (0,-2)   {gear={0.15}{8}{14}};
    \node (model) [startstop, right of=doc, xshift=2.5cm] {IRT Model};
    \draw [arrow] (0.5, 0) -- (doc);
    \draw [arrow] (0.5, -2) -- (doc);
    \draw [arrow] (doc) -- (model);
    \end{tikzpicture}
   
\end{minipage}%   
\begin{minipage}{.4\textwidth}
\begin{tabular}{p{2.5cm}p{2.5cm}}
\toprule
	Dataset \par characteristics
  & Algorithm \par  characteristics \\ \midrule
 Difficulty & Ability \\
 Discrimination &  \\
 \bottomrule
\end{tabular}
\end{minipage}
\caption{Standard IRT setting extended to algorithms working on datasets. The IRT model provides the dataset characteristics of difficulty and discrimination, and algorithm ability, as outputs.}
\label{fig:tradIRT}
\end{figure}

 Instead of using observations of a given  dataset as test items, we can also use datasets as test items. Then the parameters fitted by the IRT model would be dataset difficulty and discrimination. This is illustrated in Figure~\ref{fig:tradIRT}. Recent investigations \citep{irtensemble} showed the benefits of a flipped approach in constructing an unsupervised anomaly detection ensemble for a single dataset where observations were used as participants and algorithms as test items. In the current paper, we  explore this idea further for evaluating algorithms on many datasets, developing a full theory and framework for comprehensive algorithm evaluation.

% \todo[inline]{Uncomment Figure fig:tradIRT - commented because it takes long to compile}

% [
% roundnode/.style={circle, draw=black!60, very thick, minimum size=22mm, },
% ]

\section{Algorithmic IRT (AIRT)}\label{subsec:AlgoEvalMetrics} % Analysis
\begin{figure}[!ht]
\centering
\begin{minipage}{.6\textwidth}
\begin{tikzpicture}[font=\sffamily,
doc/.style={draw, minimum height=4em, minimum width=3em, 
            fill=white, double copy shadow={shadow xshift=4pt, shadow yshift=4pt, fill=white, draw}},]
    \node[doc] at (0,0) (doc1) {dataset};    
    \node[doc] at (0,-2) (doc2) {dataset};    
    \pic(g1)[draw] at (2.9,-1.9)   {gear={0.2}{8}{16}};
    \pic(g2)[draw] at (3.25,-0.25)   {gear={0.2}{8}{16}};
    \node (model) [startstop, right of=doc2, xshift=5.5cm, yshift = 1cm] {IRT Model};
    \draw [arrow] (doc1) -- (2.1,-0.8);
    \draw [arrow] (doc2) -- (2.1,-1.2);
    \draw [arrow] (4,-1) -- (model);
    \end{tikzpicture}
\end{minipage}%   
\begin{minipage}{.4\textwidth}
\begin{tabular}{p{2.5cm}p{2.5cm}}
\toprule
	Algorithm \par characteristics
  & Dataset  \par  characteristics \\ \midrule
 Algo-Char-1 & Dataset-Char-1 \\
 Algo-Char-2 &  \\
 \bottomrule
\end{tabular}
\end{minipage}
\caption{Inverted IRT setting with datasets acting on algorithms. The IRT model provides two algorithm characteristics in place of difficulty and discrimination, and one dataset characteristic in place of ability as outputs.}
 \label{fig:invertedIRT}
\end{figure}
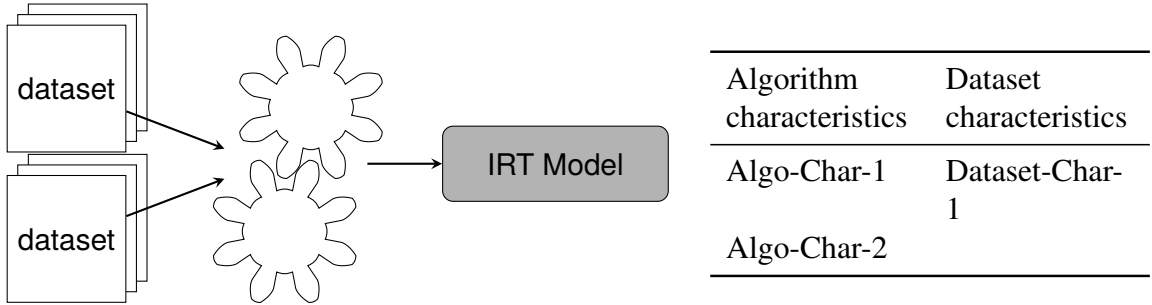

\begin{figure}[!ht]
    \centering
    \begin{tikzpicture}[
roundnode/.style={rectangle, draw=black!60, very thick,text width=2cm},
]
    \node[roundnode](participant){Dataset $i$};
    \node[roundnode](testquestion)[right=of participant]{Algorithm $j$};
    \node[roundnode](marks)[below=of participant, xshift=1.65cm]{Performance $x_{ij}$};
    \draw[ultra thick, ->] (participant) -- (marks);
    \draw[ultra thick, ->] (testquestion) -- (marks);
    \node[roundnode](ability)[right=of testquestion]{Dataset Ability $\theta_i$};
    \node[roundnode](discrimination)[right=of ability]{Algorithm Discrimination $\alpha_j$};
    \node[roundnode](difficulty)[right=of discrimination]{Algorithm Difficulty $\beta_j$};
    \node[roundnode](marks2)[below=of discrimination]{Performance $x_{ij}$};
    \draw[ultra thick, ->] (ability) -- (marks2);
    \draw[ultra thick, ->] (discrimination) -- (marks2);
    \draw[ultra thick, ->] (difficulty) -- (marks2);
    \end{tikzpicture}
    \caption{\color{black} Mapping IRT to the algorithm evaluation domain, a  participant is mapped to a dataset and a question is mapped to an algorithm. Left: The resulting DAG from the this mapping. Right: The DAG composed of dataset and algorithm attributes.}
    \label{fig:dagalgorithmic}
\end{figure}
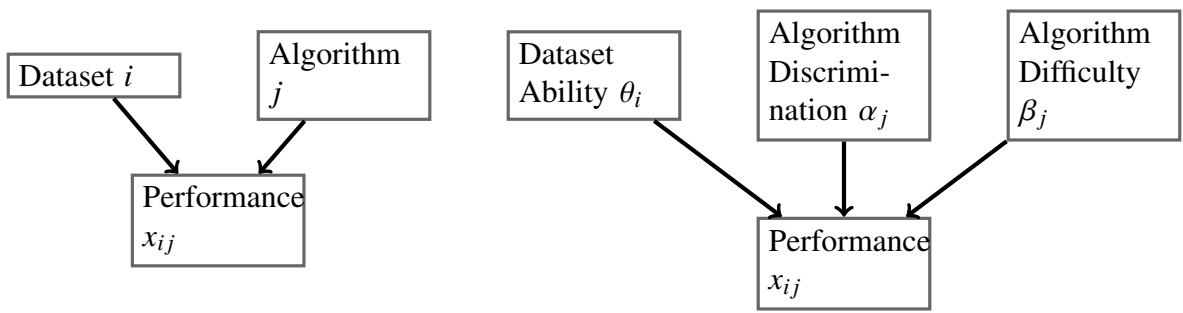

As a novel adaptation in this paper we now invert the intuitive IRT mapping discussed in the previous paragraph and consider algorithms as items and test instances as participants. This is shown in Figure~\ref{fig:invertedIRT}. This inversion results in a loss of intuition momentarily. However, by persisting with this less intuitive mapping we gain an elegant reinterpretation of the theory that enables us to analyze the strengths and weaknesses of algorithms with far more nuanced detail. Firstly, we note that this inversion produces two parameters describing algorithm properties compared to a single parameter in the standard setting. As we will see shortly, we will derive three algorithm characteristics from these two algorithm parameters. Thus, the inversion serves to offer a richer set of metrics with which to evaluate algorithms, compared to the standard approach, which focuses more on dataset/observation evaluation. Table~\ref{tab:differentIRTframework} compares the classic IRT approach with the standard and the inverted IRT approaches for algorithm evaluation.  {\color{black}
With this mapping, we presume that attributes of the problem/dataset $i$ and algorithm $j$ give rise to the performance $x_{ij}$ as shown in the DAG in Figure \ref{fig:dagalgorithmic}. 
}
% \todo[inline]{Uncomment Figure fig:invertedIRT}

\begin{table}[!ht]
		\centering
		\caption{A comparison between the classic IRT with the standard and inverted IRT approaches for algorithm evaluation. }
		\begin{tabular}{p{1.5cm}|p{3cm}|p{3cm}|p{3cm}|p{3cm}}
			\toprule %\cmidrule{2-5}
		 &	Classic IRT & Standard Approach for Algorithm \par Evaluation & Inverted Approach for Algorithm Evaluation & Inverted \par Characteristics  \\
			\midrule
       Setting    & Examinees doing test items & Algorithms working on datasets & Datasets acting on algorithms & \\ 
           \hline 
          & Test item difficulty & Dataset  difficulty & Difficulty parameter for algorithms & Algorithm difficulty limit \\ \cdashline{2-5}
      Parameters    & Test item discrimination & Dataset  discrimination &  Discrimination parameter for algorithms & Algorithm anomalousness and consistency\\ \cdashline{2-5}
          & Examinee ability & Algorithm ability & Ability trait of datasets & Dataset  difficulty \\
		    \bottomrule
		\end{tabular}
		\label{tab:differentIRTframework}
\end{table}

\begin{figure}[!ht]
  \centering
  \includegraphics[scale=0.8]{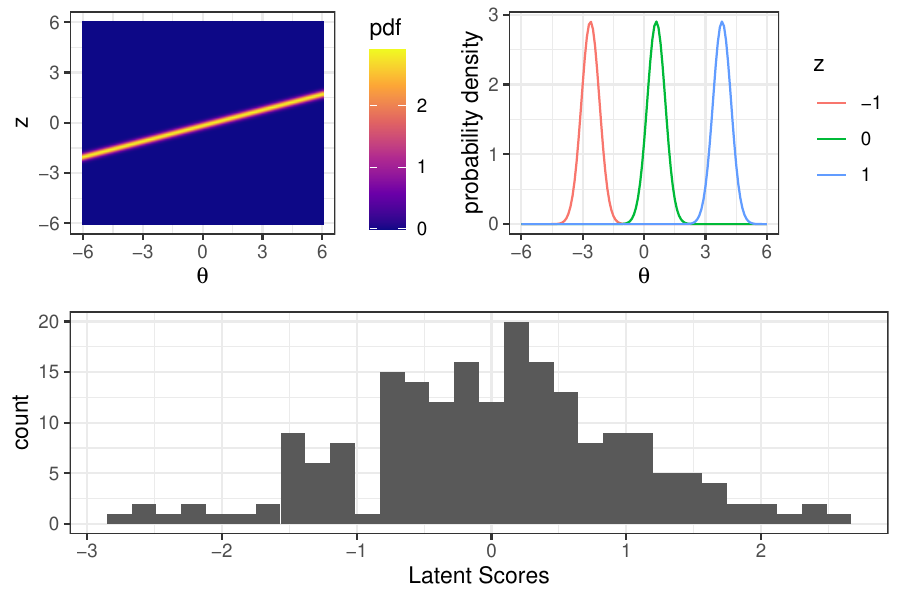}
  \caption{  
  The heatmap of LCG-Glucose-free on the top left and the trace lines for $z \in \{-1, 0, 1 \}$ for that item on the top right. The histogram of the latent scores estimated by the model is shown at the bottom. In the inverted IRT algorithm evaluation setting, the latent scores represent test instance easiness. }
  \label{fig:algoPerfHistScores}
\end{figure}

The inherent meaning of resulting IRT parameters and latent scores is changed when we map algorithms to items and test instances to participants. For example, suppose Figure~\ref{fig:algoPerfHistScores} originates from an educational testing scenario. It shows the heatmap of a test question, the set of trace lines with $\text{P}1 < \text{P}2 < \text{P}3 < \text{P}4 $ and a histogram of latent scores. The $y$-axis in the heatmap labeled $z$ denotes the normalized score and examinee's ability is denoted by $\theta$. Then, as the examinee's ability increases, the probability of getting a better grade for this particular question also increases as seen from the heatmap and the trace lines.  For algorithm evaluation, let us also consider the performance levels $\text{P}1 < \text{P}2 < \text{P}3 < \text{P}4 $ with higher levels and larger $z$ values indicating better performance. If we consider the standard IRT approach discussed in Table~\ref{tab:differentIRTframework}, then the heatmap and the trace lines give the performance of a specific dataset and the histogram of latent scores give algorithm abilities. If we consider the inverted IRT approach, the heatmap and the tracelines show the performance of an algorithm and the histogram gives the latent scores of the datasets. What do these latent scores represent? We know that algorithms gives better performance on easy test instances. For example a classification algorithm such as logistic regression will give better classification accuracy on a linearly separable dataset compared to a complex dataset. As such, in the inverted algorithm evaluation setting the latent score $\theta$ represents the easiness of the test instance. 

Furthermore, this inverted setting gives rise to important algorithm characteristics that can now be measured using the IRT parameters, as described in the following sections.

% =======================================================================
{ \subsection{Framework} }
Our algorithmic IRT (AIRT) framework consists of three main stages:
\begin{enumerate}
    \item Stage 1: Fitting an IRT model with inverted mapping \\
    We input the performance results of $n$ algorithms on $N$ test instances to a continuous or a polytomous IRT model,  mapping test instances to participants and algorithms to items. The R package \texttt{airt} fits the continuous IRT models described in Section~\ref{subsec:ctsIRTmodels} using the updated log-likelihood function and assumption. To fit polytomous models \texttt{airt} uses the functionality of the existing R package \texttt{mirt} \citep{JSSv048i06}.
    \item Stage 2: Calculation of algorithm and dataset metrics  \\
    The second stage consists of reinterpreting the results  of the IRT model, due to the inverted mapping and inherent contextual differences, so that a richer set of metrics for algorithm performance and dataset difficulty can be calculated.
    \item Stage 3: Compute strengths and weaknesses and construct algorithm portfolios \\
    Construct latent trait curves to enable algorithm ranking and  strengths and weaknesses of algorithm portfolios to be observed across test suites of varying difficulty. 
\end{enumerate}

A range of indicators are computed as additional measures that characterize algorithms and assess the goodness of the IRT model, as presented in the following sections. AIRT is applicable to both continuous and polytomous IRT models. Our results on various algorithm portfolios used to validate the approach in Section~\ref{sec:Results}  and Appendix~\ref{sec:App1}  focus on continuous IRT models, however we note that AIRT can be used to construct polytomous models. We present results for continuous scenarios because they have higher variation  and as such are more interesting. We note that the R package \texttt{airt} has the functionality to handle polytomous data as well as continuous, and details of the generalization to polytomous data are provided in Supplementary Materials. 

We will use CSP-Minizinc-2016 algorithm portfolio from ASlib repository \citep{Bischl2016} to illustrate algorithm and dataset metrics. For all algorithms in the ASlib repository certain hyperparameters and parameters were used which we do not vary.  Any conclusions we draw about algorithm performance are therefore dependent on the actual algorithm implementation they use. Further conclusions about the strengths and weaknesses of any algorithm would need to thoroughly explore the impact of its parameter values. 

\begin{figure}[!ht]
    \centering
    \includegraphics{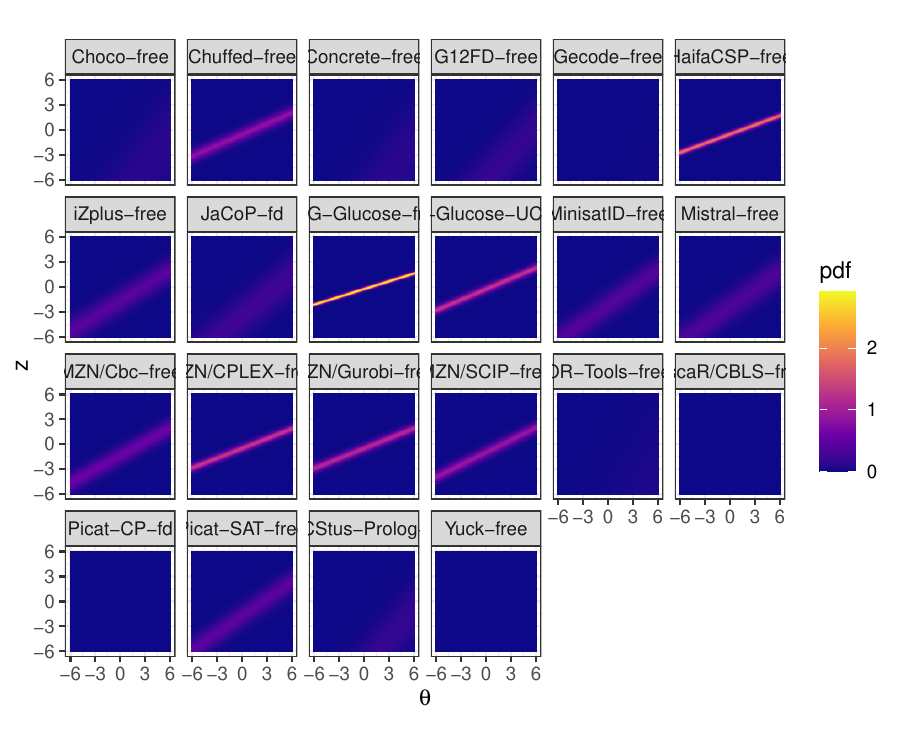}
    \caption{ The heatmap of probability density functions for all algorithms in CSP-Minizinc-2016 portfolio.}
    \label{fig:crmparameters22}
\end{figure}

CSP-Minizinc-2016  contains the results of constrained satisfaction and optimization problems. The original dataset contains the runtimes of each problem instance. As the IRT framework denotes good performance by increasing values we have taken the reciprocal of the runtimes to fit the AIRT model. Figure \ref{fig:crmparameters22} shows the heatmaps of the probability density functions for all algorithms in the portfolio. The items discussed in Figures \ref{fig:crmparameters} and \ref{fig:crmparameters2} were algorithms taken from this portfolio.

\subsection{ Dataset metric: Difficulty score}\label{latenttrait} 
As discussed previously, the latent trait denoted by $\theta$ corresponds to dataset easiness and is given by \begin{equation}\label{eq:CRM15}
     \theta_i  = \frac{\sum_j \hat{\alpha}_j^2 \left(\hat{\beta}_j + \hat{\gamma}_j z_{ij} \right)}{  \sum_j \hat{\alpha}_j^2} \, , 
\end{equation} 
where $\hat{\alpha}_j$, $\hat{\beta}_j$ and $\hat{\gamma}_j$ are the estimated discrimination, difficulty and scaling parameters for algorithm $j$, which are obtained by fitting the IRT model. Using $\theta_i$ we define dataset difficulty as
\begin{equation}\label{eq:CRMdifficultydat}
    \delta_i = -\theta_i \, ,
\end{equation}
where $ \delta_i$ denotes the difficulty of the $i^{\text{th}}$ dataset. We see that dataset difficulty is a function of discrimination, difficulty and scaling parameters of algorithms as well as the accuracy scores of the datasets. 

\cite{SHOJIMA2005} uses the normal density type CRM with normal priors making the posterior distribution of the trait parameter $\theta$  normal. Thus, we can expect dataset difficulty $\delta$ to be normally distributed. We refer to datasets/problems as easy if they have low difficulty values. Similarly, we say datasets/problems are difficult if they have high difficulty values. The semi-difficult or semi-easy instances are in the middle of the spectrum. 

%The dataset difficulty spectrum gives a way of ordering the performance values $y_{ij}$. For each algorithm $j$, we can consider the set of points $( \text{difficulty}(i), y_{ij})$ for  $i \in \{1, \ldots, N\}$. When ordered by difficulty,  $y_{ij}$ exhibits the behavior of algorithm $j$'s performance as datasets get progressively difficult. Thus, for each algorithm $j$, we can fit a  model explaining the performance by dataset difficulty values, which we will explore later. 

\subsection{Algorithm metric: Anomalous indicator}\label{subsubsec:anomalous}

\begin{figure}[!ht]
  \centering
  \includegraphics[scale=0.7]{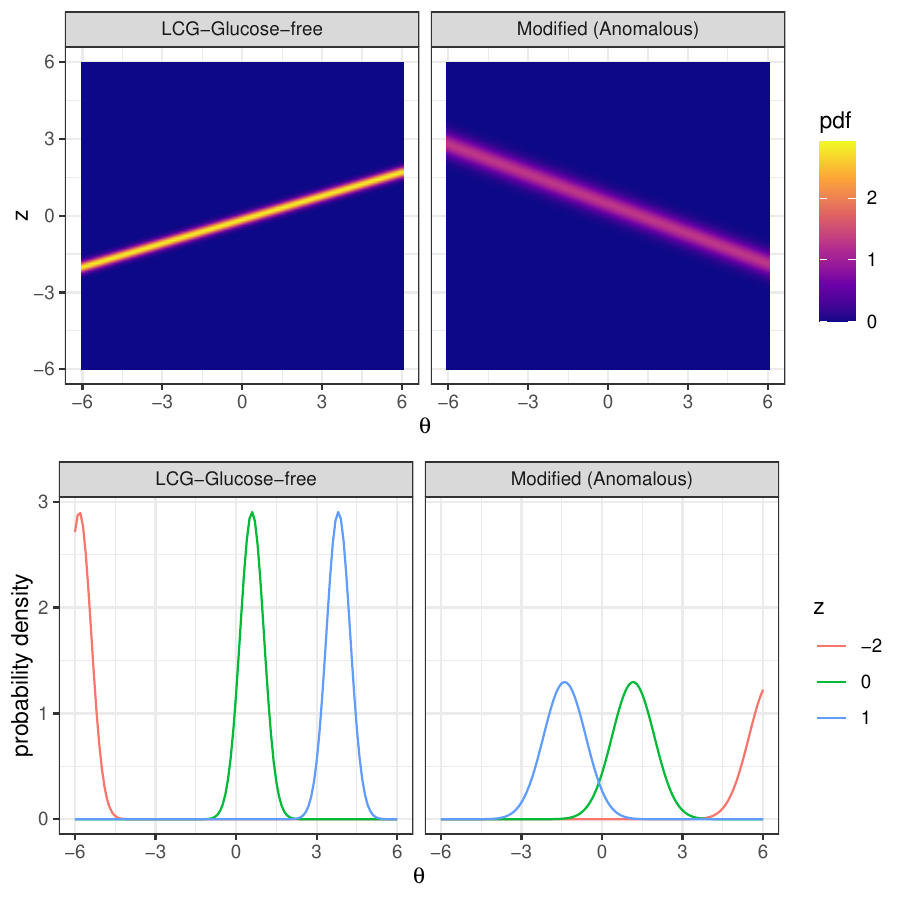}
   \caption{ The left column shows the heatmap and the trace lines for LCG-Glucose-free, a typical algorithm with increasing $\theta$ corresponding to increased performance.  The right column  shows the heatmap and the trace lines for an anomalous algorithm, which obtains high accuracy scores for difficult test instances and low accuracy scores for easy test instances.  }
    \label{fig:algoflipped}
\end{figure}

Consider the heatmap and the trace lines shown in Figure~\ref{fig:algoflipped}. The left column represents the algorithm LCG-Glucose-free and the second column   represents a different type of algorithm. The second algorithm is constructed using an algorithm in the Minizinc portfolio for illustrative purposes. Suppose these figures were generated from  an item in  educational testing. Then the left column shows the heatmap and the trace lines of a test item for which higher examinee ability corresponds to higher grades. On the other hand the right column shows a test item 
for which examinees with lower ability obtain higher grades than examinees with higher ability. Such a test item is said to  have  negative discrimination. The standard premise in educational testing is that high grades correspond to high ability. As such, a test item with negative discrimination is commonly revised to obtain a positive discrimination or removed from the pool of questions \citep{hambleton2013item}. 

However, in algorithm evaluation such a heatmap or a set of trace lines represent an algorithm or a dataset with an interesting quirk. For the standard IRT approach for algorithm evaluation, the heatmap and the trace lines represent a dataset, which gives poor performances for high ability algorithms and good performances for low ability algorithms. For the inverted IRT approach, the heatmap and trace lines represent an algorithm that performs well on difficult test instances and poorly on easy test instances. We describe such algorithms  as ``anomalous''.  Indeed, the \textit{no free lunch } concept emphasizes that no single algorithm performs better than other algorithms for all problems.

This is confirmed by the \textit{instance space} analyses conducted by Smith-Miles and co-authors \citep{smith2012measuring, Kang17, Munoz16b}. Furthermore, the instance space analyses for different problems show that even though some algorithms perform poorly on average, they often hold a niche in the instance space where they outperform  other algorithms \citep{normalizationoutliers}. As this is a unique strength of the algorithm, it should not be removed from the dataset as practiced in educational testing.  \\

%\subsubsection{Anomalous indicator}
For continuous and polytomous IRT models, the standard parameters for item $j$ comprise the discrimination parameter $\alpha_j$ and the difficulty parameter $\beta_j$ for continuous models,  and the intercepts $\bm{d}_j = \left(d_1, \ldots, d_{C_j-1} \right)$ for polytomous models. The discrimination parameter, which is present in both continuous and polytomous models highlights two aspects of algorithm performance. The sign of the discrimination parameter tells us if the algorithm is typical or anomalous. If $a_j <0$ then algorithm $j$ gives better performance values for difficult test instances and  low performances for easy test instances, and is considered anomalous. So we define the anomalous indicator as 
\begin{equation}\label{eq:defanomalous}
    \text{anomalous}(j) =   \left\{
                \begin{array}{ll}
                    \text{TRUE}  & a_j < 0 \, ,  \\
                     \text{FALSE} & \text{otherwise} \, .
            \end{array} 
            \right.
\end{equation}

\subsection{  Algorithm metric: Algorithm consistency score}\label{subsubsec:stability}

\begin{figure}[!ht]
    \centering
    \includegraphics[scale=0.8]{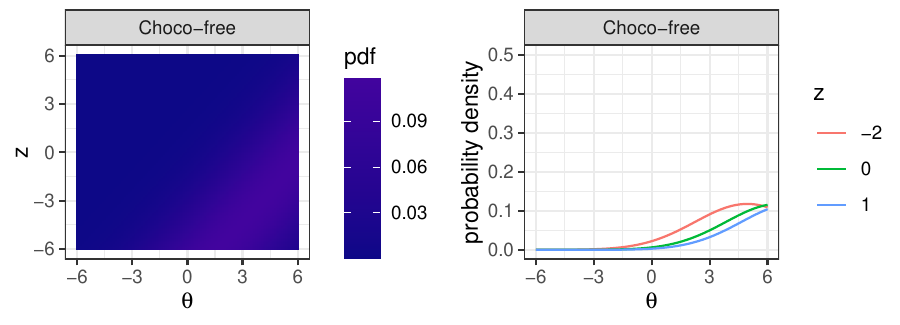}
    \caption{ The heatmap and the trace lines for Choco-free, a relatively consistent algorithm in this portfolio.  }
    \label{fig:algoreliable}
\end{figure}

Consider the heatmap and the trace lines in Figure~\ref{fig:algoreliable}. Suppose these trace lines relate to a test item in an educational testing scenario. Then this item does a poor job in discriminating examinees with different abilities, because all examinees are most  likely to obtain a similar score regardless of their ability. 

In algorithm evaluation using the standard IRT approach, such a heatmap and trace lines indicate that the dataset in question does not discriminate between the algorithms. That is, the dataset might be too difficult for all algorithms or too easy for all algorithms. 
Similarly, in algorithm evaluation using the inverted IRT approach, such a heatmap and trace lines indicate that the algorithm  does not discriminate. That is, regardless of the easiness/difficulty of the test instance,  this algorithm is most likely to give a similar score, i.e. its sensitivity to test instances is quite low. Thus, the algorithm is  consistent and non-discriminative. The  consistency or robustness of an algorithm is an important characteristic that is sometimes overlooked in the quest for peak performance.   

Stability or robustness can be defined in different ways. For example, \cite{Eiben2011} discuss 3 types of  robustness indicators: robustness with respect to parameters, problem specification and random seeds. Often robustness or stability is defined as a measure of the change of the output with respect to a small perturbation of the input. In our case, we do not perturb the input; however datasets positioned close to each other in the latent trait continuum are considered to have similar easiness/difficulty level. As such, a measure of the change of performance values across the latent trait continuum is an indication of stability  or robustness. However, stability or robustness  are positive attributes. The algorithm quality we want to encapsulate is slightly different in the sense that some algorithms can consistently perform poorly irrespective of the problem while others can consistently perform well. We capture this notion by defining algorithm consistency. 

% \subsubsection{Stability indicator}
The absolute value of the discrimination parameter $|a_j| $ gives the discrimination power of the algorithm, which is linked to the consistency of the algorithm. If $|a_j| $ is small, then the algorithm will produce trace curves with slower transitions similar to those in Figure~\ref{fig:algoreliable}, signifying a more consistent algorithm than one with a larger $|a_j| $. As such, we define  consistency as 
\begin{equation}\label{eq:defstability}
    \text{consistency}(j) = \frac{1}{\vert a_j \vert} \, .
\end{equation}

{\color{black} Tying this back to the heatmaps, the discrimination power of the algorithm is connected with the sharpness of the lines/bands on the heat map.  In Figure \ref{fig:crmparameters22} we see that some algorithms have sharp lines while others have blurry lines. Algorithms with sharp lines are more discriminating than algorithms with blurry lines, i.e., algorithms with blurry lines or no lines are more consistent than algorithms with sharp lines.    }

\subsection{ Algorithm metric: Difficulty limit}\label{subsubsec:easiness}
\begin{figure}[!ht]
    \centering
    \includegraphics[scale=0.7]{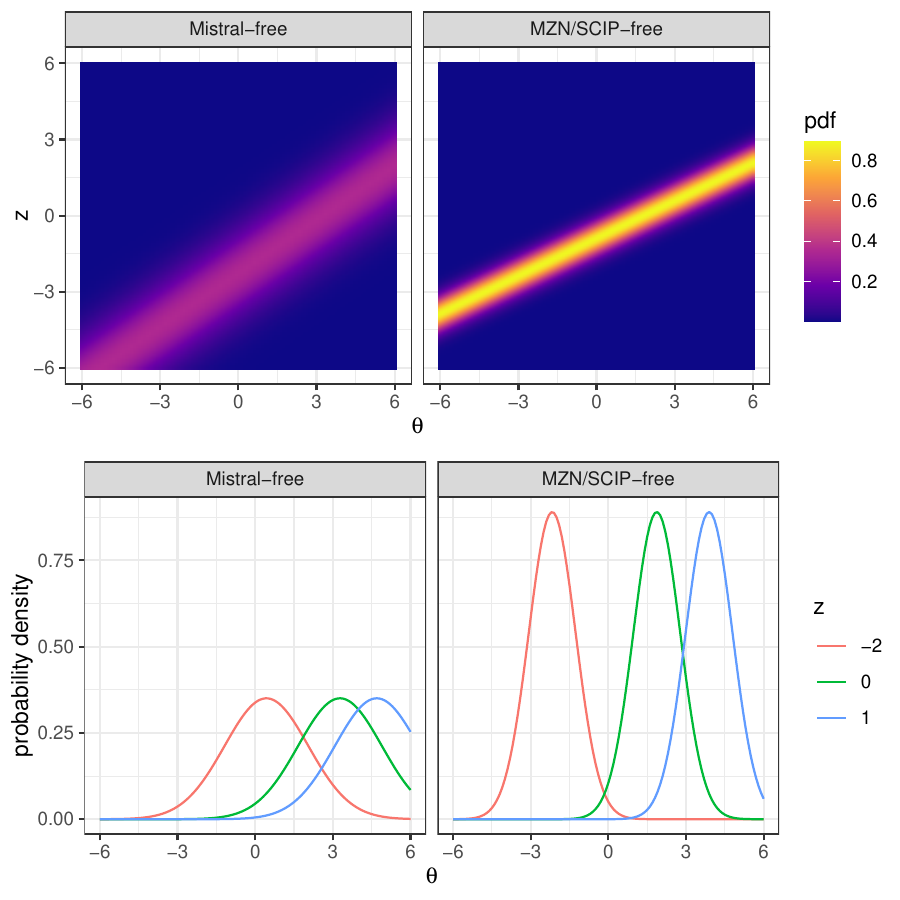}
    \caption{  Two algorithms with different difficulty limits. Mistral-free has a higher difficulty limit than MZN/SCIP-free.}
    \label{fig:easinessthreshold}
\end{figure}

Both consistency and anomalousness relate to the IRT discrimination parameter. Next, we discuss the role of the item difficulty parameter in the inverted IRT algorithm evaluation approach. Suppose Figure~\ref{fig:easinessthreshold} represents two items in educational testing. The first and the second columns in Figure~\ref{fig:easinessthreshold} show the trace lines and the heatmaps of two items, with the item in the left column having higher difficulty. We see that for any given ability $\theta$, the most probable score in the heatmap in the right column is higher than that of the left column. 

In the inverted IRT approach, the heatmaps and the tracelines represent algorithms with the algorithm in the left column, Mistral-free, giving lower performance for similar datasets compared to the algorithm in the right column,  MZN/SCIP-free.  When we consider dataset difficulty ($-\theta$), we see that as datasets get more difficult the algorithm performance goes down. Thus, each algorithm has an upper limit in terms of dataset difficulty. If the difficulty of a dataset is lower than this limit, we expect the algorithm to give good results, but if it is higher than the limit, the algorithm would perform poorly. Therefore, we define the algorithm difficulty limit as
\begin{equation}\label{eq:CRMdifficulty}
\text{difficulty}(j) = -\beta_j \, ,
\end{equation}
where $\beta_j$ is the traditional IRT difficulty parameter. Higher values of $\text{difficulty}(j)$ indicate better algorithms that can handle more difficult datasets.

For polytomous IRT, as there are multiple difficulty parameters $(d_1, d_2, \ldots, d_{C_j -1})$, we use $-d_{C_j -1}$ as the difficulty limit, because this denotes the threshold for the highest performance level.   

\section{ Evaluating algorithm portfolios using AIRT}\label{sec:airteval}

\subsection{Modelling algorithm performance based on dataset difficulty}\label{latenttrait} 
The dataset difficulty spectrum gives a way of ordering the performance values $y_{ij}$. For each algorithm $j$, we can consider the set of points $(\delta_i, y_{ij})$ for  $i \in \{1, \ldots, N\}$. When ordered by $\delta_i$,  $y_{ij}$ exhibits algorithm $j$'s performance as datasets get progressively difficult. Thus, for each algorithm $j$, we can fit a  model explaining the performance by dataset difficulty values. These models can be denoted by functions $\{ h_j(\delta)\}_{j=1}^n$, where $j$ denotes the algorithm and $\delta$ the dataset difficulty. For simplicity our $h_j$'s are smoothing splines.  

The smoothing spline $h_j$ minimizes the function
\begin{equation}\label{eq:smoothingsplines1}
    \sum_{i=1}^N \left( y_{ij} - h_j(\delta_i) \right)^2 + \lambda \int h''_j(t) \, dt \, , 
\end{equation}
where the first term denotes the sum of squared errors and the second term is a penalty for wiggliness. It is the second term -- the integral of the second derivative -- that gives the smoothness to the spline.  The parameter $\lambda$ is a tuning parameter and is computed by using a closed-form expression that minimizes the leave-one-out cross validation squared error \citep{hastietibshirani}. 

An advantage of using smoothing splines is that we do not need to specify any parameters to fit the splines. Furthermore, by graphing the splines we can visualize regions of the latent trait where algorithms give good or weak performance. 

We note that this is a feature-less way of exploring algorithm performance. For example, in instance space analysis we compute features of datasets and explain algorithm performance using these features. AIRT explains algorithm performance using dataset difficulty, which is computed from fitting an IRT model without using external features.  

\begin{figure}[!ht]
    \centering
    \includegraphics[scale=0.8]{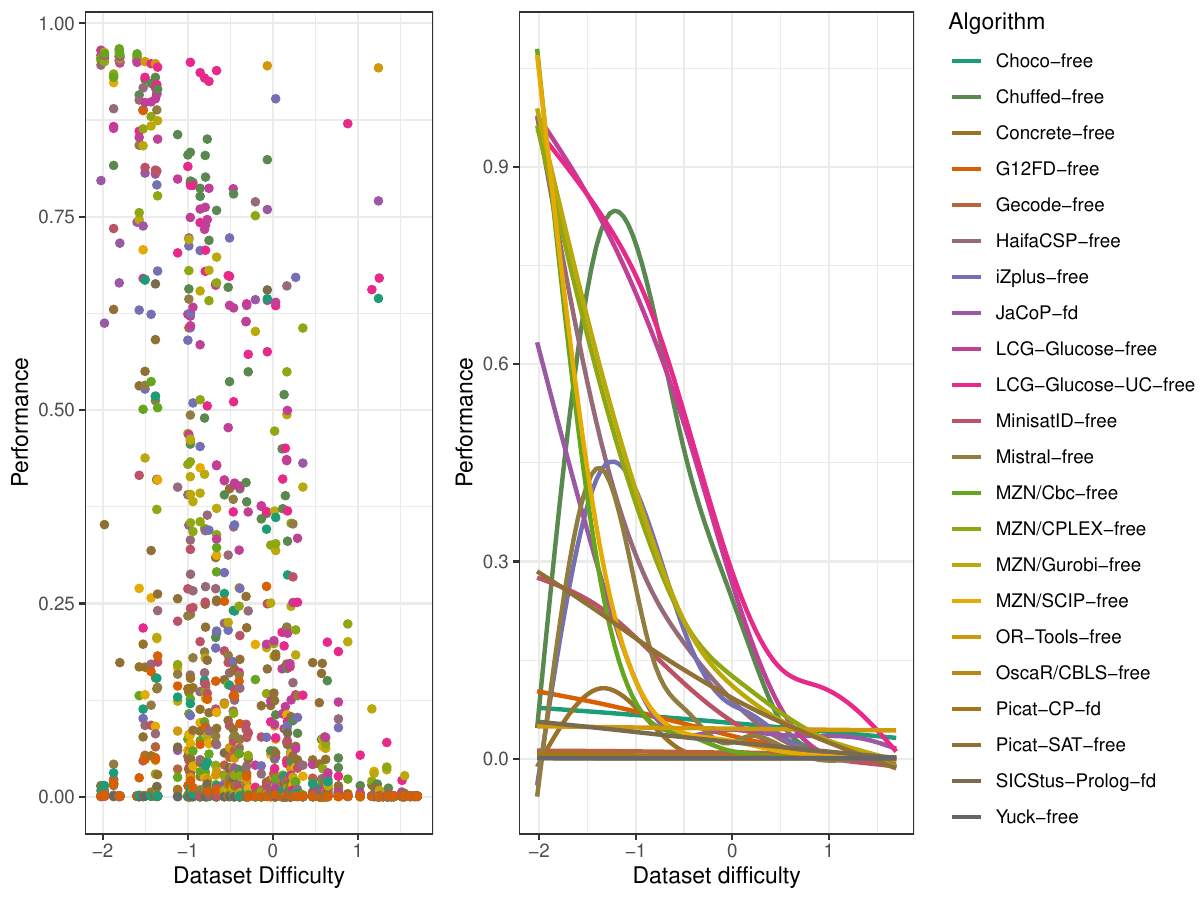}
    \caption{  The dataset difficulty spectrum of CSP-Minizinc-2016 explored. Left: Algorithm performance against dataset difficulty for all 4 algorithms. Right: Smoothing splines fitted to algorithm performance values. }
    \label{fig:latenttrait}
\end{figure}

CSP-Minizinc-2016 algorithm portfolio ordered by dataset difficulty and the fitted smoothing splines are shown in Figure~\ref{fig:latenttrait}. From this diagram we see that  different algorithms perform better for different values of dataset difficulty.

\subsection{Strengths and weaknesses of algorithms}\label{sec:strengthsweaknesses}  
We can compute the strengths and weaknesses of algorithms using the dataset/problem difficulty spectrum. To find the algorithm strengths we first find the best algorithm performance for each value $\delta$ in the problem difficulty spectrum. That is, 
\begin{align}\label{eq:maxlatent}
    h_{j_*}(\delta) & = \max_j h_j(\delta) \, . 
\end{align}
Next, for a given $\epsilon > 0$ we define the strengths of algorithm $j$ as
\begin{equation}\label{eq:strength}
    \text{strengths}(j, \epsilon) = \left\{\delta : |h_j(\delta) - h_{j_*} (\delta)|   \leq \epsilon \right\} \, .
\end{equation}

That is, the strengths of algorithm $j$ denote the regions in the problem difficulty spectrum where algorithm $j$ gives good performance. Here good is defined as close to best, specifically within $\epsilon$ from the best. As such, we can get multiple contiguous regions of strengths for some algorithms while others may not have any strengths in the spectrum for a given $\epsilon$. 

Algorithm weaknesses are found similarly. To compute the weaknesses we first find the poorest algorithm performance for every point in the problem difficulty spectrum: 
\begin{align}\label{eq:minlatent}
    h_{j_\#}(\delta) & = \min_j h_j(\delta) \, . 
\end{align}
Then, we define the weaknesses of algorithm $j$ as 
\begin{equation}\label{eq:weakness}
    \text{weaknesses}(j, \epsilon) = \left\{\delta : |h_j(\delta) - h_{j_\#} (\delta)|   \leq \epsilon \right\} \, .
\end{equation}  
Weaknesses represent regions in the problem difficulty spectrum where algorithms give poor performance. 

\begin{figure}[!ht]
    \centering
    \includegraphics[scale=0.8]{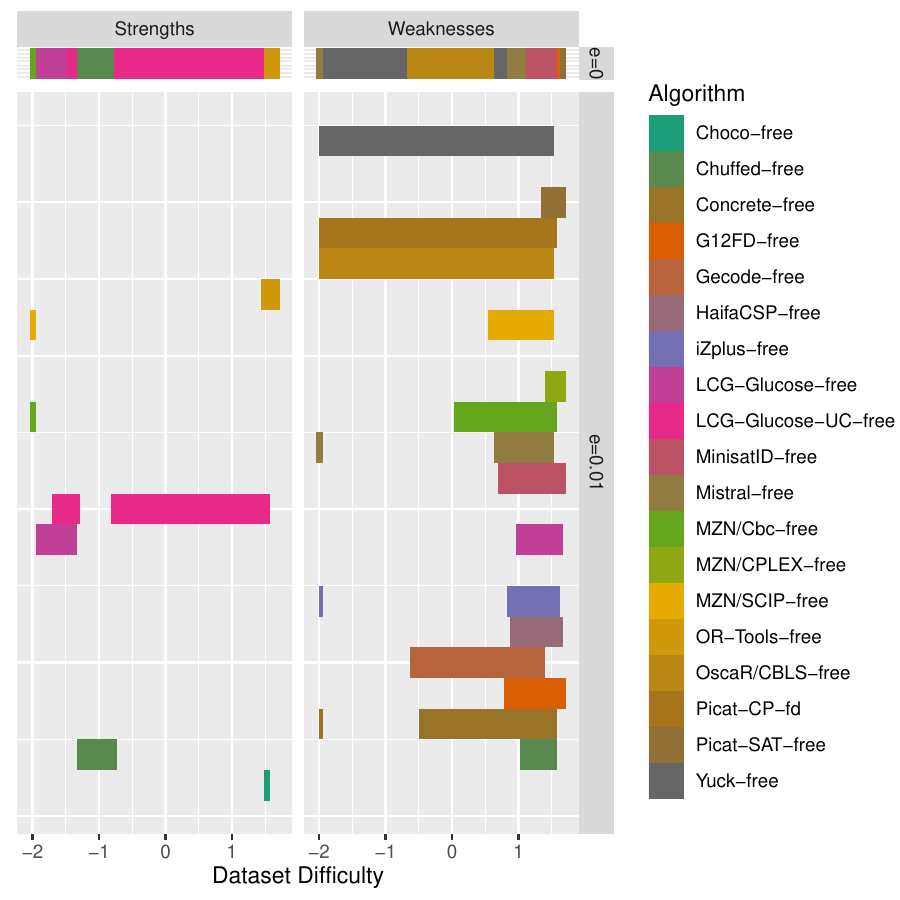}
    \caption{  Strengths and weaknesses of CSP-Minizinc-2016 algorithms  for $\epsilon = 0$ and $\epsilon = 0.01$. }
    \label{fig:strwknes}
\end{figure}

Figure~\ref{fig:strwknes} shows the strengths and weaknesses of   CSP-Mnizinc-2016 algorithm portfolio. The top row shows the strengths and weaknesses for $\epsilon = 0$ and the bottom part for $\epsilon = 0.01$. The difference between the two values of $\epsilon$ is that when $\epsilon = 0$, for each value $\delta$ in the dataset difficulty spectrum there is only one algorithm that is strong. When $\epsilon \neq 0$ multiple algorithms can display strengths for the same $\delta$. 

In Figure~\ref{fig:strwknes} we see that when $\epsilon = 0$  LCG-Glucose-UC-free is strong for a large part of the problem space, including difficult and medium-difficult problems. OR-Tools-free is better for more difficult problems and LCG-Glucose-free and Chuffed-free for easy problems. {\color{black} For $\epsilon = 0$ only 5 algorithms have strengths.} When $\epsilon = 0.01$ we see a little overlap. {\color{black} However, when $\epsilon = 0.01$ only 7 algorithms out of 22 algorithms exhibit strengths. In contrast, 16 algorithms have weaknesses when $\epsilon = 0.01$.} Both LCG-Glucose-UC-free and LCG-Glucose-free have strengths for easier problems {\color{black} but LCG-Glucose-UC-free remains the more powerful algorithm}. In the weaknesses space, we see Picat-CP-fd, OscaR/CBLS-free and Yuck-free displaying weaknesses for most of the problem space. A large number of algorithms are weak for difficult problems as seen for $\epsilon = 0.01$.  

 Using the strengths we compute the \textit{latent trait occupancy} (LTO) for each algorithm. LTO gives the proportion of datasets supported by each algorithm in the region of its strength. We define it as 
\begin{equation}
    \text{LTO}(j, \epsilon) = \frac{ \vert \{i: i \in \text{strengths}(j, \epsilon) \}\vert }{N}\, , 
\end{equation}
where $i$ and $j$ denote the datasets and algorithms respectively. The total number of datasets/problems is denoted by $N$. For the strengths shown in Figure~\ref{fig:strwknes} for $\epsilon = 0$,  LCG-Glucose-UC-free occupies the largest portion of the latent trait followed by Chuffed-free. When $\epsilon > 0$, the quantity $\sum \text{LTO} > 1$ if the strengths of algorithms overlap as shown in Figure~\ref{fig:strwknes}.  The LTO values for both $\epsilon = 0$ and $\epsilon = 0.01$ is listed in Table \ref{tab:ltominizinc}. 

{\color{black} Combining Figure~\ref{fig:strwknes} and Table \ref{tab:ltominizinc} we see that for very easy problems $(\delta \approx -2)$ algorithms MZN/Cbc-free and MZN/SCIP-free display strengths. However, we see that the latent trait occupancy LTO = 0.02, which is very small. Therefore, even if these two algorithms have strengths for very easy problems, it is risky to use them because of small LTO. For easy problems ($\delta \leq 0)$ we have 3 candidates: LCG-Glucose-UC-free, Chuffed-free and LCG-Glucose-free. The LTO of these algorithms are 0.828, 0.141 and 0.111 respectively. Basically, this reiterates that LCG-Glucose-UC-free is the most powerful algorithm. For very hard datasets ($\delta > 1)$, we have 3 candidates, Choco-free, OR-Tools-free and LCG-Glucose-UC-free. Of these, Choco-free has an LTO of 0.04, and thus can be disregarded.  OR-Tools-free occupies the same position in the strengths diagram for both $\epsilon = 0$ and $\epsilon = 0.01$ and thus has a unique strength for very difficult problems. }

\begin{table}[!ht]
		\centering
		\caption{ AIRT Latent Trait Occupancy (LTO) for CSP-Minizinc-2016 algorithms. }
        {
		\begin{tabular}{lrr}
			\toprule
            Algorithm & LTO ($\epsilon = 0$) & LTO ($\epsilon = 0.01$) \\
			\midrule
 
 LCG-Glucose-UC-free    &  0.717    &   0.828  \\
 Chuffed-free           &  0.121    &   0.141  \\
 LCG-Glucose-free       &  0.071   &   0.111  \\
 OR-Tools-free          &  0.071   &   0.071 \\
 MZN/Cbc-free           &  0.020   &   0.020 \\
 Choco-free             &  0       &    0.040 \\
 MZN/SCIP-free          &  0      &    0.020 \\
 
		    \bottomrule
		\end{tabular} }
		\label{tab:ltominizinc}
\end{table}

\subsection{Algorithm portfolio  selection}\label{sec:portfolioselection}  
The analysis in the previous section can be used understand the strengths and weaknesses of algorithms, adding to the exploratory data analysis domain of algorithm portfolios. We can also use AIRT for algorithm portfolio selection. We construct the airt portfolio by selecting the set of strong algorithms for a given $\epsilon$.

Formally, the airt portfolio is defined as
\begin{align}\label{eq:strongairt}
 \mathscr{A}(\epsilon)    & =   \left\{j : |h_j(\delta) - h_{j_*} (\delta)|   \leq \epsilon, \,  \text{for all} \, \delta \right\} \, , \\ 
 & = \{j : \text{strengths}(j, \epsilon) \neq \emptyset \} \, .
\end{align}

When $\epsilon = 0 $ we obtain $\mathscr{A}(0) = \bigcup j_*$,  i.e., the strongest  set of algorithms in the latent space. 

We use lowercase letters `airt' when describing portfolio specific results and uppercase AIRT when describing more general aspects. The number of algorithms in the airt portfolio depends on $\epsilon$. However, we do not directly specify the number of algorithms. It is a result of the smoothing splines $\{ h_j(\delta)\}_{j=1}^n$, which use the dataset difficulty spectrum $\delta$ as the input. But, $\delta_i = - \theta_i$, which is computed using $\alpha_j$, $\beta_j$, $\gamma_j$ and $z_{ij}$ as dictated by equation~\eqref{eq:CRM15}. Therefore, the AIRT model has a direct influence on the portfolio.

Of course, the airt portfolio, strengths and weaknesses and other indicators of algorithm performance are only reliable if the IRT model providing the parameters has a good fit.  In the following section we provide some measures of goodness of the IRT model to support interpretation of the results.

\subsection{IRT Model goodness measures} 
We are using IRT to model algorithm performance, that is the IRT model is effectively a meta-model. Checking the accuracy or the goodness of the IRT model is important because it determines the confidence we can place on the IRT model parameters, which describe the algorithms. If the IRT model is accurate, then we can trust the relationships it has modeled between instances and algorithm performances. 

\begin{figure}[!ht]
    \centering
    \includegraphics[scale=0.8]{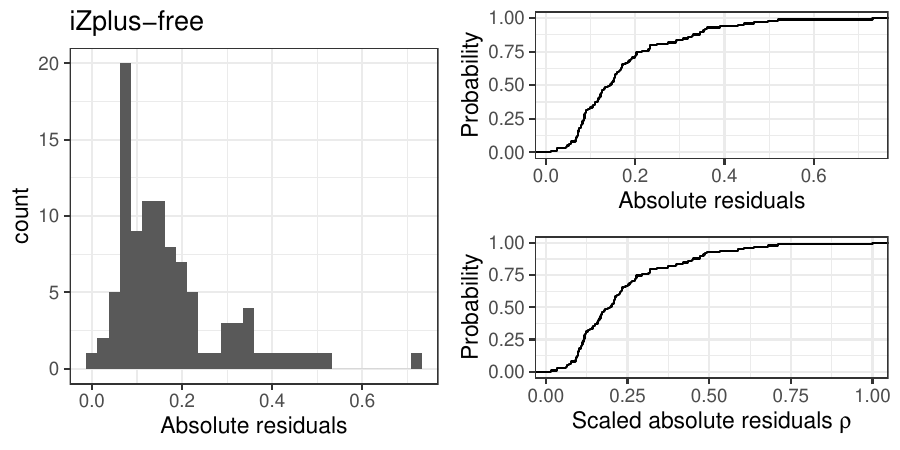}
    \caption{ The histogram of the absolute residuals $|e_{ij}|$ of iZplus-free is shown on the left. The CDF of the absolute residuals is shown on the top right and the CDF of the scaled absolute residuals is shown on the bottom right. Notice the difference in the domain for the two CDFs. }
    \label{fig:scaledresiduals}
\end{figure}

After fitting a continuous (polytomous) IRT model we define the predicted result (category) for a test instance $i$, with latent score $\theta_i$ as the result (category) with the highest probability for latent score $\theta_i$. We denote the predicted result (category) for test instance $i$ and algorithm $j$ by $\hat{x}_{ij}$. Then the residuals $e_{ij} = x_{ij} - \hat{x}_{ij}$ are of interest to us.  For a fixed $j$, let $ e_j = \{e_{ij} \}_{i=1}^N$ denote the residuals of the $j^{\text{th}}$ algorithm.
We consider the scaled absolute residuals $ \rho_{ij} =  c\vert e_{ij}  \vert$, such that $\rho_{ij} \in [0, 1]$. As we are interested in the algorithms we define $\rho_j = \{ \rho_{ij} \}_{i=1}^N$ and consider the empirical cumulative distribution function (CDF) of $\rho_{j}$ for each $j$, which we denote by $F(\rho_j)$:
\begin{equation}\label{eq:modelgood1}
    F(\rho_j)  = P(\rho_j  \leq \rho) \quad \text{for} \quad \rho \in [0, 1] \, .
\end{equation}
Figure~\ref{fig:scaledresiduals} shows a histogram of the absolute residuals $|e_{ij}|$, the  empirical cumulative distribution functions of $|e_{ij}|$, and the scaled absolute residuals $\rho_{ij}$ for iZplus-free algorithm in CSP-Minizinc-2016 portfolio. The only difference between the two CDFs is the $x$ values, which are in the interval $[0,1]$ for the scaled absolute residuals. 

By rescaling the absolute residuals to $[0,1]$ we make sure that the area under the CDF $F(\rho_{j})$, denoted by $\text{AUCDF}(\rho_j)$, is bounded by 1. $\text{AUCDF}(\rho_j)$ provides a measure of goodness of the IRT model for algorithm $j$. A higher AUCDF signifies better IRT model fit. 

We compute the mean square error (MSE) of the residuals and $\text{AUCDF}(\rho_j)$ for each algorithm $j$. IRT may fit some algorithms better than others. We note that the log-likelihood obtained from fitting the IRT model is an aggregate and therefore does not show how well each algorithm is fitted. By computing the residual metrics such as mean square error and $\text{AUCDF}(\rho_j)$ we gain a better understanding of the IRT model in relation to each algorithm.

\subsection{Predicted and actual effectiveness } 
We are interested in how well algorithms perform on test instances, especially the high performance results. If an algorithm gives good performance results for most test instances, then that algorithm is effective. As such, we focus on the performance results in decreasing order and study effectiveness via the  cumulative distribution function (CDF) for each algorithm. First we denote the algorithm performance results for algorithm $j$ by $x_j = \{x_{ij} \}_{i=1}^N$. By defining $t_j = \max(x_j) - x_j$ we reverse the performance results so that small values of $t_j$ denote high performance results. The variable $t_j$ can be thought of as a tolerance parameter, i.e. small tolerances give better performance. Then we compute the effectiveness of the algorithm by  
\begin{equation}\label{eq:ccdf1}
    \bar{F}_j(\ell) = P(t_j \leq \ell)\, , 
\end{equation}
where $P$ denotes the probability. The function $\bar{F}_j(\ell)$ is also related to the complementary cumulative distribution (CCDF), which is defined as
\begin{equation}\label{eq:ccdf3}
    \bar{F}_{x}(\ell) = P(x \geq \ell)\, ,  
\end{equation}
since,
\begin{align}\label{eq:ccdf2}
    P(t_j \leq \ell) & = P\left( \max(x_j) - x_j \leq  \ell \right)\, , \\
     & = P\left(  x_j \geq  \max(x_j) - \ell \right)\, . 
\end{align}
As such, $\bar{F}_j(\ell)$ denotes the CCDF of $x_j$ with the $x$ axis reversed.  

We call the curve $y= \bar{F}_j(\ell)$, the effectiveness curve.  By scaling $t_j$ to lie in $[0, 1]$, we make sure that the area under the effectiveness curve  is bounded by 1. For polytomous IRT with categories $\{0, 1, \ldots, C_{j-1}\}$,  we consider a step size of $\Delta = \frac{1}{C_{j-1}}$ for the $x$ axis with $\ell \in \{0, 1, \ldots, C_{j-1}\}$, so that the curve $y = \bar{F}_j(\ell)$ is defined by the points $\left(0, \bar{F}_j(C_{j-1}) \right),  \left(\Delta, \bar{F}_j(C_{j-2}) \right), \ldots, \left(1, \bar{F}_j(0) \right)$. Figure~\ref{fig:effectiveness_example} shows the histogram, CDF of performance values (bottom-left) and the effectiveness curve (bottom-right) of  Chuffed-free algorithm in CSP-Minizinc-2016 portfolio.    

\begin{figure}[!ht]
    \centering
    \includegraphics[scale=0.7]{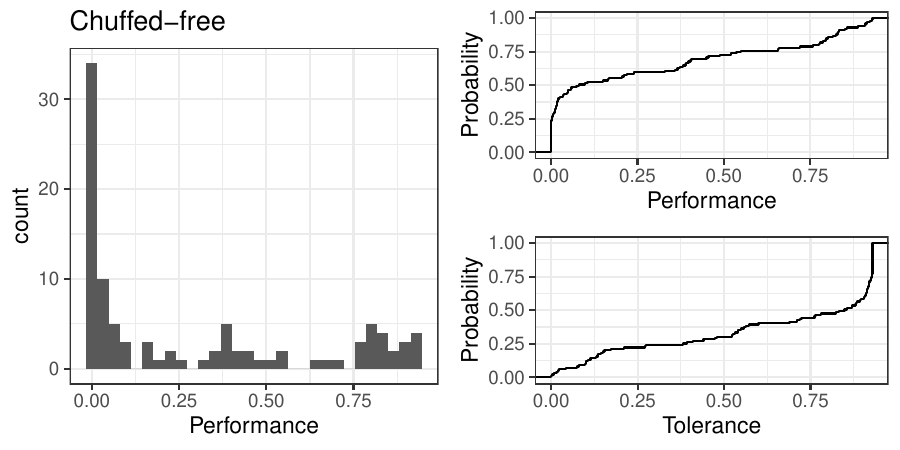}
    \caption{ The histogram of performance values Chuffed-free algorithm is shown on the left. The graph on the top right shows the CDF of the performance values. The graph on the bottom right shows the effectiveness curve $y= \bar{F}_j(\ell)$. }
    \label{fig:effectiveness_example}
\end{figure}

Similarly, we can compute the effectiveness for the IRT predicted algorithm performance values by defining $\hat{x}_j = \{\hat{x}_{ij} \}_{i=1}^N$, and $\hat{t}_{j} = \max{\hat{x}_j} - \hat{x}_j$ where $\hat{x}_{ij}$ denotes the predicted result for algorithm $j$ and test instance $i$. This gives the predicted effectiveness 
\begin{equation}\label{eq:ccdf4}
    \bar{F}_{j}(\hat{\ell}) =P(\hat{t}_j \leq \ell)\, , 
\end{equation}
where we have indicated that it is a predicted quantity by using  $\hat\ell$. We have denoted the effectiveness by $\bar{F}$ for both predicted and actual values, while changing from $\ell$ to $\hat{\ell}$ for predicted effectiveness. We compute the area under the actual and predicted effectiveness curves as this is a measure of an algorithm's ability to produce high performance results. We denote the area under the actual effectiveness curve $ y= \bar{F}_j(\ell)$ by $\text{AUAEC}(j)$, and area under the predicted effectiveness curve $ y= \bar{F}_j(\hat\ell)$  by $\text{AUPEC}(j)$. A high $\text{AUAEC}(j)$ indicates that algorithm $j$ has a large proportion of high performance results and a high $\text{AUPEC}(j)$ indicates that the IRT model predicts algorithm $j$ to have a large proportion of high performance results. 

\begin{figure}[!ht]
    \centering
    \includegraphics[scale=0.7]{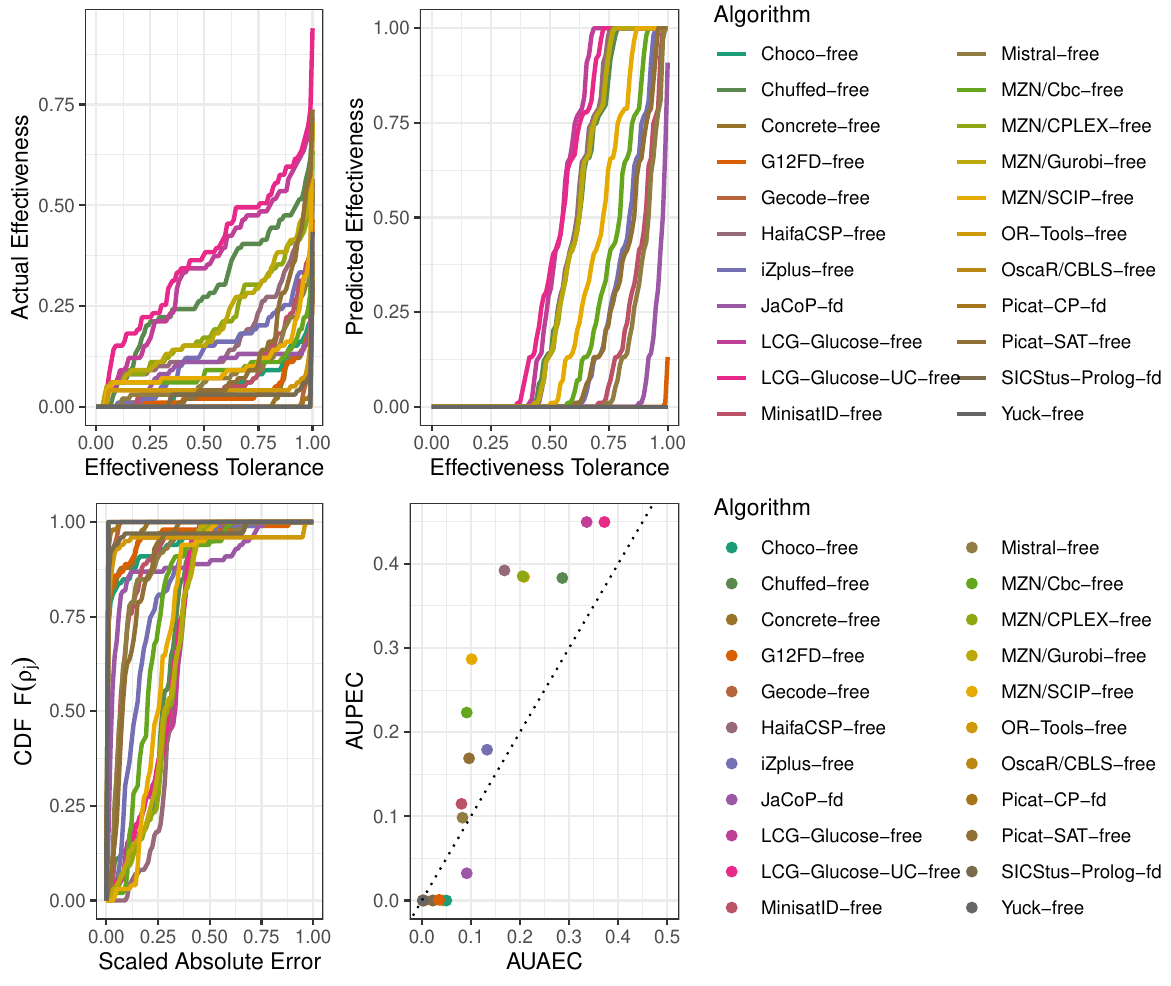}
    \caption{ The model goodness graphs for CSP-Minizinc-2016 portfolio. The top row shows actual and predicted effectiveness curves. The graph on the bottom left shows the CDF of the absolute residuals and the graph on the bottom right shows the actual and predicted effectiveness of the algorithms. }
    \label{fig:effectiveness_example}
\end{figure}

For a single algorithm the pair of values (AUAEC, AUPEC) gives an indication about the algorithm's actual and perceived ability to produce high performance results. If the absolute difference between the predicted and actual effectiveness, $| \text{AUAEC} - \text{AUPEC}|$ is large, then the trustworthiness of the IRT model is low for that algorithm. It may be the case that AUAEC $\approx$ AUPEC for most algorithms in a portfolio, but for one algorithm the absolute difference between AUAEC and AUPEC is higher. A larger absolute difference between AUAEC and AUPEC will concur with a lower AUCDF for that algorithm. For example if the IRT model over estimates the performance of an algorithm, AUPEC will be higher than AUAEC. This will also result in lower agreement between the predicted and the actual results giving rise to lower AUCDF. {\color{black} Table \ref{tab:AccuracyReliabilityCTS} gives the model goodness measures for each algorithm. We see that the MSE is low for most algorithms apart from HaifaCSP-free, which also has the highest $| \text{AUAEC} - \text{AUPEC}|$. In terms of goodness of fit, we can say that the IRT model is a good fit for mostly all algorithms, apart from HaifaCSP-free.  }

\begin{table}[!ht]
		\centering
		\caption{ \color{black} MSE, AUCDF, Area Under Actual Effectiveness Curve  (AUAEC) and Predicted Effectiveness Curves (AUPEC) and |AUAEC - AUPEC| for CSP-Minizinc algorithms. }
  {\color{black}
		\begin{tabular}{lrrrrr}
			\toprule
		Algorithm  & MSE & AUCDF & AUAEC & AUPEC  & |AUAEC - AUPEC|  \\
			\midrule
iZplus-free      &   0.046 & 0.823 &  0.133   &  0.179     &      0.046 \\
MZN/SCIP-free    &   0.072 & 0.746 & 0.102    &  0.287     &      0.185 \\ 
Chuffed-free     &   0.085 & 0.733 & 0.286    &  0.383     &      0.097 \\
LCG-Glucose-UC-free & 0.088 & 0.725 & 0.372   &  0.450     &      0.078 \\
Concrete-free    &   0.003 & 0.974 & 0.022    &  0.000     &      0.022 \\ 
JaCoP-fd         &   0.051 & 0.894 & 0.092    &  0.032     &      0.059 \\
Mistral-free     &   0.027 & 0.892 & 0.083    &  0.098     &      0.016 \\ 
OscaR/CBLS-free  &   0.000 & 0.995 & 0.002    &  0.000     &      0.002 \\
HaifaCSP-free    &   0.100 & 0.690 & 0.168    &  0.392     &      0.224 \\ 
Gecode-free      &   0.000 & 0.989 & 0.005    &  0.000     &      0.005 \\
OR-Tools-free    &   0.037 & 0.951 & 0.043    &  0.000     &      0.043 \\
SICStus-Prolog-fd &   0.013 & 0.972 & 0.023   &  0.000     &      0.023 \\
Picat-CP-fd      &   0.000 & 0.993 & 0.002    &  0.000     &      0.002 \\
Picat-SAT-free   &   0.017 & 0.894 & 0.096    &  0.169     &      0.073 \\ 
MZN/Gurobi-free  &   0.089 & 0.720 & 0.208    &  0.384     &      0.176 \\
MZN/CPLEX-free   &   0.093 & 0.713 & 0.205    &  0.385     &      0.180 \\ 
LCG-Glucose-free &   0.089 & 0.722 & 0.336    &  0.450     &      0.114 \\
MZN/Cbc-free     &   0.058 & 0.786 & 0.092    &  0.223     &      0.132 \\
Yuck-free        &   0.000 & 0.995 & 0.002    &  0.000     &      0.002 \\
Choco-free       &   0.021 & 0.944 & 0.050    &  0.000     &      0.050 \\
MinisatID-free   &   0.022 & 0.898 & 0.081    &  0.115     &      0.034 \\ 
G12FD-free       &   0.014 & 0.961 & 0.035    &  0.001     &      0.034 \\

		    \bottomrule
		\end{tabular}
  }
		\label{tab:AccuracyReliabilityCTS}
\end{table}

The measures we have proposed for algorithm consistency score, anomalous indicator and difficulty limit are algorithm evaluation metrics while the  absolute residuals curve, actual and predicted effectiveness curves, along with AUCDF, |AUAEC - AUPEC| comprise AIRT's model goodness metrics. In the discussion that follows we refer to both AIRT and the underlying IRT model.  AIRT refers to the reinterpreted IRT model with the additional evaluation metrics discussed above. When we discuss standard IRT concepts such as trace lines we refer to the IRT model. 

This concludes the discussion on different aspects of the AIRT framework. The pseudocode given in Algorithm~\ref{algo:airt} summarizes the steps and functionality of AIRT.

\DontPrintSemicolon
\begin{algorithm}\fontsize{11}{12}\selectfont
	\SetKwInOut{Input}{input~~~}
	\SetKwInOut{Output}{output}
	\Input{~ The matrix $Y_{N \times n}$, containing accuracy measures of $n$ algorithms for $N$ datasets/problem instances. }
	\Output{ 1. AIRT indicators of algorithms and dataset/problem difficulty \\ 2. The strengths and weaknesses of algorithms \\
	3. airt algorithm portfolio \\
	4. Model goodness measures}
	{\textbf{Stage 1 - Fitting the IRT model with inverted mapping}} \\
	 {1. Transform the accuracy measures $y_{ij}$ by defining $z_{ij} = \ln\frac{ y_{ij}}{k - y_{ij}}$. } \\
	 {2. Let $Z = \{z_{ij}\} \in \mathbb{R}^{N \times n}$, where $N$ denotes the number of problems/datasets and $n$ denotes the number of algorithms.  } \\
	 {3. Fit a continuous IRT model to $Z$ by maximizing the log-likelihood function   
$$ E_{\bm{\theta}|\Lambda^{(t)}, \bm{Z}}\left[ \ln p\left( \bm{\Lambda} \vert \bm{\theta}, \bm{Z} \right) \right] = N \sum_{j=1}^n \left(\ln |\alpha_j|  + \ln|\gamma_j| \right) - \frac{1}{2} \sum_{i=1}^N \sum_{j=1}^n \alpha_j^2 \left( \left(  \beta_j + \gamma_j z_{ij} - \mu_i^{(t)} \right)^2 + \sigma^{(t)2} \right) + \ln p\left(\bm\Lambda \right) + \text{const} \, ,
$$ } 
	 {4. From this model we obtain (after $t$ iterations) the IRT discrimination and  difficulty parameters $\alpha_j$ and $\beta_j$  and the scaling parameter $\gamma_j$ for algorithms $j \in \{1, \ldots, n\}$ as follows:
 \begin{align*}
   \gamma_j^{(t+1)} & = \frac{V \left(\mu_i^{(t)}\right) + \sigma^{(t)2}}{C_j\left(z_{ij}, \mu_i^{(t)} \right)}  \, , \\
    \beta_j^{(t+1)} & = M\left( \mu_i^{(t)}\right) - \gamma_j^{(t+1)} M_j \left( z_{ij}\right) \, ,  \\
    \alpha_j^{(t+1)}  & =  \sign\left(\gamma_j^{(t+1)}\right)\left( \gamma_j^{(t+1)2} V_j(z_{ij}) - V \left(\mu_i^{(t)}\right) - \sigma^{(t)2} \right)^{-1/2} \, .
\end{align*}    
 } \\
    5. Using these IRT parameters we compute the latent trait $\theta_{N \times 1}$ as
  $$
  \theta_i  = \frac{\sum_j \hat{\alpha}_j^2 \left(\hat{\beta}_j + \hat{\gamma}_j z_{ij} \right)}{  \sum_j \hat{\alpha}_j^2} \, . 
  $$
  
	 {\textbf{Stage 2 - Calculation of algorithm and dataset metrics}} \\
	 { 6. For each algorithm $j$ compute the anomalous indicator,  algorithm consistency} score and difficulty limit using 
   
   \begin{align*}
    \text{anomalous}(j) & =   \left\{
                \begin{array}{ll}
                    \text{TRUE}  & a_j < 0 \, ,  \\
                     \text{FALSE} & \text{otherwise} \, .
            \end{array} 
            \right.  \, , \\ 
       \text{{ consistency}}(j) & = \frac{1}{\vert a_j \vert} \, , \\
       \text{difficulty}(j) & = -\beta_j \, ,
   \end{align*} 
  \\ 
	 { 7. For each dataset $i$ compute the dataset difficulty using 
  
$
 \delta_i = -\theta_i \, .
$} \\
  { \textbf{Stage 3 - Computing strengths and weaknesses and construct airt portfolio }}\\
	 { 8. Using the dataset difficulty spectrum $\delta$ fit smoothing splines $h_j(\delta)$ to performance values $y_{ij}$ for each algorithm $j$   minimizing
  $ \sum_{i=1}^N \left( y_{ij} - h_j(\delta_i) \right)^2 + \lambda \int h''_j(t) \, dt \, . $ } 
  \caption{\itshape AIRT framework.}
  \label{algo:airt}
\end{algorithm}

  \begin{algorithm}
	 { 9. Compute the strengths and weaknesses of algorithms using
  \begin{align*}
      \text{strengths}(j, \epsilon) & = \left\{\delta : |h_j(\delta) - h_{j_*} (\delta)|   \leq \epsilon \right\} \, , \\
       \text{weaknesses}(j, \epsilon) & = \left\{\delta : |h_j(\delta) - h_{j_\#} (\delta)|   \leq \epsilon \right\} \, ,
  \end{align*}
  and use the strengths and weaknesses for exploratory data analysis purposes.
  }  \\
	 { 10. Construct the airt portfolio using 
  $  \mathscr{A}(\epsilon) = \{j : \text{strengths}(j, \epsilon) \neq \emptyset \} \, .
  $ }  \\
	 { 11. Check the fit of the IRT model by computing model goodness measures MSE, AUCDF and |AUAEC - AUPEC|.  }
\end{algorithm}

\subsection{Computational complexity of AIRT}\label{sec:timecomplex}
To fit the IRT model, we use the non-iterative item parameter solution  proposed by \cite{SHOJIMA2005}. They use expectation maximization (EM) and in each EM cycle a non-iterative solution is found by optimizing the expectation in equation~\eqref{eq:CRM12} item-by-item. By computing partial derivatives and solving a set of simultaneous equations they find the exact solutions for item parameters $\alpha_j$, $\beta_j$ and $\gamma_j$ in each cycle. The optimization stops when solutions of successive cycles converge or when the maximum number of cycles is reached. Let $c$ denote the number of cycles. Hence, the computation is repeated $c$ times. For a $N \times n$ matrix $Z$, there are $n$ items and $N$ participants. The non-iterative solution is found for each item $j \in \{1, \ldots, n\}$. For a fixed $j$, solving for $\alpha_j$, $\beta_j$ and $\gamma_j$ involves computing various quantities such as mean, variance and covariance. The computational complexity of these operations is $\mathscr{O}(N)$. When they are computed for each item $j$ for  $c$ cycles the overall complexity of fitting the IRT model becomes $\mathscr{O}(Nnc)$. Of the three variables $c$ has an upper bound of $200$ and $n$ is much smaller than $N$. As such the most influencing variable is $N$. 
 
After fitting the IRT model we compute the anomalous indicator, algorithm consistency score and difficulty limit for each algorithm. These computations take a fixed amount of time for each algorithm $j$. Therefore, computing the indicators have $\mathscr{O}(n)$ complexity. Computing dataset difficulty values $\delta_i$ for $N$ datasets using equations~\eqref{eq:CRM15} and~\eqref{eq:CRMdifficultydat} takes $\mathscr{O}(N)$ complexity. Therefore, computing AIRT indicators and dataset difficulty values have $\mathscr{O}(N + n) \approx \mathscr{O}(N)$ complexity as $n$ is much smaller compared to $N$. 

Smoothing splines can be fitted in $\mathscr{O}(N)$ computational time. In statistical software packages, they are fitted using a much smaller number of points, approximately $\log(N)$ when $N > 50$ \citep{hastie2009elements}. Strengths and weaknesses of algorithms are computed mainly for visualization purposes. As such, the strengths and weaknesses horizontal bar graph has a smaller number of points compared to $N$; let us say it has $M$ points. For each of these points we compute the strengths and weaknesses of $n$ algorithms. This computation involves $\mathscr{O}(Mn)$ complexity; however vectorized computations make it much faster. The airt algorithm portfolio can be computed in fixed time as they take the union of strong or weak algorithms. 

Model goodness measures involve $n$ algorithms with $N$ data points for each algorithm. Computing the MSE, and the CDF for $\rho_j$  as in equation~\eqref{eq:modelgood1} have $\mathscr{O}(nN)$ complexity. Computing the area under the curve using trapezoidal integration takes $\mathscr{O}(N)$ time. Similarly, actual and predicted effectiveness have $\mathscr{O}(nN)$ complexity.

\section{Results}\label{sec:Results}  
We now test AIRT on 10 algorithm portfolios hosted on ASlib data repository \citep{Bischl2016}. ASlib hosts performance data and test instance features for a large number of algorithm portfolios. Section~\ref{subsec:openml} contains a detailed analysis of classification algorithms using AIRT. We explore AIRT metrics, model goodness measures and the strengths and weakness of algorithms using the dataset difficulty spectrum. In addition, we compare different algorithm portfolios. The analysis of classification algorithms encompasses the full functionality of AIRT. We carry out more concise analyses for other ASlib scenarios in Appendix~\ref{sec:App1}. We include the latent trait curves, strengths and weaknesses and algorithm portfolio comparisons for each ASlib scenario.

% =======================================================================
\subsection{  Detailed case study: Classification}\label{subsec:openml}
% The classification performance data on ASlib is extracted from the website \url{https://www.openml.org/} and contains results from the 2017 algorithm selection challenge \citep{lindauer2017open}. 
{\color{black} This scenario was introduced by \cite{Rijnthesis} and uses a selection of WEKA algorithms \citep{Hall2009}.  It was later used in the 2017 algorithm selection challenge by \cite{lindauer2017open}.} The dataset contains predictive accuracy results from 30 classification algorithms on 105 test instances. The default parameters and hyperparameters used by the classification algorithms were not varied.  For ease of plotting graphs, we have shortened the names of many algorithms. For example, there are 3 multilayer perceptron algorithms; algorithm 8990\_MultilayerPerceptron is renamed to 8990\_MLP.

\subsubsection{AIRT algorithm metrics}
\begin{figure}[!ht]
    \centering
    \includegraphics[scale=0.8]{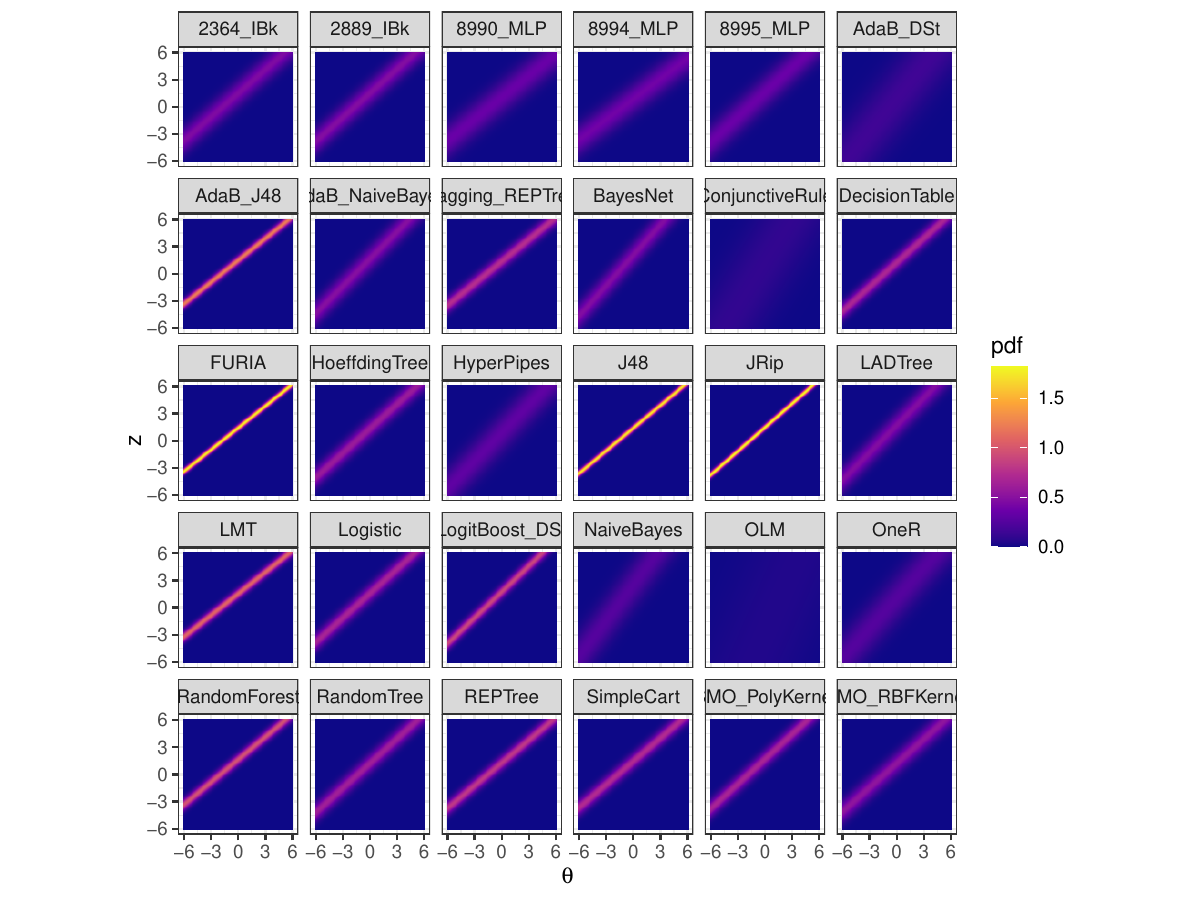}
    \caption{ The heatmap of probability density functions for classification (OpenML-weka-2017) algorithms by fitting a continuous IRT model}
    \label{fig:graphSurfaces}
\end{figure}

Figure~\ref{fig:graphSurfaces} shows the heatmaps of AIRT fitted probability distribution functions for the classification algorithms. We see that  OLM and ConjunctiveRule are more stable comparatively. AIRT did not find any algorithm to be anomalous.

Table~\ref{tab:graphColuringairtmetrics} gives AIRT metrics for the  classification algorithms. Even though OLM has the highest algorithm consistency, it has the lowest difficulty limit. Therefore, OLM gives poor performances consistently. Thus,  algorithm consistency by itself is not an indicator of a good algorithm. The  RandomForest has the highest difficulty limit. Hence, the RandomForest can handle very difficult instances. Algorithms LMT, NaiveBayes, SMO\_PolyKernel, AdaBoostM1\_J48 and BayesNet also have high difficulty limits meaning that these algorithms can handle hard instances. 

The RandomForest occupies the largest proportion in the latent trait (LTO) for $\epsilon = 0$ and the second largest for $\epsilon = 0.01$. Therefore, it is an excellent algorithm suited for a large number of diverse instances. Notably, LMT, the second best algorithm in terms of LTO for $\epsilon = 0$ surpasses the RandomForest and becomes the best algorithm for $\epsilon = 0.01$. This means, that even though it is not the topmost curve for most part of the latent trait, it is  $\epsilon$-close to the top curve mostly, and coupled with its own strengths on the latent trait it surpasses the RandomForest.  Algorithm AdaBoostM1\_J48 has a similar latent trait occupancy (LTO) as LMT when $\epsilon = 0$. Even though AdaBoostM1\_J48's LTO increases when $\epsilon = 0.01$, it doesn't increase as much as LMT's LTO does. Curiously, REPTree and 8990\_MLP have a similar proportion on the latent trait for both $\epsilon$ values. In contrast, algorithms such as J48, JRip and Bagging\_REPTree, increase their LTO from 0 to values greater than $0.1$ when $\epsilon$ increases from 0 to 0.01 -- a bigger increase than REPTree and 8990\_MLP undergo with the increase in $\epsilon$.  This observation suggests the two algorithms REPTree and 8990\_MLP  have unique strengths in the latent trait and not in other parts where more algorithms perform well.

\begin{table}[!ht]
		\centering
		\caption{  AIRT Metrics: algorithm consistency Score, Anomalousness indicator, Difficulty Limit and Latent Trait Occupancy (LTO) for classification algorithms. }
  {
		\begin{tabular}{lp{1.5cm}p{1.5cm}p{2.5cm}p{1.5cm}p{1.5cm}}
			\toprule
            Algorithm & Consistency & Difficulty Limit & Anomalousness & LTO ($\epsilon = 0$) & LTO ($\epsilon = 0.01$) \\
			\midrule
      8990\_MLP & 1.401   &   1.427  &  FALSE   &    0.010    &    0.038 \\
      8994\_MLP  & 1.349   &   1.271  &  FALSE   &    0.000    &    0.038 \\
      8995\_MLP & 1.201   &   1.842  &  FALSE   &    0.000    &    0.019 \\ 
            SMO\_PolyKernel & 0.655   &   1.950  &  FALSE   &    0.000    &    0.000 \\
                      OneR & 1.426   &   1.026  &  FALSE   &    0.000    &    0.000 \\ 
                       J48 & 0.274   &   1.749  &  FALSE   &    0.000    &    0.162 \\
                       2364\_IBk & 1.010   &   1.798  &  FALSE   &    0.000    &    0.000 \\
                   REPTree & 0.595   &   1.663  &  FALSE   &    0.029    &    0.076 \\ 
                 RandomTree & 0.709   &   1.457   & FALSE    &   0.000     &   0.000 \\ 
              RandomForest & 0.500    &  2.064   & FALSE    &   0.410     &   0.790 \\
                       LMT & 0.467    &  1.994   & FALSE    &   0.276     &   0.895 \\
             HoeffdingTree & 0.757    &  1.553   & FALSE    &   0.000     &   0.000 \\ 
             SMO\_RBFKernel & 0.842    &  1.508   & FALSE    &   0.000     &   0.000 \\
                      JRip & 0.253    &  1.741   & FALSE    &   0.000     &   0.124 \\ 
                      2889\_IBk & 0.950    &  1.812   & FALSE    &   0.000     &   0.000 \\
                HyperPipes & 1.272    &  0.919   & FALSE    &   0.000     &   0.000 \\
                NaiveBayes & 1.173    &  1.968   & FALSE    &   0.000     &   0.000 \\
                       OLM & 3.768    & -1.176   & FALSE    &   0.000     &   0.000 \\
                     FURIA & 0.281    &  1.806   & FALSE    &   0.000     &   0.314 \\
                  BayesNet & 0.752    &  1.942   & FALSE    &   0.000     &   0.000 \\
           ConjunctiveRule & 2.473    &  0.845   & FALSE    &   0.000     &   0.000 \\
                SimpleCart & 0.643    &  1.819   & FALSE    &   0.010     &   0.105 \\
     AdaBoostM1\_NaiveBayes & 0.819    &  1.750   & FALSE    &   0.000     &   0.000 \\
                   LADTree & 0.852    &  1.793   & FALSE    &   0.000     &   0.010 \\
                  Logistic & 0.669    &  1.824   & FALSE    &   0.000     &   0.000 \\ 
  AdaBoostM1\_DecisionStump & 2.069    &  0.882   & FALSE    &   0.000     &   0.000 \\
            AdaBoostM1\_J48 & 0.408    &  1.947   & FALSE    &   0.267     &   0.448 \\
           Bagging\_REPTree & 0.660    &  1.837   & FALSE    &   0.000     &   0.105 \\
             DecisionTable & 0.645    &  1.532   & FALSE    &   0.000     &   0.067 \\
  LogitBoost\_DecisionStump & 0.473    &  1.927   & FALSE    &   0.000     &   0.000 \\
		    \bottomrule
		\end{tabular}
  }
		\label{tab:graphColuringairtmetrics}
\end{table}

\subsubsection{Strengths and weaknesses of algorithms via AIRT}

\begin{figure}[p]
    \centering
    \includegraphics[scale = 0.75]{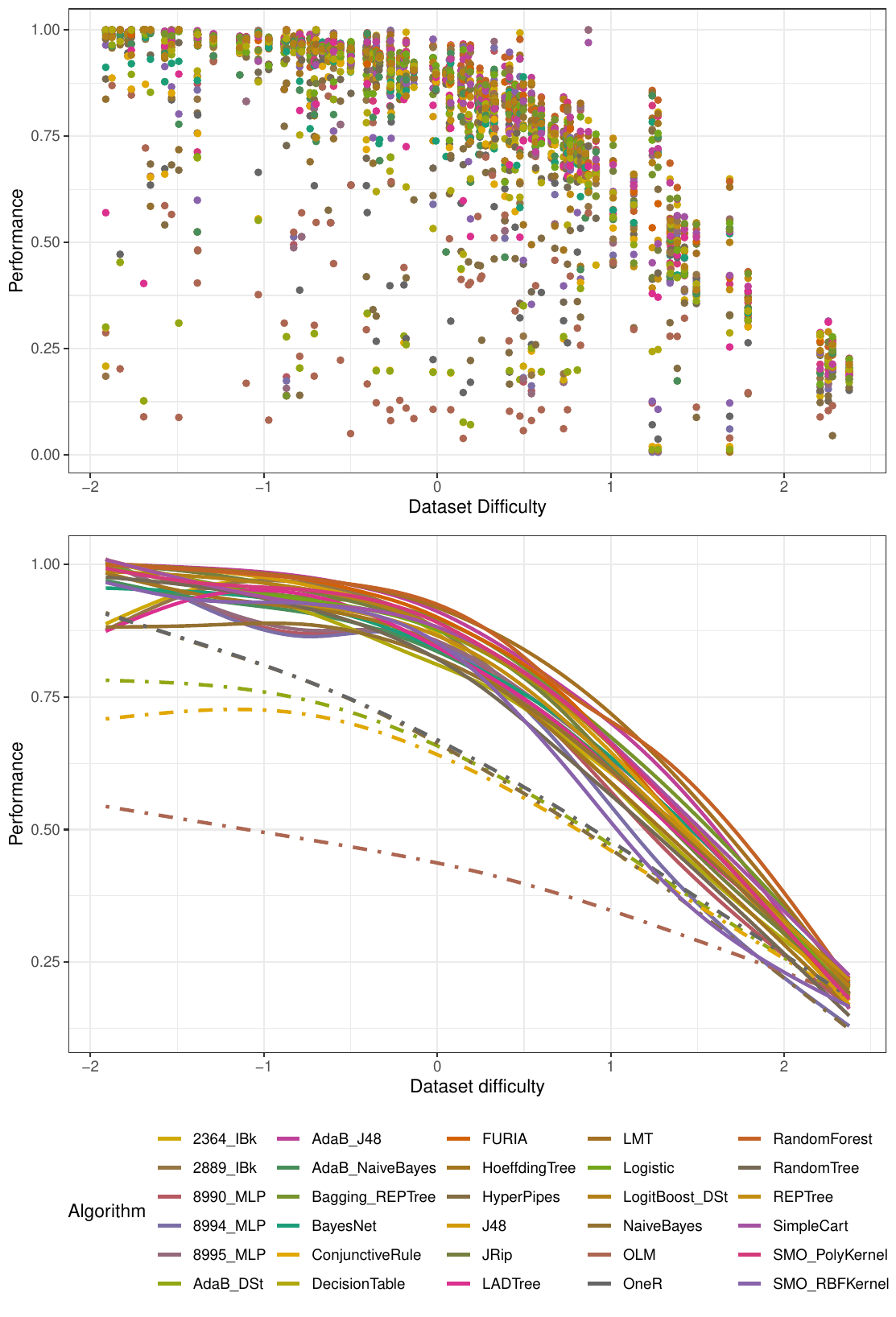}
   \caption{ Algorithm performance with dataset/problem difficulty for classification algorithms. Top: Algorithm performance against dataset difficulty. Bottom: Latent trait curves for each algorithm with AdaB\_DSt, ConjunctiveRule, HyperPipes, OLM and OneR in dashed lines.}
    \label{fig:latentgraph}
\end{figure}

\begin{figure}[!ht]
    \centering
    \includegraphics[scale = 0.85]{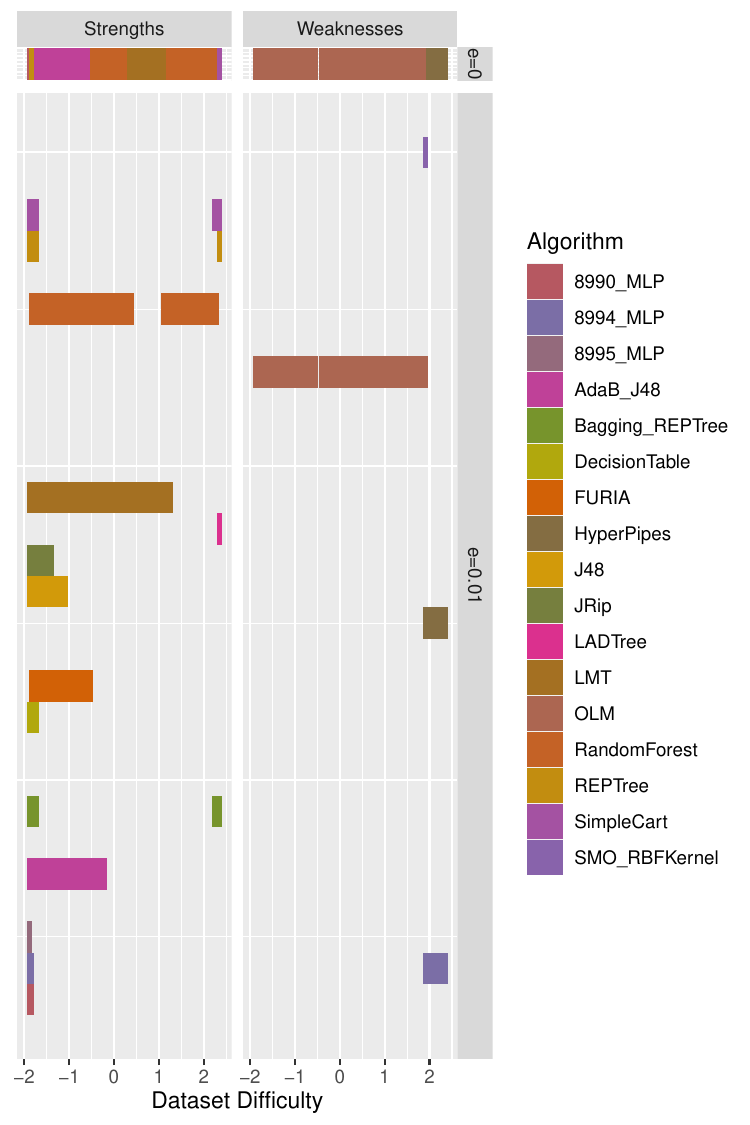}
   \caption{ Strengths and weaknesses of OpenML Weka classification algorithms. The top bar shows the strengths and weaknesses for $\epsilon = 0$ and the bottom graph for $\epsilon = 0.01$.}
    \label{fig:latentgraph1}
\end{figure}

Figures~\ref{fig:latentgraph} and \ref{fig:latentgraph1}  show the latent trait analysis for OpenML Weka classification algorithms. Figure~\ref{fig:latentgraph} shows the performance of the algorithms with respect to problem difficulty and the resulting smoothing splines. The strengths and weaknesses of different algorithms are shown in  Figure~\ref{fig:latentgraph1}. The strengths and weaknesses are calculated for two values of $\epsilon$, $\epsilon = 0$ and $\epsilon = 0.01$ as discussed in Section~\ref{sec:strengthsweaknesses}. 

Of the 30 algorithms, 6 have strengths on the dataset difficulty spectrum when $\epsilon = 0$. These are 8990\_MLP, AdaBoostM1\_J48, LMT, RandomForest, REPTree and SimpleCart algorithms. In contrast 14 algorithms exhibit strengths when $\epsilon = 0.01$ showing the competitiveness of algorithms. The RandomForest displays strengths on a large region of the problem space followed by LMT when $\epsilon = 0.01$. We see that many algorithms have strengths for easy problems while not so many are strong for difficult problems. For the region when dataset difficulty is between 0.5 and 1, only LMT displays a strength. Similarly, when dataset difficulty is between 1.5 and 2, the RandomForest is the only algorithm that displays an advantage. In terms of weaknesses, OLM is weak for most of the problem space for both $\epsilon$ values. Hyperpipes are weak for more difficult problems for both $\epsilon$ values.  The latent trait curves lying relatively below are shown in dashed lines so that they can be identified easier. These are AdaB\_DSt, ConjunctiveRule, HyperPipes, OLM and OneR.  

{\color{black} We can make some observations from Figure \ref{fig:latentgraph1} and Table \ref{tab:graphColuringairtmetrics}. The first is that the RandomForest and LMT cover almost all of the latent trait in the strengths diagram for $\epsilon = 0.01$. For $\epsilon = 0$, these two algorithms coupled with AdaB\_J48 cover most of the strengths spectrum. Thus, these three algorithms, or even just RandomForest and LMT make a good combination in tackling diverse datasets.  The second observation is that when increasing $\epsilon $ from 0 to 0.01, even though the number of algorithms increased from 6 to 14, most of them have strengths for very easy problems. Of the additional 8 algorithms, DecisionTable, 8994\_MLP, 8995\_MLP and LADTree have $\text{LTO} < 0.1$. Thus, we can disregard some of the algorithms with small LTO when $\epsilon = 0.01$. Considering the key algorithms, the main change from $\epsilon = 0$ to $\epsilon = 0.01$ is the increase in LTO for algorithm LMT.        }

\subsubsection{AIRT model goodness metrics}
\begin{figure}[!ht]
    \centering
     \includegraphics[scale=0.7]{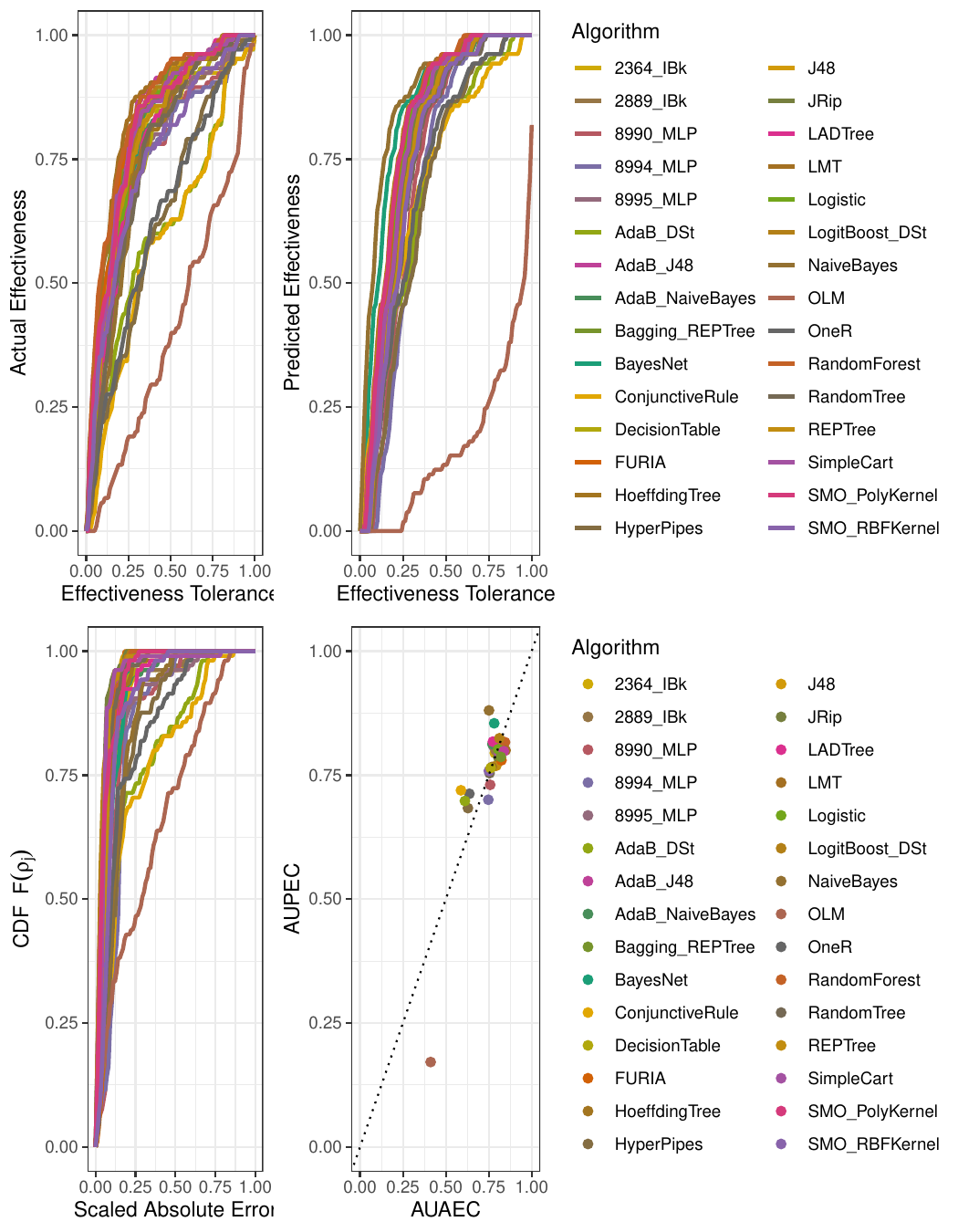}
    \caption{ Model goodness metrics for classification algorithms: the actual and predicted effectiveness curves on the top row and the CDFs  $F(\rho_j)$ and (AUAEC, AUPEC) on the bottom row.}
    \label{fig:latentgraph2}
\end{figure}

Table~\ref{tab:graphColuringAccuracyReliabilityCTS} gives the model goodness results for classification algorithms. The MSE is less than 0.1 for all algorithms apart from OLM. Furthermore, the difference between predicted and actual effectiveness $|\text{AUPEC} - \text{AUAEC}|$ is less than 0.1 for all algorithms apart from OLM, NaiveBayes and ConjunctiveRule.  Figure~\ref{fig:latentgraph2} shows the effectiveness curves and the CDFs for this portfolio of algorithms. We see that most points on the AUAEC-AUPEC plane are close to the AUAEC = AUPEC line, which is shown by a dotted line. OLM is the exception. In general, the model has fitted the algorithm performances well.
\begin{table}[!ht]
		\centering
		\caption{ MSE, AUCDF, Area Under Actual Effectiveness Curve  (AUAEC) and Predicted Effectiveness Curves (AUPEC) and |AUAEC - AUPEC| for classification algorithms. }
  {
		\begin{tabular}{lrrrrr}
			\toprule
		Algorithm  & MSE & AUCDF & AUAEC & AUPEC  & |AUAEC - AUPEC|  \\
			\midrule

         8990\_MLP & 0.032 & 0.863  & 0.758 &  0.730 & 0.028 \\
         8994\_MLP & 0.036 & 0.844  & 0.747 &  0.700 &  0.047 \\
         8995\_MLP & 0.029 & 0.899  & 0.776 &  0.809 & 0.033 \\
   SMO\_PolyKernel & 0.007 & 0.940  & 0.814  & 0.820 & 0.006 \\
             OneR & 0.051 & 0.840  & 0.637  & 0.712 & 0.075 \\
              J48 & 0.004 & 0.944  & 0.810 &  0.782 & 0.028 \\
         2364\_IBk & 0.022 & 0.910 &  0.785 &  0.795 & 0.010 \\
          REPTree & 0.010 & 0.932  & 0.794 &  0.769 & 0.025 \\
       RandomTree & 0.007 & 0.939  & 0.754  &  0.754 & 0.000 \\
    RandomForest & 0.006 & 0.938 &  0.845 &  0.816 & 0.029 \\
             LMT & 0.007 & 0.928 &  0.848 &  0.800 & 0.048 \\
   HoeffdingTree & 0.011 & 0.921 &  0.768  & 0.767 & 0.001 \\
   SMO\_RBFKernel & 0.017 & 0.899 &  0.750 &  0.759 & 0.009 \\
            JRip & 0.003 & 0.950 &  0.805 &  0.787 & 0.018 \\
        2889\_IBk & 0.020 & 0.920 &  0.789 &  0.800 & 0.011 \\
      HyperPipes & 0.040 & 0.846  & 0.629 &  0.683 &  0.054 \\
      NaiveBayes & 0.029 & 0.868 &  0.751 &  0.881  & 0.130 \\
             OLM & 0.164 & 0.681  & 0.411  & 0.171 &  0.240\\
           FURIA & 0.006 & 0.929 &  0.824  & 0.780  & 0.044 \\
        BayesNet & 0.011 & 0.920  & 0.782 &  0.855  & 0.073 \\
 ConjunctiveRule & 0.093 & 0.784 &  0.588  & 0.719 & 0.131 \\
      SimpleCart & 0.009 & 0.943  & 0.811 &  0.794  & 0.017 \\
 AdaBoostM1\_NaiveBayes & 0.013 &  0.928 &  0.771  & 0.813 & 0.042\\
         LADTree & 0.012 & 0.938  & 0.774  & 0.818 & 0.044 \\
        Logistic & 0.005 & 0.942 &  0.805 &  0.802  &  0.003\\
        AdaBoostM!\_DSt & 0.082 &  0.794  & 0.611 &  0.698  & 0.087 \\
        AdaBoostM1\_J48 & 0.006 &  0.934  &  0.837  & 0.798  & 0.039 \\
 Bagging\_REPTree & 0.011 & 0.929  & 0.820  & 0.787  & 0.033 \\
   DecisionTable & 0.009 & 0.931  &  0.761 &  0.764  & 0.003 \\
  LogitBoost\_DSt & 0.004 & 0.956 &  0.812  & 0.824  & 0.012\\
		    \bottomrule
		\end{tabular}
  }
		\label{tab:graphColuringAccuracyReliabilityCTS}
\end{table}
    
\subsubsection{Algorithm portfolio selection}\label{sec:basketeval}
We compare the airt portfolio with 2 additional algorithm portfolios: 
\begin{enumerate} 
\item \textit{Shapley-portfolio}:  a subset of algorithms selected using Shapley values \citep{Kevin2016}. Shapley values measure an algorithm's marginal contribution to the portfolio by using concepts from coalition game theory.  For Shapley-portfolio we select algorithms with the top-$n$ Shapley values.  
\item \textit{topset-portfolio}: a subset of algorithms having the best on-average performance at a per-instance level. The highest-ranked algorithm in the topset-portfolio gives the best performance for the most number of instances. For topset-portfolio we select the top-$n$ best on-average algorithms
\end{enumerate} 

We construct Shapley, topset and airt portfolios with $n$ algorithms and compare their performance for different values of $n$.  As the evaluation metric we use the performance gap. Performance gap is computed using the best per-instance performance for each portfolio and the best per-instance performance using all the algorithms. We define the difference as the performance gap at a per-instance level. Let the best performance for instance $i$ using the full set of algorithms be denoted by $b_i$. Let $\mathscr{A}_n, \mathscr{S}_n$ and $\mathscr{T}_n$ denote airt, Shapley and topset portfolios having $n$ algorithms. Let $b_{\mathscr{A},i,n}$ denote the best performance for instance $i$ using the airt portfolio with $n$ algorithms. Similarly, let $b_{\mathscr{S},i,n}$ and $b_{\mathscr{T},i,n}$ denote the best performance for instance $i$ using Shapley and topset portfolios with $n$ algorithms. Then we define the performance gap for instance $i$ for each portfolio as 
\begin{equation}\label{eq:perfdiff}
    \text{Perf. gap}_{\mathscr{A}, i, n} = b_i -b_{\mathscr{A},{i},n} \, , \quad \text{Perf. gap}_{\mathscr{S},i,n} = b_i -b_{\mathscr{S},i,n}\, \quad \text{and} \quad \text{Perf. gap}_{\mathscr{T},i,n} = b_i -b_{\mathscr{T},i,n} \, .
\end{equation}
For each algorithm portfolio and $n$ we get an $N \times 1$ vector of performance gap values. We compute the mean performance gap for each $n$. For each algorithm scenario we use 10-fold cross validation and report the average cross validated performance gap for Shapley, topset and airt portfolios. Additionally, we compute the standard errors using different folds.  {\color{black} We note that \textit{Perf. gap} is the same as \textit{misclassification penalty} discussed in \cite{Bischl2016}. However, we have used the term \textit{Perf. gap} because we think it is more intuitive and applicable to non-classification scenarios.}

\begin{figure}[!ht]
    \centering
    \includegraphics[scale=0.7]{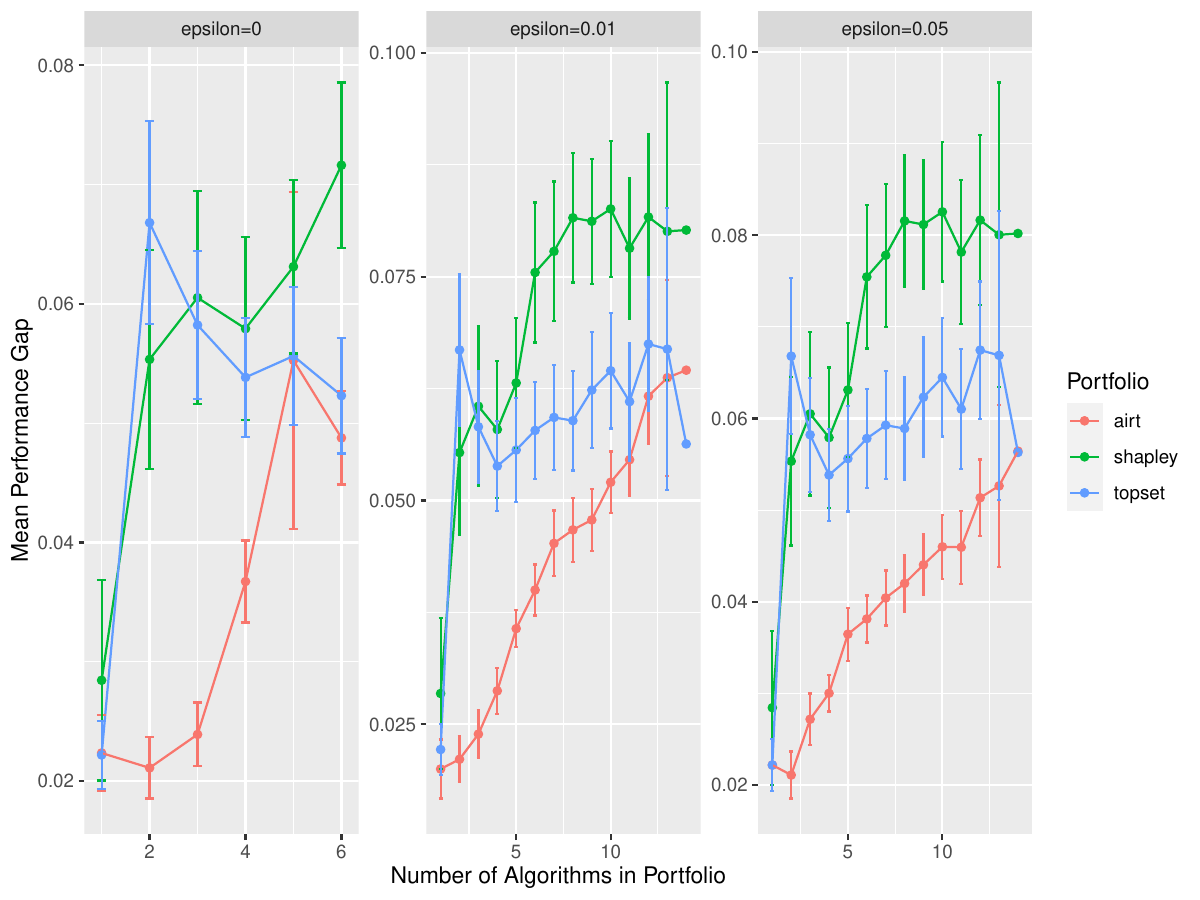}
    \caption{  Performance analysis of Shapley, topset and airt portfolios for different $\epsilon$ values. The mean cross-validated performance gap is shown with standard errors denoted by vertical lines.    }
    \label{fig:perfCurves}   
 \end{figure}

Figure~\ref{fig:perfCurves} shows the mean performance gap of the 3  portfolios using 10-fold cross validation for OpenML Weka algorithms for different values of $\epsilon$. A lower gap is preferred as it indicates the portfolio has better algorithms. The vertical lines at each point show the standard errors. We see that airt generally has lower performance gaps. The number of algorithms in the airt portfolio changes with $\epsilon$. For each $\epsilon$, as the limiting number of algorithms (the maximum $x$ value) we select the minimum number of algorithms from airt, Shapley and topset. For $\epsilon = 0$ airt selects 6 algorithms, which decides the limiting number of algorithms. For other $\epsilon$ values Shapley decides the limiting number of algorithms in this example. For each fold, different algorithms may get selected by different portfolio selection methods. Thus, for $n = 14$ standard errors are not computed because 14 algorithms are selected only in 1 fold.

\subsection{Additional case studies}

\begin{table}[!ht]
		\centering
		\caption{  Additional ASlib case studies. Mean Performance Gap (MPG) of a portfolio of 5 algorithms is reported using 10-fold CV with the best in bold.  }
  {
		\begin{tabular}{lp{2cm}p{1.5cm}p{1.5cm}p{1.5cm}p{1.5cm}p{1.5cm}}
			\toprule
		Scenario & Measurement &  Num. Obs. & Num. Algorithms &   airt MPG & Shapley MPG  & topset MPG \\
			\midrule
         OPENML\_WEKA & accuracy & 105 & 31 & \textbf{0.0553} & 0.0631 & 0.0556 \\
         ASP\_POTASSCO &  runtime & 1294 & 11 & 78.0 & 92.7 & \textbf{77.8}\\
         CSP\_MINIZINC\_2016 & par10 & 100 & 21 &  \textbf{1962} & 2371 & 2026  \\
         GRAPHS\_2015  & runtime & 5725 & 8 & \textbf{1689346} & 6127229 &  6763210 \\
         MAXSAT\_PMS\_2016 & par10 & 601 & 20 & \textbf{1019} & 1469 & 1305 \\
         PROTEUS\_2014 & runtime & 4021 & 23 & \textbf{293} & 648 & 1125 \\
         SAT11\_INDU & runtime & 300 & 19 & 882 & \textbf{826} & 855 \\
         SAT12\_ALL & runtime & 1614 & 32 & \textbf{456} & 523 & 683 \\
         SAT18\_EXP\_ALGO & runtime & 353 & 38 & \textbf{1677} & 1823 & 1822 \\
         BNSL\_2016 & runtime & 1179 & 9 & \textbf{1210} & 1448 & 2030 \\
        \bottomrule 
        \end{tabular}
   }
   \label{tab:aslibsummary}
\end{table}

We conduct shorter analyses for 9 additional ASlib scenarios, which are given in Appendix~\ref{sec:App1}. We explore the latent trait curves, strengths and weaknesses of algorithms for $\epsilon \in \{0, 0.05\}$ and compare different algorithm portfolios. As each scenario other than OPENML-Weka has  runtimes or par10 values as the evaluation metric, we  transform these values by multiplying with -1 and scaling to the interval $[0,1]$ with 1 denoting good performance and 0 denoting poor performance. Some summary statistics of these analyses are given in Table \ref{tab:aslibsummary}. As it is difficult to encapsulate the strengths and weaknesses or the latent trait curves by a single numeric value, we give the mean performance gap of the  portfolios with 5 algorithms in  Table \ref{tab:aslibsummary}. Figures of the mean performance gap  for different number of algorithms with standard errors and other details are given in the Appendix. 

For most scenarios airt performs well. Even though airt does not perform well for SAT11\_INDU, from the MPG curves in the Appendix we see that the standard errors of the different portfolios overlap.  We also see that the latent trait curves for SAT11\_INDU are all bundled up together. This tells us that SAT11\_INDU algorithms are similar in performance.  From other scenarios, we notice that airt is better at identifying a good portfolio of algorithms when algorithms are diverse, i.e., when the latent trait curves display high variability. The construction of the latent trait $\theta$ involves IRT discrimination and difficulty parameters as well as the actual performance. Therefore, selecting algorithms based on fitting splines to $\theta$ takes into account this underlying hidden quantity uncovered by IRT that denotes the dataset difficulty spectrum. This allows us to select a good portfolio of algorithms when a diverse set of algorithms are present. This is another use of AIRT in addition to its exploratory aspect.

% =======================================================================
\section{Conclusions}\label{sec:conclusions}
Beyond standard statistical analysis, which often hides useful insights, there are not many techniques  that can be used to rigorously evaluate a portfolio of algorithms and identify their strengths and weaknesses. One such technique is the instance space analysis methodology which can be used to visualize the strengths and weaknesses of algorithms. As the instance space incorporates both the algorithms and the test instances, computing features of test instances is an essential step to constructing an instance space. Devising suitable features of test instances that capture their intrinsic difficulties for algorithms is a significant challenge that can limit the applicability of the method. In this paper we have taken a different approach to achieve the same goal that avoids the need to devise instance features.  We have presented AIRT, an IRT based algorithm evaluation method, which evaluates algorithms using only performance results. We demonstrated its usefulness on a diverse set of algorithm portfolios arising from a wide variety of problem domains.  The scenarios used are taken from the ASlib repository containing algorithm implementations with given parameter and hyperparameter settings. We have not explored different parameter settings in this study and this is a limitation. Each parameter setting would give rise to a different algorithm implementation that would result in a different algorithm curve. Thus, by considering a single algorithm with different parameter settings, AIRT has potential to select parameter settings that are advantageous for easy or difficult problems.

Recasting the IRT framework as an inverted model, AIRT focuses on evaluating  algorithm attributes such as consistency, anomalousness and difficulty limit thereby helping to broaden the understanding of algorithm behaviors and their dependence on test instances. AIRT can be used to visualize the strengths and weaknesses of algorithms in different parts the problem space. Using algorithms with strengths we construct an algorithm portfolio and show that it has a low performance gap compared to other portfolios. In addition, IRT model goodness measures can be derived, showing the level of trustworthiness of the underlying IRT model. Due to the fact that AIRT extends the IRT framework, it also has the desirable mathematical and optimality properties inherited from the embedded maximum likelihood estimation techniques.  Furthermore, the explainable nature of IRT parameters gets translated to the algorithm evaluation domain. 

As future research avenues we plan to consider the role of AIRT in parameter selection and  alternative remappings of the IRT framework to increase understanding of the strengths and weaknesses of dataset repositories, thereby providing means to select an unbiased yet diverse collection of datasets,  drawing deeper insights into their abilities to support meaningful conclusions about algorithm strengths and weaknesses.

\subsection*{Acknowledgments}
%\acks{\textbf{Acknowledgements}}
% =======================================================================

Funding was provided by the Australian Research Council through the Australian Laureate Fellowship FL140100012, and the ARC Training Centre in Optimisation Technologies, Integrated Methodologies and Applications (OPTIMA) under grant IC200100009. The  authors would like to thank Prof Rob J. Hyndman for his suggestion of the name AIRT for our method. 
% =======================================================================
\subsection*{Supplementary Material}
The algorithm performance datasets used in this paper are found at \url{https://github.com/coseal/aslib_data} and the programming scripts using AIRT are available at \\ \url{https://github.com/sevvandi/airt-scripts}. 
% =======================================================================

%\pagebreak
\bibliographystyle{agsm}
\bibliography{citations}
\appendix 
\section{ASlib scenarios}\label{sec:App1}
% \subsection{ASlib scenarios}
 In this section we explore 9 ASlib scenarios: ASP-POTASSCO, CSP-MiniZinc-Time-2016, GRAPHS-2015, MAXSAT-PMS-2016,  PROTEUS-2014, SAT11-INDU, SAT12-ALL, BNSL-2016 and SAT18-EXP-ALGO. For each scenario we fit an AIRT model and conduct a smaller analysis compared to the OpenML-Weka example in Section~\ref{subsec:openml}. Using the fitted model, we plot the latent trait curves. Then we compute the strengths and weaknesses of algorithms on the dataset difficulty spectrum for  $\epsilon = 0$  and $\epsilon = 0.05$. By visualizing this spectrum, we see which algorithms have strengths for easy problems and which ones are better suited for difficult problems. Similarly, we see their weaknesses as well. Using 10-fold cross validation, we evaluate airt, topset and Shapley algorithm portfolios and examine the mean performance gap as explained previously.     

\begin{figure}[p]
    \centering
    \includegraphics[scale=0.65]{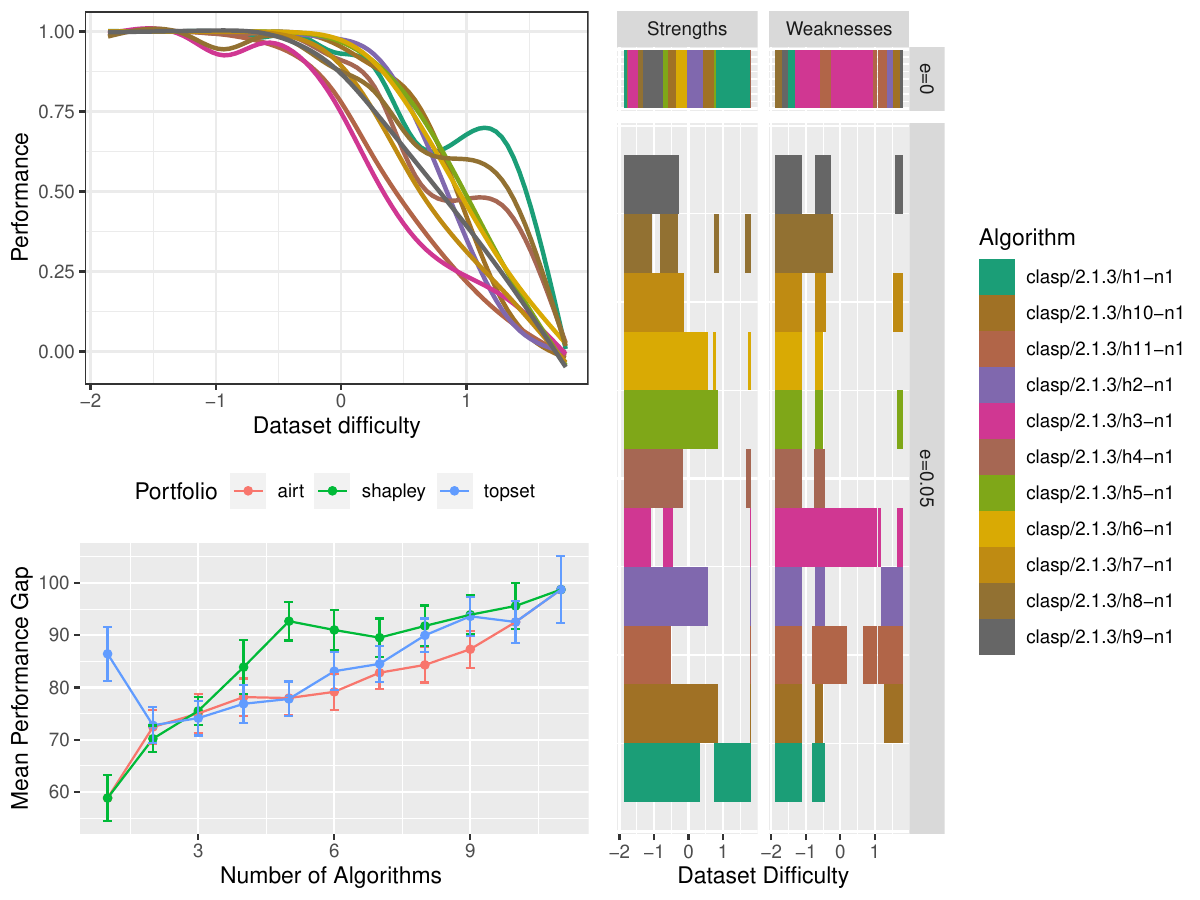}
   \caption{ Latent trait curves, strengths and weaknesses and 10-fold CV portfolio comparison for ASP\_POTASSCO scenario. }
    \label{fig:asppoassco}
    \includegraphics[scale=0.65]{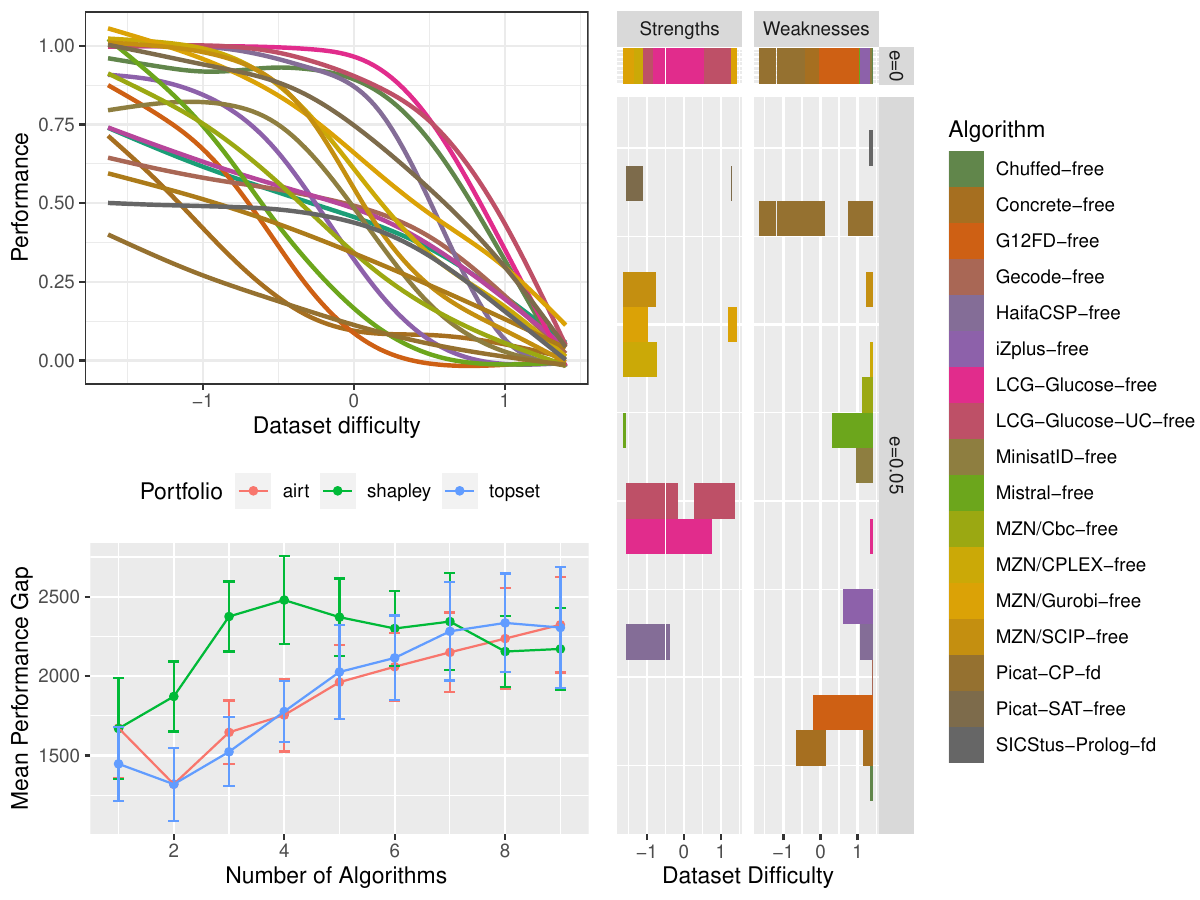}
   \caption{ Latent trait curves, strengths and weaknesses and 10-fold CV portfolio comparison for CSP\_MiniZinc scenario. }
    \label{fig:cspminizinc}
\end{figure}

\subsubsection{ASP\_POTASSCO}
Figure~\ref{fig:asppoassco} shows the analysis for ASP\_POTASSCO scenario. Algorithm \textit{clasp/2.1.3/h3-n1} is the weakest in the portfolio as we can see from the strengths and weaknesses figure and the latent trait curves. Algorithm \textit{clasp/2.1.3/h1-n1} is better suited for difficult problems as seen by the hump in the latent trait curve around $ \delta \approx  1 $. Many algorithms perform well for very easy problems as seen by the leftmost part of the strengths in the problem difficulty spectrum. {\color{black} A point of interest about these curves is that some curves, including that of  \textit{clasp/2.1.3/h1-n1}, have a turning point around  $ \delta \approx 0.75$ followed by a positive slope, signifying an improvement in performance, culminating at $ \delta \approx 1.25$ before decreasing again. This shows locally anomalous behavior for $\delta$ in that region for certain algorithms.} The cross-validated mean performance gap of different algorithm portfolios show that   for $n \in \{1, \ldots, 5\}$ airt is similar to either Shapley or topset, but for  $n \in \{6, \ldots, 9\}$ airt has a lower mean performance gap. However, the standard errors show that the differences are not significant. 

\subsubsection{CSP\_MiniZinc\_2016}
Figure~\ref{fig:cspminizinc} shows the analysis for CSP\_MiniZinc\_2016. {\color{black} The latent trait curves are spread out well and thus show high variability. This has resulted in a sparse set of strengths and weaknesses. Even though there are many algorithms, only a few exhibit strengths and similarly only a few have weaknesses at other places apart from the rightmost end, which has the most difficult problems.} As seen from the latent trait curves and the strengths and weaknesses figure, algorithm \textit{LCG-Glucose-UC-free} shows continued strength for difficult and semi-difficult problems. Algorithm \textit{MZN/Gurobi-free} is better suited for easy and very difficult problems. The weakest algorithm is \textit{Picat-CP-fd}, which is weak for easy and semi-difficult problems. While many algorithms are good for easy problems, both \textit{LCG-Glucose-UC-free} and \textit{LCG-Glucose-free} displays strengths for a large region of the problem space for $\epsilon = 0.05$.  The cross-validated mean performance gap graphs show that airt and topset behave similarly, while Shapley has higher mean performance gaps initially but converges with airt and topset for higher $n$.

\subsubsection{Graphs\_2015}
Figure~\ref{fig:graphs} shows the analysis for Graphs\_2015. From the latent trait curves and the strengths and weaknesses figure we see that \textit{glasgow2} and  \textit{glasgow3} are suited for a large part of the  problem space. \textit{supplemantallad} is good for easy and very difficult problems and many algorithms have strengths for easy problems. The weakest algorithm is \textit{vf2} as seen from the weaknesses spectrum. {\color{black} For dataset difficulty $\delta \lessapprox 0.5$, all algorithms apart from \textit{vf2} perform well. However, after that point, the algorithms diverge in their performance as seen from the curves.} The cross-validated mean performance gap of different portfolios show that airt has the smallest performance gap for most $n$.

\subsubsection{MAXSAT-PMS-2016}
\begin{figure}[!p]
    \centering
    \includegraphics[scale=0.65]{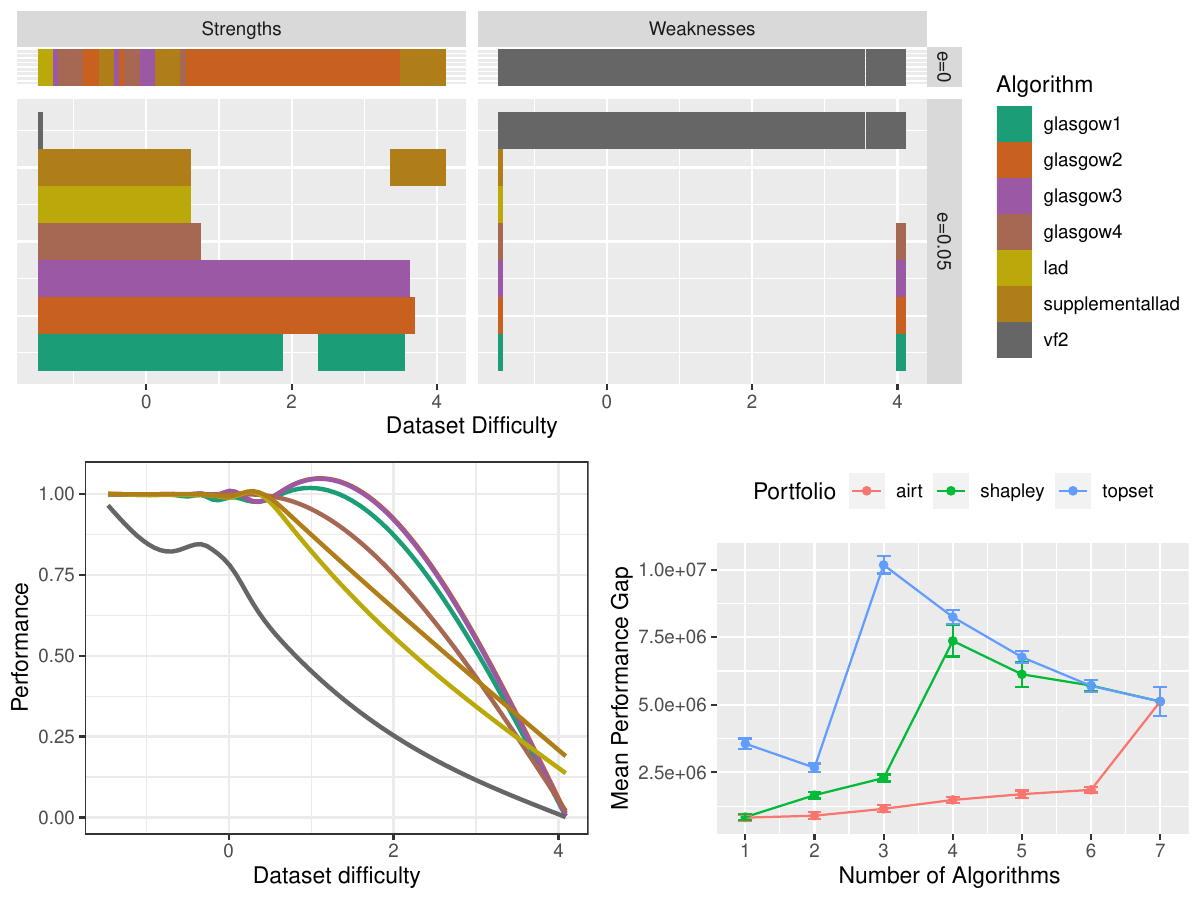}
    \caption{ Strengths and weaknesses, latent trait curves and 10-fold CV portfolio comparison for Graphs\_2015 scenario. }
    \label{fig:graphs}
    \includegraphics[scale=0.65]{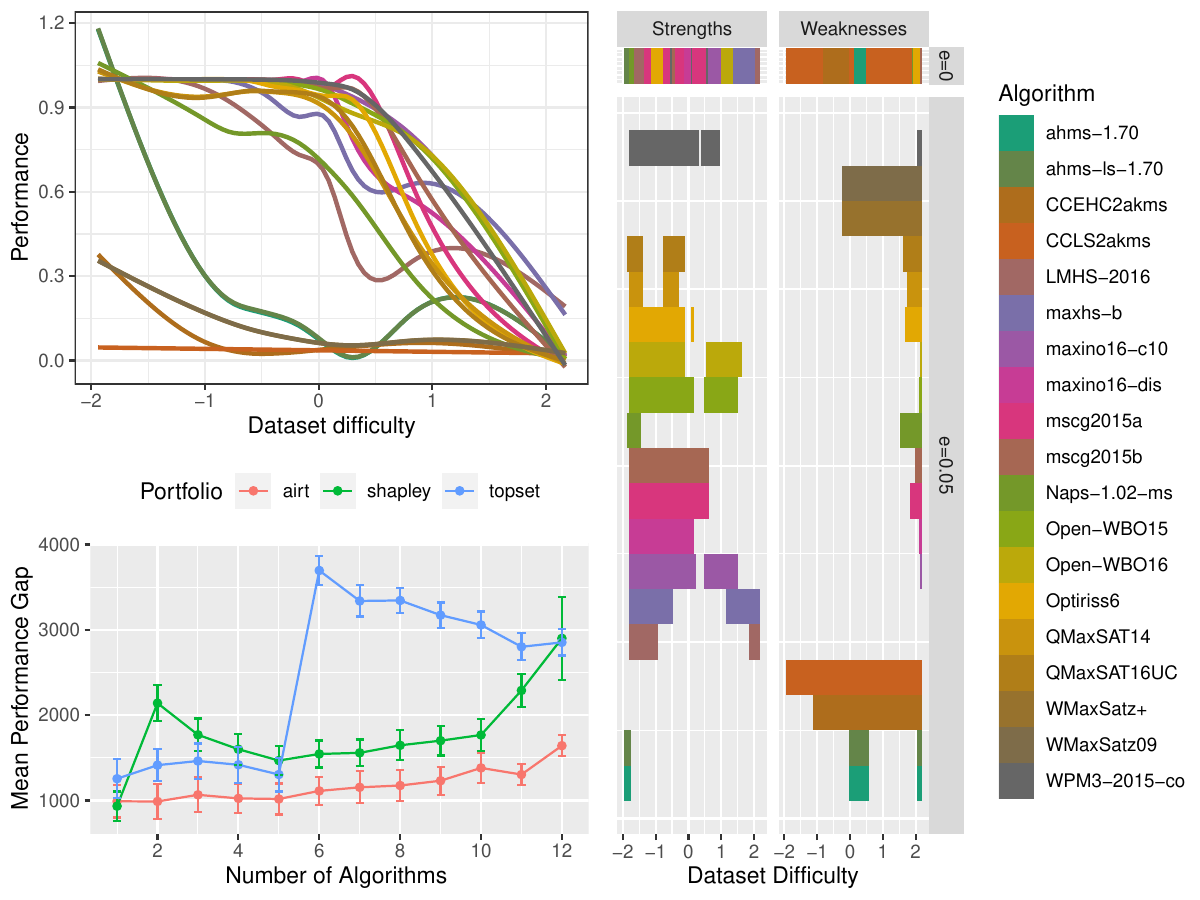}
   \caption{ Strengths and weaknesses, latent trait curves and 10-fold CV portfolio comparison for MAXSAT\_PMS\_2016 scenario. }
    \label{fig:maxsatpms}
\end{figure}

Figure~\ref{fig:maxsatpms} shows the analysis for MaxSAT-PMS-2016 scenario. Immediately we see variety in the latent trait curves. Some curves have low performance values for most part of the space, which is different from the other scenarios we examined so far. Some curves have varying behavior with curved sections. In the strengths diagram, we see many algorithms having strengths for easier problems.  {\color{black} Of the algorithms, 15 have strengths for dataset difficulty $\delta \leq 0$ when $\epsilon = 0.05$. These are the easy problems. For the easy problems, any of these algorithms would give good performances. Only 8 algorithms have strengths for $0 \leq \delta \leq 1$ and of these only 5 have strengths for $\delta > 1$.} \textit{LMHS-2016} and \textit{maxhs-b} are better suited for harder problems. In the weaknesses space we see that \textit{CCLS2akms} and \textit{CCEHC2akms} are very weak algorithms. The cross-validated mean performance gap shows that airt performs better compared to the other two portfolios. The topset portfolio has a sudden jump at $n = 6$, possibly due to including a volatile algorithm, which gets mitigated with subsequent algorithm additions to the portfolio.

\subsubsection{PROTEUS-2014}
The latent curves of PROTUES-2014, shown in Figure~\ref{fig:proteus} have many wiggles. {\color{black} Four curves achieve local minima at dataset difficulty $\delta \approx -0.2$. After that point, their performance increase for some part of the dataset difficulty spectrum, i.e., as the dataset difficulty increases, the performance of these algorithms get better. Thus, these algorithms are locally anomalous. They are not anomalous throughout the spectrum, but they have regions of locally anomalous behaviour.} Algorithms \textit{claspcnf\_support, claspcnf\_direct} and \textit{claspcnf\_directorder} display strengths for a large part of the problem space including difficult problems. Algorithm \textit{gecode} is the weakest algorithm as seen by the latent trait curves and the strengths and weaknesses diagram. The cross-validated mean performance gap curves show that airt performs better than the other two portfolios. The standard errors for both topset and airt are very low making them not clearly visible in the diagram.

\subsubsection{SAT11-INDU}
\begin{figure}[!p]
    \centering
    \includegraphics[scale=0.55]{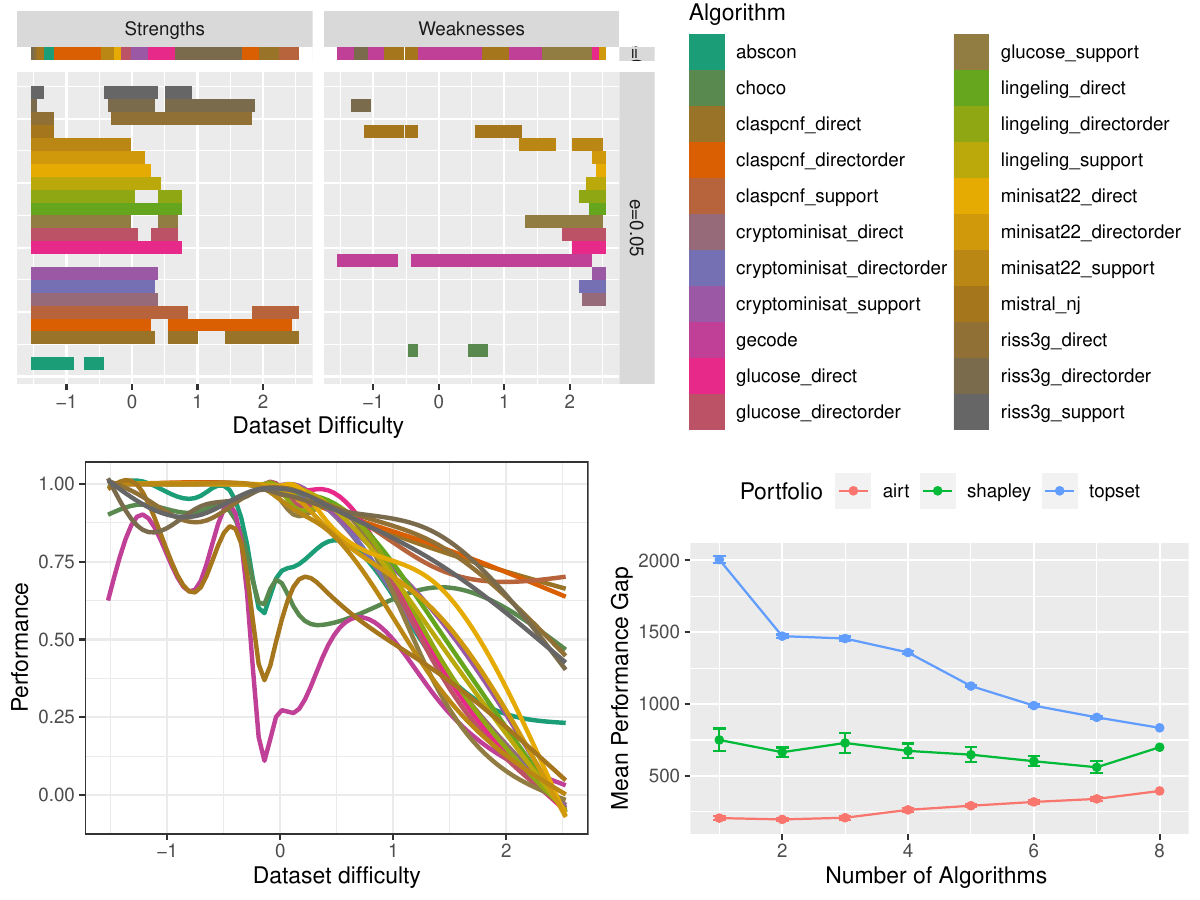}
    \caption{ Strengths and weaknesses, latent trait curves and 10-fold CV portfolio comparison for PROTEUS\_2014 scenario.  }
    \label{fig:proteus}
     \includegraphics[scale=0.55]{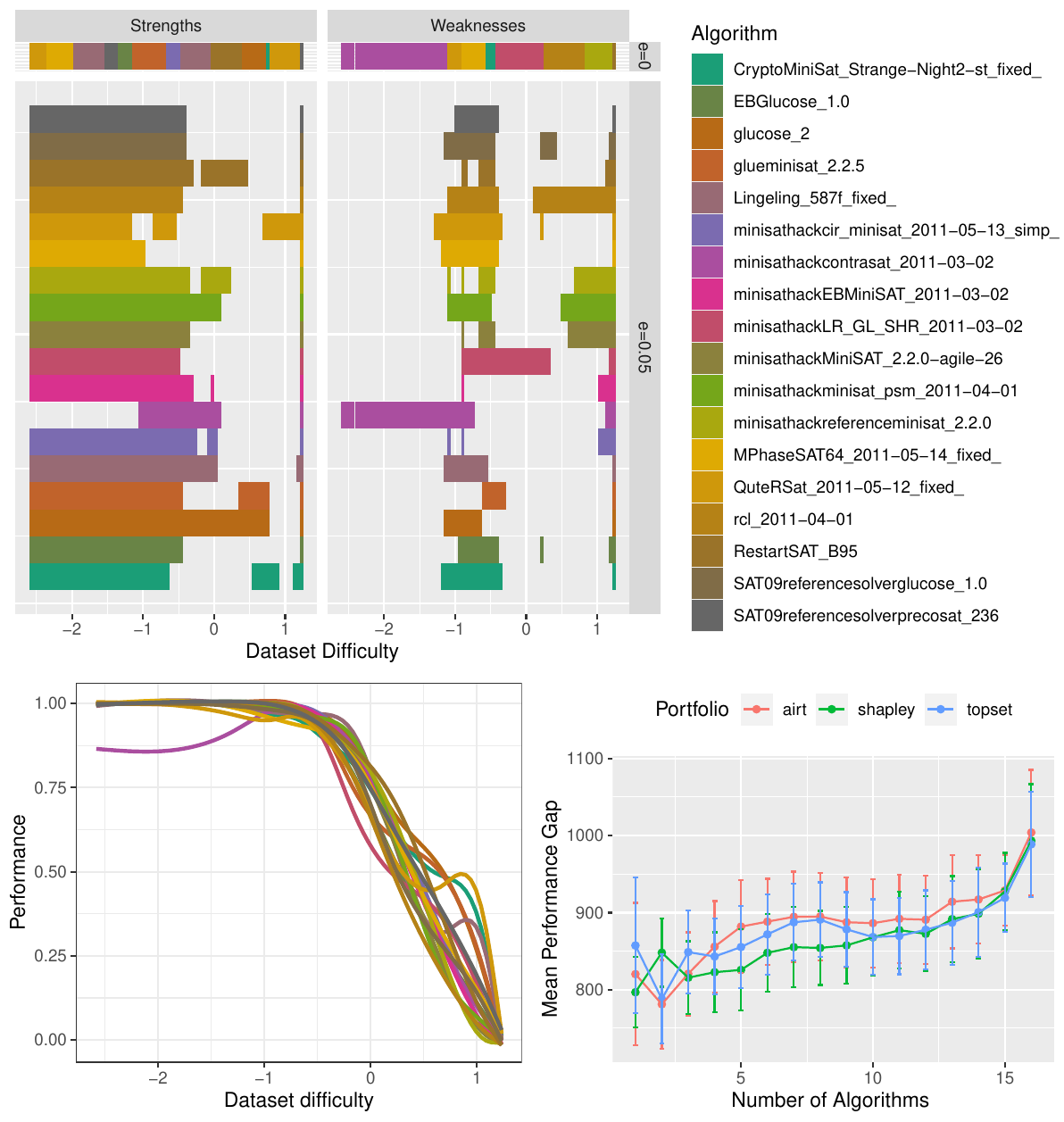}
    \caption{ Strengths and weaknesses, latent trait curves and 10-fold CV portfolio comparison for SAT11\_INDU scenario..}
    \label{fig:sat11indu}
\end{figure}

Figure~\ref{fig:sat11indu} shows the analysis of  SAT11-INDU. We see that most algorithms have similar-shaped latent trait curves. {\color{black} We do not know if the algorithms were preselected, which might account for this behaviour. The similarity of the curves implies some similarity of performance between the algorithms. } In the strengths diagram many algorithms have strengths for easy and semi-difficult problems. In the weaknesses diagram, we see a curious occurrence: many algorithms display weaknesses in the middle of the spectrum as well as on the difficult end of the spectrum. This is because the curves are packed together for most part of the problem space. Algorithm \textit{glucose\_2} occupies the highest proportion of the latent trait. From the strengths and weaknesses figure we see that \textit{QuteRSat\_2011-05-12\_fixed} is strong for difficult problems. Notably \textit{minisathackcontrasat\_2011-03-02} is weak for easy problems.  The cross-validated performance gap curves show that Shapley performs better than the others, but the standard errors of the 3 portfolios overlap for most values of $n$.

\subsubsection{ SAT12-ALL}

SAT12-ALL scenario contains SATzilla 2012 competition \citep{Xu2012} results on algorithm performance. Figure \ref{fig:sat12} shows the latent trait curves, strengths and weaknesses and performance comparison of different portfolios. The curves have diverse characteristics: some curves have an initial downward trend showing that they are weak for most part of the space but later trend upward indicating that they perform better for more difficult test instances. {\color{black} Another set of curves give good performances for easy problems with $\delta \lessapprox -1$ and decrease in performance after that.} Algorithms \textit{mphaseSATm} and \textit{mphaseSAT} are strong for a large part of the problem space including difficult instances. Algorithms \textit{spear-sw} and \textit{eagleup} are weak for most parts of the space. The cross-validated mean performance gap curves show that airt has a lower gap compared to the other 2 portfolios.

\subsubsection{ BNSL-2016}
\begin{figure}[!p]
    \centering
    \includegraphics[scale=0.55]{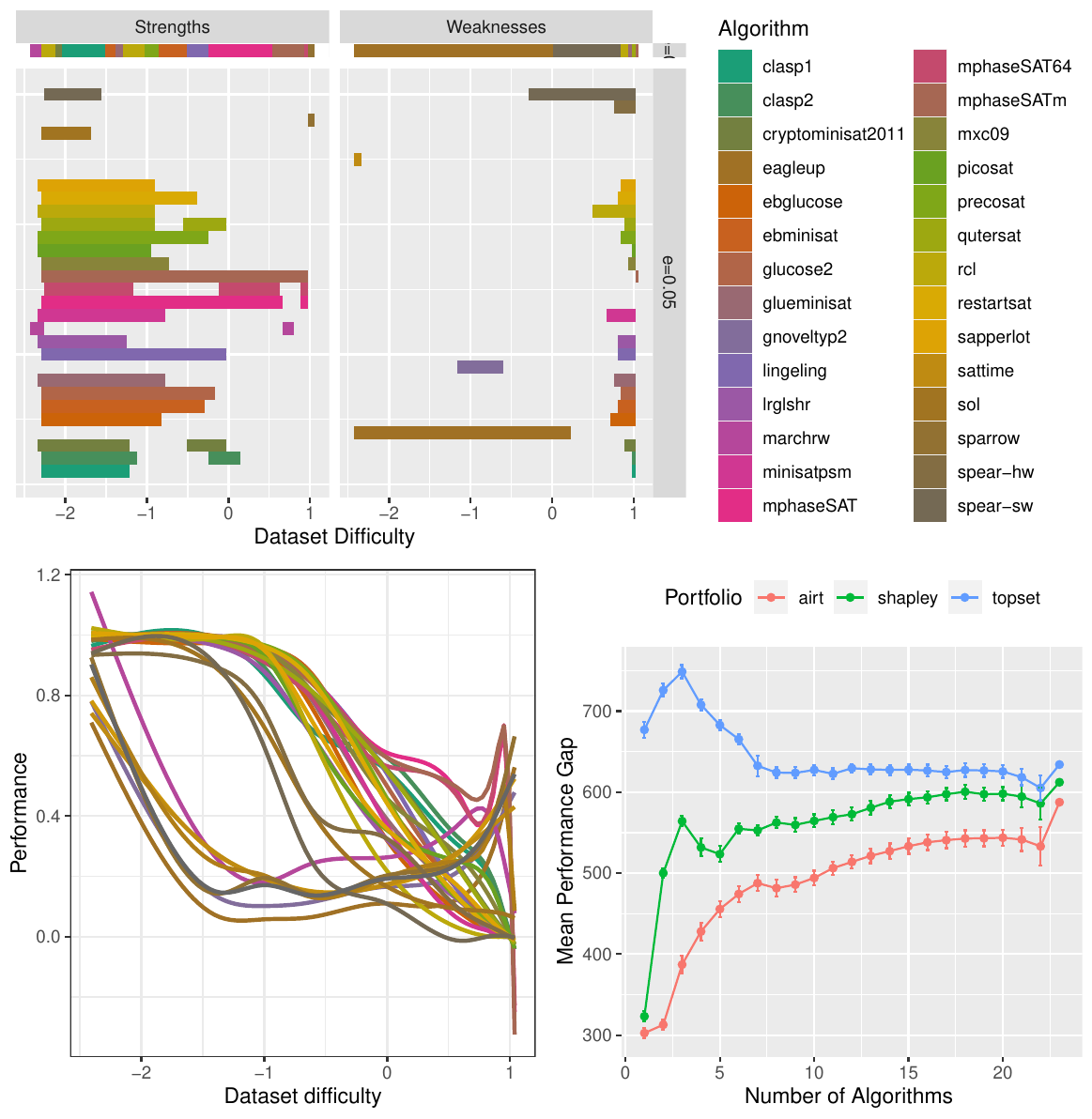}
    \caption{ Strengths and weaknesses, latent trait curves and 10-fold CV portfolio comparison for SAT12\_ALL scenario.  }
    \label{fig:sat12}
    \includegraphics[scale=0.55]{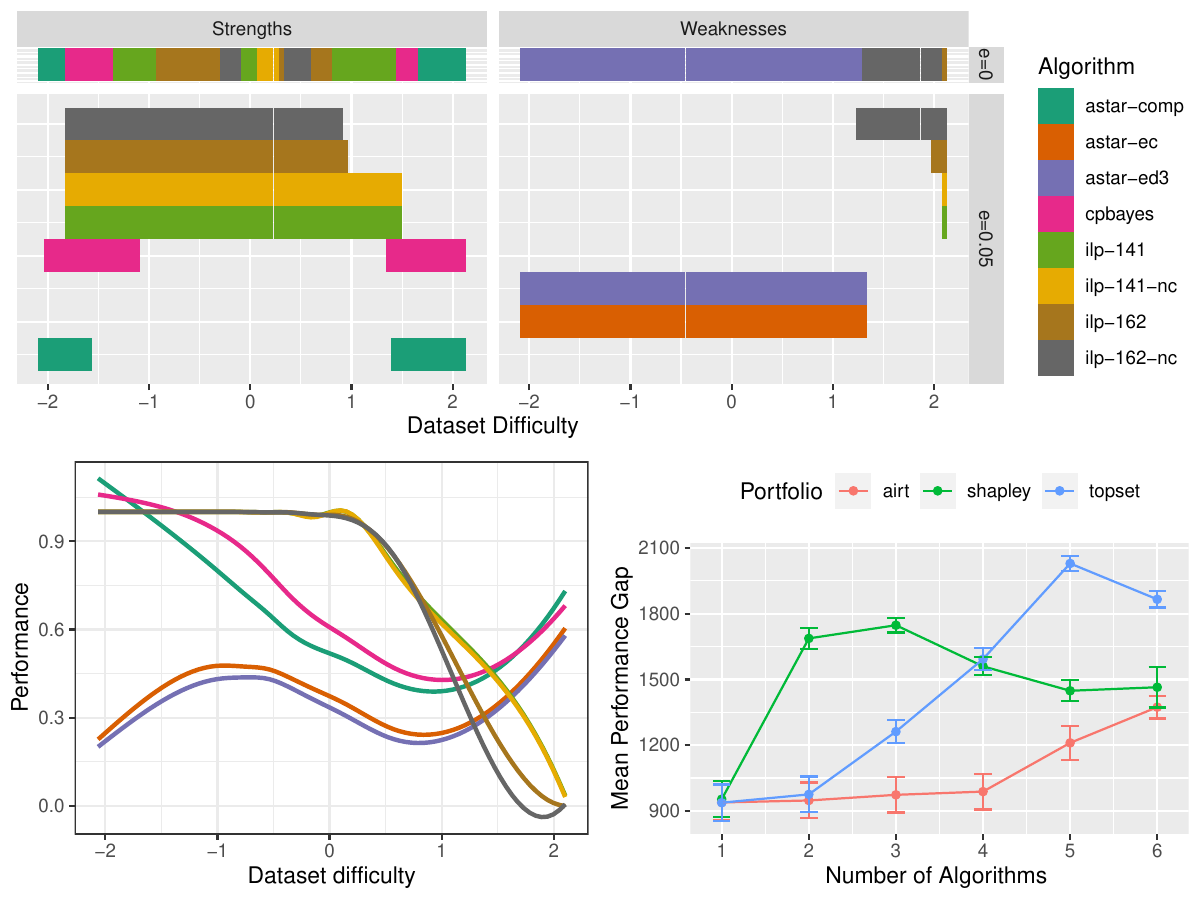}
     \caption{ Strengths and weaknesses, latent trait curves and 10-fold CV portfolio comparison for BNSL\_2016 scenario..}
     \label{fig:bnsl}
\end{figure}

Figure \ref{fig:bnsl} shows the analysis for BNSL-2016 scenario. {\color{black} Algorithms ilp-141 and ilp-141-nc have similar latent trait curves. Similarly, ilp-162 and ilp-162-nc are also similar. Furthermore, astar-ec and astar-ed3 have similar curves. Lastly, cpbayes and astar-comp have somewhat similar curves.} Algorithms \textit{cpbayes} and \textit{astar-comp} display strengths for easy and very difficult problems while \textit{astar-ec} and \textit{astar-ed3} are weak for most of the problem space. {\color{black} Algorithms ilp-141, ilp-141-nc, ilp-162 and ilp-162-nc have strengths for a large part of the problem space.} Algorithm portfolio comparison shows that airt achieves good performance.

\subsubsection{ SAT18-EXP-ALGO}
\begin{figure}[!ht]
    \centering
     \includegraphics[scale=0.7]{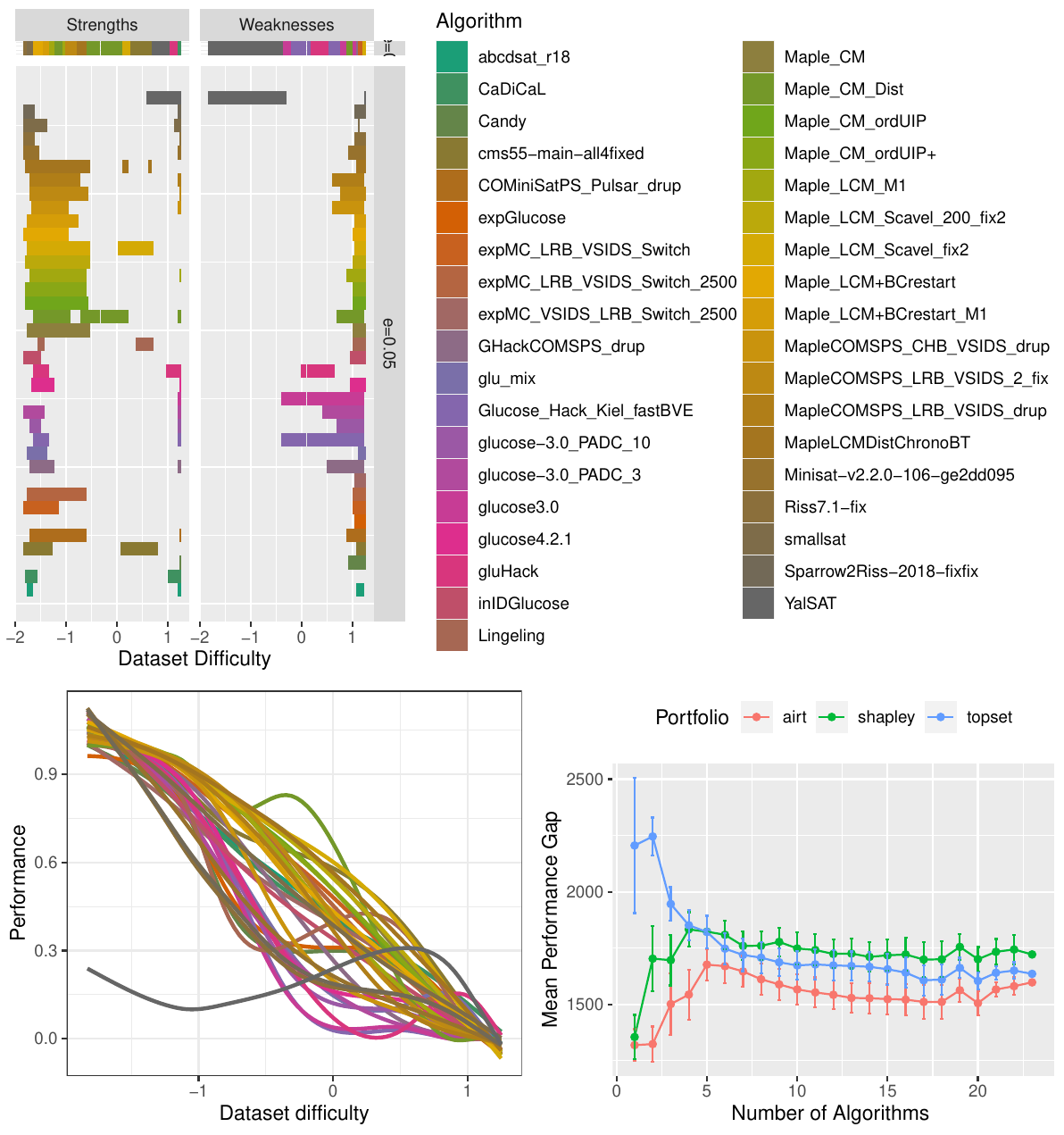}
    \caption{ Strengths and weaknesses, latent trait curves and 10-fold CV portfolio comparison for SAT18\_EXP\_ALGO scenario..}
    \label{fig:sat18}
\end{figure}

Figure \ref{fig:sat18} shows the analysis for SAT18\_EXP\_ALGO scenario. The latent trait curves are somewhat similar, but not too similar as in SAT11\_INDU. The algorithm \textit{YalSAT}, depicted by a gray shade, is weak for easier instances and strong for difficult instances. Hence is comes up in both strengths and weakness diagrams. Remarkably, the latent trait curve appears at the bottom on the left hand side and bends and ends up at the top at the right-most side. {\color{black} Another upward bend is observed at $\delta \approx -0.5$ by Maple\_CM\_Dist algorithm showing a unique strength of this algorithm.} The airt portfolio  achieves good performance for this scenario.

\end{document}